\documentclass{article}

\scrollmode
\usepackage{amsmath}
\usepackage{amsfonts}
\usepackage{amssymb}
\usepackage{latexsym}
\usepackage{stmaryrd}
\usepackage{array}
\usepackage{exscale}
\newcommand{\nc}{\newcommand}
\newcommand{\ol}{\overline}
\newcommand{\ul}{\underline}
\newcommand{\es}{\emptyset}
\newcommand{\sm}{\setminus}
\newcommand{\ve}{\varepsilon}
\newcommand{\vp}{\varphi}
\newcommand{\bw}{\bigwedge}
\newcommand{\bv}{\bigvee}
\newcommand{\bc}{\bigcup}

\newcommand{\Lra}{\Leftrightarrow}

\newcommand{\Ra}{\Rightarrow}

\newcommand{\ra}{\rightarrow}

\newcommand{\la}{\leftarrow}
\newcommand{\lra}{\leftrightarrow}

\newcommand{\sse}{\subseteq}

\newcommand{\fa}{\forall}
\newcommand{\ex}{\exists}
\newcommand{\mr}{\mathrm}
\newcommand{\mc}{\mathcal}

\newcommand{\DMO}{\DeclareMathOperator}
\newcommand{\DST}{\displaystyle}

\newcommand{\NN}{\mathbb{N}}
\newcommand{\NNZ}{\NN_0}

\newcommand{\PP}{\mathbb{P}}

\newcommand{\TT}{\mathbb{T}}
\newcommand{\und}{{\:\wedge\:}} 
\newcommand{\oder}{{\:\vee\:}} 
\newcommand{\mb}{{\:|\:}} 
\newcommand{\set}[1]{\{ #1 \}}
\newcommand{\setb}[1]{\big \{ \, #1 \, \big \}}
\nc{\simlvi}[1]{\!\sim_{#1}}
\nc{\apprel}[3]{{#1}(#2)_{(#3)}} 
\nc{\cmpli}[1]{\complement^1_{#1}} 
\nc{\cmplzi}[1]{\complement^0_{#1}} 
\nc{\cmplzoi}[1]{\complement^*_{#1}} 
\newcommand{\tb}[2]{\set{#1, \dots, #2}} 
\providecommand{\abs}[1]{\lvert #1 \rvert} 
\DeclareMathOperator{\pot}{\PP} 
\newcommand{\floor}[1]{\lfloor #1 \rfloor}

\nc{\Prim}{\mc{PR}} 

\newcommand{\Va}{\mc{V\hspace{-0.1em}A}}

\newcommand{\Lit}{\mc{LIT}}
\newcommand{\Cl}{\mc{CL}}
\newcommand{\Cls}{\mc{CLS}}

\newcommand{\Pcls}[1]{#1\mbox{--}\Cls}

\newcommand{\Pass}{\mc{P\hspace{-0.32em}ASS}}

\newcommand{\Sat}{\mc{SAT}}

\newcommand{\Usat}{\mc{USAT}}

\newcommand{\Musat}{\mc{M\hspace{0.8pt}U}} 
\newcommand{\Musati}[1]{\Musat_{\!#1}} 
\newcommand{\Smusat}{\mc{S}\Musat} 
\newcommand{\Smusati}[1]{\Smusat_{\!#1}}

\nc{\Clsoo}{\Cls^{1,1}} 
\DeclareMathOperator{\lit}{lit}
\DeclareMathOperator{\var}{var}

\newcommand{\Clash}{\mc{HIT}} 

\newcommand{\Uclash}{\mc{U}\Clash} 
\newcommand{\Uclashi}[1]{\Uclash_{\!\!#1}}

\newcommand{\Ho}{\mc{HO}} 
\newcommand{\Rho}{\mc{R}\Ho} 
\newcommand{\Qho}{\mc{Q}\Ho}

\newcommand{\Dnf}{\mathrm{DNF}}

\DeclareMathOperator{\res}{\diamond} 

\DeclareMathOperator{\comp}{Comp} 
\DeclareMathOperator{\compr}{\comp_R} 

\DeclareMathOperator{\hardness}{hd}
\DMO{\phardness}{phd} 
\DMO{\whardness}{whd} 
\DMO{\hts}{hs} 
\newcommand{\php}{\mathrm{PHP}}
\newcommand{\ephp}{\mathrm{EPHP}} 
\DeclareMathOperator{\nds}{nds} 
\DeclareMathOperator{\lvs}{lvs} 
\DeclareMathOperator{\nlvs}{\#lvs} 
\DeclareMathOperator{\nnds}{\#nds} 
\DeclareMathOperator{\height}{ht}
\DeclareMathOperator{\depth}{d}

\newcommand{\pab}[1]{\langle #1 \rangle}
\newcommand{\pao}[2]{\langle #1 \ra #2 \rangle}

\nc{\bth}[1]{\langle{#1}\rangle} 
\DMO{\rsub}{r_S} 
\DMO{\rk}{r} 
\DMO{\rki}{r_{\infty}} 
\nc{\rslur}{\xrightarrow{\text{SLUR}}} 
\nc{\rslurs}{\rslur_{\!*}} 
\DMO{\slur}{slur} 
\nc{\Slur}{\mc{SLUR}} 
\nc{\rkslur}[1]{\xrightarrow{\text{SLUR}_{#1}}} 
\nc{\rkslurs}[1]{\rkslur{#1}_{\!*}} 
\nc{\Altsluri}[1]{\Slur(#1)}
\nc{\Altslurstari}[1]{\Slur\text{\textasteriskcentered}(#1)}
\nc{\Canoni}[1]{\mr{CANON}(#1)}
\nc{\rkslurstar}[1]{\xrightarrow{\text{SLUR\textasteriskcentered}#1}} 
\nc{\rkslursstar}[1]{\rkslurstar{#1}_{\!*}} 
\DMO{\slurstar}{\slur\!\text{\textasteriskcentered}}
\nc{\Urefc}{\mc{UC}}
\nc{\Propc}{\mc{PC}}
\nc{\Wrefc}{\mc{WC}} 
\DeclareMathOperator{\nsat}{\#sat}

\DeclareMathOperator{\vdeg}{vd} 
\DeclareMathOperator{\minvdeg}{\mu\!\vdeg} 
\DMO{\varmvd}{\var_{\minvdeg}} 
\DMO{\nfc}{fc} 
\DMO{\maxnfc}{\nu\!\nfc} 
\newcommand{\OKsolver}{\texttt{OKsolver}}

\nc{\svbf}{\mc{VB}} 
\nc{\svbfs}{\mc{VB}^*} 
\DMO{\potp}{pp} 
\DMO{\potprec}{NM} 
\DMO{\minnonmer}{\mu{}nM} 
\DMO{\varsing}{\var_s} 
\DMO{\varosing}{\var_{1s}} 
\DMO{\varnosing}{\var_{\neg1s}} 
\nc{\Musatns}{\Musat'} 
\nc{\Musatnsi}[1]{\Musati{#1}'}
\nc{\Smusatns}{\Smusat'} 
\nc{\Smusatnsi}[1]{\Smusati{#1}'}
\nc{\Uclashns}{\Uclash'} 
\nc{\Uclashnsi}[1]{\Uclashi{#1}'}
\nc{\tsdp}{\xrightarrow{\text{sDP}}}
\nc{\tsdps}{\tsdp_{\!*}}
\nc{\tosdp}{\xrightarrow{\text{1sDP}}}
\nc{\tosdps}{\tosdp_{\!*}}
\DMO{\sdp}{sDP} 
\DMO{\osdp}{sDP_1} 
\nc{\cflmusat}{\mc{CF}\Musat} 
\nc{\cflmusati}[1]{\mc{CF}\Musati{#1}}
\nc{\cflimusat}{\mc{CFI}\Musat} 
\DMO{\sNF}{sNF} 
\DMO{\eqp}{eqp} 
\DMO{\sgp}{sp} 
\DMO{\singind}{si} 
\DMO{\osingind}{si_1} 
\DMO{\shyp}{svh} 
\DMO{\sdph}{ssh} 
\DMO{\msdph}{mss} 
\DMO{\osdph}{ssh_1} 
\DMO{\mosdph}{mss_1} 
\newcommand{\Mps}{\mathcal{MPS}} 
\DMO{\mps}{mps} 
\DMO{\purec}{puc} 
\DMO{\doping}{D}
\DeclareMathOperator{\primec}{prc} 
\DeclareMathOperator{\vcan}{vct} 
\DeclareMathOperator{\cant}{ct} 
\nc{\glue}[4]{\mr{glue}((#1,#2), (#3,#4))} 
\DMO{\fvdglue}{\boxplus} 
\nc{\gluea}[3]{#1 \boxplus_{#3} #2} 
\usepackage{theorem} 
\usepackage[driverfallback=hypertex]{hyperref}
\newtheorem{defi}{Definition}[section]
\newtheorem{lem}[defi]{Lemma}
\newtheorem{thm}[defi]{Theorem}
\newtheorem{corol}[defi]{Corollary}

\newtheorem{conj}[defi]{Conjecture}

\newtheorem{examp}[defi]{Example}

\newenvironment{prf}{\noindent\textbf{Proof:}\;}{\par\noindent\ignorespacesafterend}

\newcommand{\Qed}{\hfill $\square$}
\nc{\bm}{\boldmath}
\nc{\bmm}[1]{\mbox{\bm$\DST #1$}}
\nc{\mi}[1]{\bmm{\mathrm{(#1):}} \quad}

\usepackage[active]{srcltx}
\usepackage{a4}
\usepackage[all,poly]{xy}
\usepackage{float}

\nc{\trighyp}[2]{T_{\hspace{-0.08ex}#2}(#1)}

\DMO{\thardness}{thd}

\DMO{\exstrahler}{HS}
\nc{\exhst}[2]{\exstrahler(#1,#2)}
\nc{\exstrahlersize}[2]{\alpha(#1,#2)}

\DMO{\smuo}{F^1}
\DMO{\tsmuo}{T^1}
\nc{\Tsmuo}{\mc{T}_1}

\DMO{\cantm}{\cant^{\text{--}}}

\begin{document}

\title{Towards a theory of\\ good SAT representations}

\author{Matthew Gwynne and Oliver Kullmann\\
  \qquad {\small \url{http://cs.swan.ac.uk/~csmg/}} \qquad
  {\small \url{http://cs.swan.ac.uk/~csoliver}}\\
  \href{http://www.swan.ac.uk/compsci/}{Computer Science Department}\\
  \href{http://www.swan.ac.uk/}{Swansea University}\\
  Swansea, UK
}

\maketitle

\begin{abstract}
  We consider the fundamental task of representing a boolean function $f$ by a conjunctive normal form (clause-set) $F$ for the purpose of SAT solving. The boolean function $f$ here acts as a kind of constraint, like a cardinality constraint or an S-box in a cryptosystem, while $F$ is a subset of the whole SAT problem to be solved. The traditional approach towards ``good'' properties of $F$ considers ``arc consistency'', which demands that for every partial instantiation of $f$, all forced assignments can be recovered from the corresponding partial assignment to $F$ via unit-clause propagation (UCP). We propose to consider a more refined framework: First, instead of considering the above \emph{relative condition}, a relation between $f$ and $F$, we consider an \emph{absolute condition}, namely that goodness of $F$ is guaranteed by $F$ being element of a suitable target class. And second, instead of just considering UCP, we consider hierarchies of target classes, which allow for different mechanisms than UCP and allow for size/complexity trade-offs.

  The hierarchy $\Urefc_k$ of \emph{unit-refutation complete clause-sets of level $k$}, introduced in \cite{GwynneKullmann2012SlurSOFSEM,GwynneKullmann2012Slur,GwynneKullmann2012SlurJ}, provides the most basic target classes, that is, $F \in \Urefc_k$ is to be achieved for $k$ as small as feasible. Here $\Urefc_1 = \Urefc$ has been introduced in \cite{Val1994UnitResolutionComplete} for the purpose of knowledge compilation. In general, $\Urefc_k$ is the set of clause-sets $F$ such that unsatisfiable instantiations (by partial assignments) are recognisable by $k$-times nested unit-clause propagation. We also touch upon the hierarchy $\Propc_k$ of \emph{propagation complete clause-sets of level $k$}, where $\Propc_1 = \Propc$ has been introduced in \cite{BordeauxMarquesSilva2012KnowledgeCompilation}. The hierarchy $\Propc_k$ refines the hierarchy $\Urefc_k$ by providing intermediate layers. In order to make use of full resolution, we consider the hierarchy $\Wrefc_k$ of \emph{width-refutation complete clauses-sets of level $k$}, employing an improved notion of width (so that we always have $\Urefc_k \sse \Wrefc_k$).

  Via the absolute condition, the quality of the representation $F$ is fully captured by the target class, and the only relation between $f$ and $F$ is that $F$ must ``represent'' $f$. If $F$ does not contain new variables, then this means that $F$ is equivalent to $f$, while with new variables the satisfying assignments of $F$ projected to the variables of $f$ must be precisely the satisfying assignments of $f$. Without new variables, the relative and absolute condition coincide, but with new variables, the absolute condition is stronger. As we remark in this article, for the relative condition and new variables at least the hierarchies $\Urefc_k$ and $\Propc_k$ collapse, and we also conjecture that the $\Wrefc_k$ hierarchy collapses. The main result of this article is that without new variables, none of these hierarchies collapses. That means that there are boolean functions with only exponential-size equivalent clause-sets at level $k$, but with poly-size equivalent clause-sets at level $k+1$.

  Representations with new variables in general allow shorter representations. However representations without new variables can be systematically searched for, opening a new algorithmic avenue for good SAT representations, where in a pre-processing phase the representation is being optimised. When using a two-stage approach, then first non-algorithmically a representation with new variables can be constructed, which then can be optimised by searching for an equivalent better clause-set.

  We believe that many common CNF representations either already fit into the $\Urefc_k$ scheme or can be made fit by slight improvements. We give some basic tools to construct representations in $\Urefc_1$, now with new variables and based on the Tseitin translation. We conclude with a discussion of open problems and future directions, with special emphasis on separations for the various hierarchies involved.
\end{abstract}

\tableofcontents

\section{Introduction}
\label{sec:intro}

It has been shown that the practical performance of SAT solvers can depend heavily on the SAT representation used. See for example \cite{BailleuzBoufkhad2003CardinalityConstraints,Sinz2005CardinalityConstraints,Een2006Translating} for work on cardinality constraints, \cite{TamuraTagaKitagawaBanbara2009OrderEncoding,2011CompactOrderEncoding} for work on general constraint translations, and \cite{JovanovicKreuzer2010AlgAttackSAT,GwynneKullmann2011TranslationsPrelim} for investigations into different translations in cryptography. In order to obtain ``good'' representations, until now the emphasis has been on translating constraints into SAT such that ``arc-consistency'' is ``maintained'', via unit-clause propagation; for an introduction into the literature see Section 22.6.7 in \cite{RM09HBSAT}, while various case-studies can be found in \cite{Gent2002ArcConsistency,BailleuzBoufkhad2003CardinalityConstraints,Sinz2005CardinalityConstraints,Een2006Translating,BarahomaJungKatsirelosWalsh2008EncodingDNNF,BailleuxBoufkhadRoussel2009PBCNF}. That is, for all (partial) assignments to the variables of the constraint, the task is to ensure that if there is a forced assignment (i.e., some variable which must be set to a particular value to avoid inconsistency), then unit-clause propagation (UCP) is sufficient to find and set this assignment. In a similar vein, there is the class $\Propc$ of propagation-complete clause-sets (see \cite{BordeauxMarquesSilva2012KnowledgeCompilation}), containing all clause-sets for which unit-clause propagation is sufficient to detect all forced assignments.

Maintaining arc-consistency and propagation-completeness may at a glance seem the same concept. However there is an essential difference. When translating a constraint into SAT, typically one does not just use the variables of the constraint, but one adds auxiliary variables to allow for a compact representation. Now when speaking of maintaining arc-consistency, one only cares about assignments to the \emph{constraint variables}. But propagation-completeness deals only with the representing clause-set, thus can not know about the distinction between original and auxiliary variables, and thus it is a property on the (partial) assignments over \emph{all} variables! So a SAT representation, which maintains arc-consistency via UCP, will in general not be propagation-complete, due to assignments over both constraint \emph{and} new variables yielding a forced assignment or even an inconsistency which UCP doesn't detect; see Example \ref{exp:relhdct} and Lemma \ref{lem:exphddnf}. In contrast to this, for the basic concepts of ``good'' representations investigated in this paper, considering \emph{all} variables is a fundamental feature. This motivates our focus on \emph{classes of clause-sets} (as the target of good SAT representations), rather than \emph{maintaining} consistency over some higher level constraint network, since in this way we have full control of the properties at the level of the SAT solver (at the CNF-level).

In \cite{JarvisaloJunttila2009LimitRestrictedLearning} it is shown that conflict-driven solvers with branching restricted to input variables have only superpolynomial run-time on $\ephp_n'$, an Extended Resolution extension to the pigeon-hole formulas, while unrestricted branching determines unsatisfiability quickly (see Subsection \ref{sec:conclhard} for more on this). Also experimentally it is demonstrated in \cite{JarvisaloNiemala2008StructuralBranchingExperiments} that input-restricted branching can have a detrimental effect on solver times and proof sizes for modern CDCL solvers. This adds motivation to our fundamental choice of considering \emph{all} variables (rather than just input variables), when deciding what properties we want for SAT translations. We call this the ``absolute (representation) condition'', taking also the auxiliary variables into account, while the ``relative condition'' only considers the original variables. Besides avoiding the creation of hard unsatisfiable sub-problems, the absolute condition also enables one to study the target classes, like $\Propc$, on their own, without relation to what is represented.

In a certain sense, the underlying idea of maintaining arc-consistency and pro\-pa\-ga\-tion-com\-ple\-te translations is to compress all of the constraint knowledge into the SAT translation, and then to use UCP to extract this knowledge when appropriate. Motivated by the absolute condition, in \cite{GwynneKullmann2012SlurSOFSEM,GwynneKullmann2012Slur,GwynneKullmann2012SlurJ} we considered the somewhat more fundamental class $\Urefc$ of refutation complete clause-sets, introduced in \cite{Val1994UnitResolutionComplete} as a method for propositional knowledge compilation, and studied its properties. Rather than requiring that UCP detects all forced assignments (as for $\Propc$), a clause-set is in $\Urefc$ iff for all partial assignments resulting in an unsatisfiable clause-set UCP detects this.

So we have $\Urefc$ and $\Propc$ as potential target classes for ``good'' SAT representations. In both cases we know, that if the SAT solver ends up in an unsatisfiable part of the search space, then the ubiquitous unit-clause propagation will immediately determine this and the solver will avoid potentially exponential work. However, how to come up with representations in $\Urefc$? There are easy examples of ``good'' clause-sets which are not in $\Urefc$, e.g., 2-CNF. Given that UCP is a relatively simple mechanism, perhaps it would be better to consider more powerful inference methods allowing for a greater variety and possibly shorter representations (``more compression'')? For this end, to add more power, we introduce ``hardness measurement''.

\subsection{A general framework: hierarchies and measurement}
\label{sec:introframe}

 In \cite{GwynneKullmann2012SlurSOFSEM,GwynneKullmann2012Slur,GwynneKullmann2012SlurJ}, using generalised unit-clause propagation $\rk_k$ (with $\rk_1$ being UCP) introduced in \cite{Ku99b,Ku00g}, we developed a hierarchy $\Urefc_k$ (with $\Urefc_1 = \Urefc$) of clause-sets of ``hardness'' at most $k$, that is, refutation is (always) possible via $\rk_k$. Replacing $\rk_1$ with $\rk_k$ in the same way in $\Propc$ yields the propagation-completeness hierarchy $\Propc_k$ (with $\Propc_1 = \Propc$). In the limit these hierarchies cover all clause-sets, with the levels of the hierarchy offering the possibility to trade complexity of the inference method ($\rk_k$) for size of the representation. Generalising existing results we show in Lemma 6.5 of \cite{GwynneKullmann2012Slur,GwynneKullmann2012SlurJ} that various poly-time solvable SAT classes are contained within levels of the $\Urefc_k$ hierarchy. That is $\Ho \subset \Rho \subset \Urefc_1$ (Horn and renamable Horn clause-sets), $\Pcls{2} \subset \Qho \subset \Urefc_2$ (2-clause-sets and q-Horn clauses-sets, see Section 6.10.2 in \cite{CramaHammer2011BooleanFunctions} and \cite{Ma99j}) and $\Ho_k \subset \Urefc_k$ (generalised Horn clause-sets, see \cite{Kl93}).

There are strong proof theoretic connections for $\Urefc_k$ to tree-resolution. In \cite{JarvisaloMatsliahNordstromZivny2012ProofCompHardnessSAT} the argument is made that tree-resolution complexity can not provide a good measure of hardness of instances for SAT solving, citing the ability of CDCL solvers to simulate exponentially more powerful full resolution (see \cite{AtseriasFichteThurley2009ClauseLearningBoundedWidth} for evidence that CDCL solvers can simulate full resolution). However, the aim of $\Urefc_k$ is not to \emph{measure} hardness, but to offer a \emph{target class} for SAT translation. In this respect tree-resolution complexity measures are ideal, because they provide the strongest translations, and upper-bound measures for full resolution.

On the other hand, for tighter target classes in the case of full resolution, we also consider the notion of width-hardness as introduced in \cite{GwynneKullmann2012Slur,GwynneKullmann2012SlurJ}, based on the width-based hierarchies of unsatisfiable clause-sets in \cite{Ku99b,Ku00g}. That is, a clause-set is in $\Wrefc_k$, the hierarchy of clause-sets of width-hardness $k$, iff under any partial assignment resulting in an unsatisfiable clause-set there is a ``$k$-resolution'' refutation as introduced in \cite{Kl93}. Here, unlike the typical notion, we allow resolutions where only \emph{one} parent clause needs to have length at most $k$, and thus properly generalising unit-resolution (one could speak of ``asymmetric width'' here, compared to the standard ``symmetric width''). This allows to simulate nested input resolution, and thus we have $\Urefc_k \sse \Wrefc_k$ for all $k$, whereas otherwise in the standard (symmetrical) sense even Horn clause-sets require unbounded width (recall that $\Ho \subset \Urefc_1$).

Fundamental for each hierarchy is an underlying measure $h_0: \Usat \ra \Cls$, measuring the ``hardness'' of unsatisfiable clause-sets, which is extended to $h: \Cls \ra \NNZ$, where $h(F)$ for an arbitrary clause-set $F$ measures the ``hardness'' to derive any conclusion $F \models C$ for clauses $C$, by letting $h(F)$ be the maximum of $h_0(\vp * F)$ over all partial assignments $\vp$ such that application yields an unsatisfiable result $\vp * F$. The hierarchy at level $k$ collects all $F$ with $h(F) \le k$. For the PC-UC hierarchy the corresponding measure $\phardness(F)$ resp.\ $\hardness(F)$ can be described in many ways; most intuitive from a SAT point of view is to say that it measures the necessary nesting level of UCP, that is, which $\rk_k$ is required.

\paragraph{A precursor} A generalisation of $\Urefc$ was already discussed in \cite{Val2000UnitResolutionComplete}. Assuming a polytime SAT-decision algorithm $P: \mc{C} \ra \set{0,1}$ for some $\mc{C} \sse \Cls$, the class $P\mc{C} \sse \Cls$ of ``P-complete'' clause-sets is defined as the set of $F \in \Cls$ such that for all implicates $C$ holds $P(\vp_C * F) = 0$.\footnote{\cite{Val2000UnitResolutionComplete} actually favours adding unit-clauses to $F$, but we consider applying partial assignments as more fundamental.} This is an obvious generalisation of $\Urefc_k$, when using $\mc{C}_k := \set{F \in \Cls : \rk_k(F) \in \set{\top,\set{\bot}}}$ and $P_k: \mc{C}_k \ra \set{0,1}$ with $P_k(F) = 1 \Lra \rk_k(F) = \top$. But it does not cover the hierarchies $\Propc_k$ or $\Wrefc_k$, which are based on different principles.\footnote{This is obvious for $k \ge 1$ and $\Propc_k$, since $\Propc_k \cap \Usat = \Urefc_k \cap \Usat$, while $\Propc_k \cap \Sat \subset \Urefc_k \cap \Sat$. We conjecture that for $k \ge 3$ there is no (polytime) $P$ with $P\mc{C} = \Wrefc_k$ (as remarked in Subsection \ref{sec:prelimwhdWC}, for $k \in \set{0,1,2}$ there exists such a $P$).} We note the conceptual weakness of demanding a SAT-decision algorithm $P$, where actually only a means for detecting \emph{unsatisfiability} is needed.

\cite{Val2000UnitResolutionComplete} continues by considering the (generic) hierarchy $(\Pi_k)_{k \in \NNZ}$ from \cite{Pr96}, a precursor of \cite{Ku99b}. $\Pi_0 \sse \Cls$ in principal is arbitrary, but is assumed to be polytime decidable and SAT-decidable. Then $\Pi_k$ for $k > 0$ is the set of $F \in \Cls$ such that $F \in \Pi_{k-1}$ or there is a literal $x \in \lit(F)$ with $\pao x1 * F \in \Pi_{k-1}$ and $\pao x1 * F \in \Pi_k$. We note that if we choose $\Pi_0 = \set{F \in \Cls : \bot \in F}$, then $\Pi_k = \Urefc_k \cap \Usat$ for all $k \ge 0$. However this choice for $\Pi_0$ was never considered for that hierarchy from \cite{Pr96}, which might have two reasons: Implicit preference is given to classes $\Pi_0$ closed under sub-clause-set formation (see Section 6.3 in \cite{GwynneKullmann2012SlurJ} for more discussions on this issue). And furthermore SAT and UNSAT is not distinguished in \cite{Pr96} and in subsequent work directly relying on it; see Subsection 1.2 in \cite{Ku99b} for a discussion of this. So the four choices for $\Pi_0$ considered in \cite{Val2000UnitResolutionComplete} are $\Ho$, $\Pcls{2}$, $\Rho$ and $\Qho$. Accordingly $\Urefc_0 \cap \Usat$ is not contained at any $\Pi_k$, and so not even $\rk_1$ on unsatisfiable clause-set is covered by the considered hierarchies.

Due to these weaknesses, \cite{Val2000UnitResolutionComplete} does not consider a hierarchy generalising $\Urefc$. From our point of view one could say, that $\Pi_k$ is only considered as a resource for polytime recognition of certain instances for $\Urefc_k$ resp.\ $\Urefc_{k+1}$; compare Subsections 6.2, 6.3 in \cite{GwynneKullmann2012SlurJ} for results in this direction.

\subsection{Representation of boolean functions}
\label{sec:introrep}

By definition each $\Propc_k, \Urefc_k, \Wrefc_k$ is just a class of clause-sets. However when using these classes for representing boolean functions, then there are further aspects. In general, for translations to SAT a typical path is
\begin{center}
  \textbf{Problem} \quad $\ra$ \quad \textbf{Constraints} \quad $\ra$ \quad $\underbrace{\text{\textbf{Boolean functions} \quad $\ra$ \quad \textbf{SAT}}}_{\textbf{our focus}}$.
\end{center}
By considering target classes for ``good'' SAT representations. we focus our attention on the final stage, the translation of boolean functions to SAT, ignoring the issue of encoding non-boolean domains into the boolean. Now there are three main dimensions to consider (choices to make):
\begin{enumerate}
\item Inference properties (\bmm{\Propc_k} versus \bmm{\Urefc_k} versus \bmm{\Wrefc_k}, and the value of $k$): How strong a property we require of the clause-set we translate to ($\Propc_k$ is strongest, $\Wrefc_k$ weakest, and the lower $k$ the stronger the condition).
\item \textbf{Logical equivalence} versus \textbf{new variables}: whether the SAT translation is equivalent to the input function (i.e., no new variables), or uses new variables to extend the original function.
\item \textbf{Relative} versus \textbf{absolute condition}: in case new variables are used, whether the property we require for the translation refers to partial assignments only on the original variables or also on the new variables.
\end{enumerate}
See Subsection \ref{sec:cnfrep} for more on the relative condition; our point of view is that the absolute condition is fundamental for the representation of boolean functions, not the relative condition (which has been dominant in the literature until now).

In the area of Knowledge Compilation, the task is also to represent (``compile'') boolean functions to allow good inference under (repeated) queries. In particular, one wants to find a representation for a boolean function which allows queries such as clausal entailment ($F \models C$), equivalence, and model counting to be answered efficiently (in polynomial time). In this sense, we can think of ``finding a good representation'' as a form of \emph{SAT knowledge compilation}, where we care (only) about clausal entailment, since CNF-clauses directly correspond to falsifying partial assignments. \cite{CadoliDonini1997SurveyKnowledgeComp} gives an overview of the CNF-based target languages (prime implicates, $\Urefc$, $\Pcls{2}$, Horn clause-sets). \cite{FargierMarquis2008ExtendKCMapKromHornAff} consider disjunctions of simple CNF classes. \cite{DarwicheMarquis2002KCmap} provides an overview of target compilation languages based on ``nested'' (graph-based) classes, namely variants of NNF, DNNF and BDDs. In all cases query complexity and succinctness is investigated. We focus on CNF representations, since we want good representations for current resolution-based SAT solvers. All of the CNF classes studied in \cite{CadoliDonini1997SurveyKnowledgeComp,DarwicheMarquis2002KCmap,FargierMarquis2008ExtendKCMapKromHornAff} are included at the first three levels of the hierarchy $\Urefc_k$, namely, sets of prime implicates in $\Urefc_0$, (renamable) Horn clause-sets in $\Urefc_1 = \Urefc$, and $\Pcls{2}$ in $\Urefc_2$. Translations from target classes such as DNNF to CNF are also of interest and fit into the framework of $\Urefc_k$ via using new variables; see Section \ref{sec:cantrans} for the most basic positive considerations. And see \cite{BarahomaJungKatsirelosWalsh2008EncodingDNNF} for a basic negative result, characterising what can be represented under the relative condition (i.e., arc consistency).

\subsection{Strictness of hierarchies}
\label{sec:introstrictgen}

A fundamental question is the strictness of these hierarchies $\Propc_k$, $\Urefc_k$, and $\Wrefc_k$ in each of those two remaining dimensions. That is, whether each level offers new possibilities for polysize representations of (sequences of) boolean functions within the confines of the specified dimensions, i.e., relative versus absolute and without versus with new variables. Using the basic choice of the absolute condition, we have six proper hierarchies (3 conjectured, 3 proven), namely $\Propc_k$, $\Urefc_k$ and $\Wrefc_k$ for representations without (Theorem \ref{thm:separation}) and with new variables (Conjecture \ref{con:sepext}).\footnote{Regarding $\Propc_k$ we get only a separation of $\Propc_k$ and $\Propc_{k+2}$; this will be addressed in future work.} However when using representations based on the relative condition (and using new variables), then all these hierarchies collapse to their first level: two collapses are similar to existing results, while the collapse for $\Wrefc_k$ should follow also in this way, and is spelled out as Conjecture \ref{con:collapseWrefc}.

Considered together, under the relative condition only the levels $\Propc_0 \subset \Urefc_0 \subset \Propc_1$ are strict regarding polysize representations, where the two classes $\Propc_0 \subset \Urefc_0$ do not gain anything from the new variables, while everything of $\Propc_k$, $\Urefc_k$ and $\Wrefc_k$ for $k \ge 1$ can be reduced (in polytime, with exponent depending on $k$) to $\Propc_1 = \Propc$. And $\Propc$ under the relative condition is the same as the well-known condition of ``arc consistency'' for SAT translation. The main result of this paper, that $\Propc_k$, $\Urefc_k$ and $\Wrefc_k$ for the absolute condition and without new variables do not collapse, shows that a rich structure was hidden under the carpet of the relative condition aka arc consistency. A basic difference between relative and absolute condition is that under the relative condition the new variables can be used to perform certain ``computations'', since there are no conditions on the new variable other than not to distort the satisfying assignments. This is used to show the collapse to arc consistency, by encoding the stronger condition into the clause-sets in such a way that UCP can perform the ``computations''.

\subsection{Understanding the combinatorial structure of satisfiable clause-sets}
\label{sec:introcombsat}

To be able to prove properties about all equivalent representations of some clause-set $F$, we must be able to understand its combinatorial structure in relation to the set of all its prime implicates. The notion of minimal unsatisfiability (MU) and minimally unsatisfiable subsets (MUS) is important in understanding the combinatorics of unsatisfiable clause-sets (see \cite{Kullmann2007HandbuchMU,MarquesSilva2012MUS}). To understand the structure of satisfiable clause-sets and their associated boolean functions, we now consider the concept of ``minimal premise sets'' (MPS) introduced in \cite{Kullmann2007ClausalFormZII}. The notion of MPS generalises that of MU by considering clause-sets $F$ which are minimal w.r.t implying \emph{any} clause $C$ rather than just those implying $\bot$. And accordingly we consider the minimal-premise subsets (MPSS) of a clause-set $F$.

Every prime implicate $C$ of a clause-set $F$ has an associated MPSS (just consider the minimal sub-clause-set of $F$ that implies $C$), but not every MPSS of $F$ yields a prime implicate (e.g., consider the MPSS $\set{C}$ for some non-prime clause $C \in F$). However, by ``doping'' the clause-set, i.e., adding a new unique variable to every clause, every clause in an MPSS $F'$ makes a unique contribution to its derived clause $C$. This results in a new clause-set $\doping(F)$ which has an exact correspondence between its minimal premise sets (which are (essentially) also those of $F$) and its prime implicates. In this way, by considering clause-sets $F$ with a very structured set of minimal premise subsets, we can derive clause-sets $\doping(F)$ with very structured set of prime implicates.

\subsection{The UC hierarchy is strict regarding equivalence}
\label{sec:introstricteq}

A sequence $(f_h)_{h \in \NN}$ of boolean functions, which separates $\Urefc_{k+1}$ from $\Urefc_k$ w.r.t.\ clause-sets equivalent to $f_h$ in $\Urefc_{k+1}$ resp.\ $\Urefc_k$, should have the following properties:
\begin{enumerate}
\item \textbf{A large number of prime implicates}: the number of prime implicates for $f_h$ should at least grow super-polynomially in $h$, since otherwise already the set of prime implicates is a small clause-set in $\Urefc_0$ (see Definition \ref{def:hardness}) equivalent to $f_h$.
\item \textbf{Easily characterised prime implicates}: the prime implicates of $f_h$ should be easily characterised, since otherwise we can not understand how clause-sets equivalent to $f_h$ look like.
\item \textbf{Poly-size representations}: there must exist short clause-sets in $\Urefc_{k+1}$ equivalent to $f_h$ for all $h \in \NN$.
\end{enumerate}

\cite{SloanSzoerenyiTuran2005Primimplikanten_1} introduced a special type of boolean functions, called \ul{N}on-repeating \ul{U}nate \ul{D}ecision trees (NUD) there, by adding new variables to each clause of clause-sets in $\Smusati{\delta=1}$, which is the class of unsatisfiable hitting clause-sets of deficiency $\delta=1$. These boolean functions have a large number of prime implicates (the maximum regarding the original number of clauses), and thus are natural to consider as candidates to separate the levels of $\Urefc_k$. In Section \ref{sec:minpsdopingcls} we show that it is actually the underlying $\Smusat_{\delta=1}$ clause-sets that contribute the structure. The clause-sets in $\Smusat_{\delta=1}$ are exactly those with the maximum number of minimal premise sets, and then doping elements of $\Smusat_{\delta=1}$ yields clause-sets with the maximal number of prime implicates. We utilise the tree structure of $\Smusat_{\delta=1}$ to prove lower bounds on the size of equivalent representations of doped $\Smusat_{\delta=1}$ clause-sets in $\Urefc_k$.

In Section \ref{sec:lowerb} we introduce the basic method (see Theorem \ref{thm:triggersetmethod}) for lower bounding the size of equivalent clause-sets of a given hardness, via the transversal number of ``trigger hypergraphs''. Using this lower bound method, in Theorem \ref{thm:nogoodksoft} we  show a lower bound on the matching number of the trigger hypergraph of doped ``extremal'' $\Smusati{\delta=1}$-clause-sets. From this follows immediately Theorem \ref{thm:separation}, that for every $k \in \NNZ$ there are polysize clause-sets in $\Urefc_{k+1}$, where every equivalent clause-set in $\Wrefc_k$ is of exponential size. Thus the $\Urefc_k$ as well as the $\Wrefc_k$ hierarchy is strict regarding equivalence of polysize clause-sets.

In \cite{Val1994UnitResolutionComplete} (Example 2) a separation was already shown between $\Urefc_0$ (clause-sets containing all of their prime implicates) and $\Urefc_1 = \Urefc$, and the question was raised of the worst-case growth when compiling from an arbitrary CNF clause-set $F$ to some equivalent $F' \in \Urefc$. This question was partly answered in \cite{BKNW2009CircuitComplexity} (although the connection was not made), where the authors provide examples of poly-size clause-sets with only super-polynomial size representations in $\Urefc$, even when allowing new variables (see Subsections \ref{sec:cnfrep}, \ref{sec:conclhard}, and \cite{BeyersdorffGwynneKullmann2013PHPER} for more on the connection between \cite{BKNW2009CircuitComplexity} and $\Urefc_k$). This shows a super-polynomial lower-bound on the worst-case growth, but no method or new (larger) target-class for knowledge-compilation. Our separation result now answers the question of worst-case growth from \cite{Val1994UnitResolutionComplete} in full generality with the hierarchy $\Urefc_k$. Each level of $\Urefc_k$ is exponentially more expressive than the previous (i.e., with possible \emph{exponential} blow-up when compiling from some $F \in \Urefc_{k+1}$ to equivalent $F' \in \Urefc_k$), and so each level offers its own new, \emph{larger} class for knowledge compilation, at the expense of increased query time (now $O(\ell(F) \cdot n(F)^{2k-2})$ for $\Urefc_k$ compared to $O(\ell(F))$ for $\Urefc$). This separation, between $\Urefc_{k+1}$ and $\Urefc_k$ for arbitrary $k$, is more involved than the simple separation in \cite{Val1994UnitResolutionComplete}, due to the parameterised use of more advanced polynomial-time methods than $\rk_1$ (UCP). Especially the separation between $\Urefc_0$ and $\Urefc_1$ is rather simple, since $\Urefc_0$ does not allow any form of compression.

\subsection{Relevance of these hierarchies for SAT solving}
\label{sec:introrelSAT}

The poly-time methods used to detect unsatisfiability of instantiations of clause-sets in $\Urefc_k$ resp.\ $\Wrefc_k$ have a running-time with an exponent depending on $k$, and in the latter case also space-complexity depends in the exponent on $k$.
\begin{enumerate}
\item This seems a necessary condition for showing a separation result as shown in this paper. It is needed that the different levels are qualitatively different. And this seems very unlikely to be achievable with a parameter which would allow fixed-parameter tractability, and which thus would only be a quantitative parameter (like the number of variables), only expressing a gradual increase in complexity. See Lemma \ref{lem:collapsecanon} for an example of a collapsing hierarchy.
\item The class $\Urefc_k$ uses generalised UCP, namely the reduction $\rk_k$. Especially $\rk_2$, which is (complete) failed-literal elimination, is used in look-ahead SAT solvers (see \cite{HvM09HBSAT} for an overview) such as \OKsolver{} (\cite{Ku2002h}), \texttt{march} (\cite{HeuleMaaren2005MarchEq}) and \texttt{Satz} (\cite{LA1996}). Also conflict-driven solvers such as \texttt{CryptoMiniSat} (\cite{Soos2010CMSDesc}) and \texttt{PicoSAT} (\cite{Biere2008picosat,Biere2010PicoSATLingelingDesc}) integrate $\rk_2$ during search, and solvers such as \texttt{Lingeling} (\cite{Biere2010PicoSATLingelingDesc,Biere2012LingelingDesc}) use $\rk_2$ as a preprocessing technique. Furthermore, in general $\rk_k$ is used, in even stronger versions, in the St\r{a}lmarck-solver (see \cite{SS90,Har96,SSt98}, and see Section 3.5 of \cite{Ku99b} for a discussion of the connections to $\rk_k$), and via breadth-first ``branch/merge'' rules in \texttt{HeerHugo} (see \cite{GW00}).
\item Our example class $G^1_{k,h}$ (see Section \ref{sec:experiments}) shows experimentally that higher levels of hardness may still be solved easily by SAT solvers. These examples have such read-once resolution refutations (linear in the size of the input) which are detectable by ``2-subsumption resolution'', i.e., the replacement of two clauses $C \cup \set{v}, C \cup \set{\ol{v}}$ by one clause $C$. So with this preprocessing SAT solvers can solve them in linear time. But also without this preprocessing SAT solvers seem to solve these problems in linear time. The alternative representations have lower hardness, but due to their bigger sizes SAT solvers perform orders of magnitudes worse on the larger instances.
\item In general, a SAT solver does not need to have these mechanisms built-in in general: as practical experience shows, SAT solvers are rather good in finding resolution refutations, and the parameter $k$ in $\Urefc_k$ resp.\ $\Wrefc_k$ is just a general way of bounding resolution complexity. In \cite{DarwichePipatsrisawat2009ClauseLearnRes,DarwichePipatsrisawat2011ClauseLearnRes} it is argued that modern SAT solvers can simulate full resolution --- and this is considered to be a good property of SAT solvers. Thus they are also capable in general of finding the refutations guaranteed by $\Urefc_k$ resp.\ $\Wrefc_k$.

  An important point here is, that for theoretical reasoning all unsatisfiable instantiations must be handled, while in a SAT-solver run only a selected set of instantiations is encountered, and thus ``leaner means'' can suffice (as the practical success of SAT solving shows).
\end{enumerate}

\subsection{Tools for good representations}
\label{sec:introtools}

We conclude our investigations by considering translations based on the Tseitin translation in Section \ref{sec:analT}, and show that interesting classes of boolean function can be polynomially translated to $\Urefc$ under the absolute condition using new variables. First we discuss the notion of ``representation'' in general in Subsection \ref{sec:cnfrep}, with special emphasise on the ``relative'' versus the ``absolute condition''.

The Tseitin translation for DNF's we call ``canonical translation'', and we investigate it in Subsection \ref{sec:cantrans}. In particular, in Lemma \ref{lem:hitct} we show that every orthogonal (or ``disjoint'', or ``hitting'') DNF is translated to $\Urefc$, while in Lemma \ref{lem:hdctm} we show that actually every DNF is translated to $\Urefc$, when using the ``reduced'' canonical translation, which uses only the necessary part of the equivalences constitutive for the Tseitin translation. Applied to our examples yielding the separation of $\Urefc_{k+1}$ from $\Wrefc_k$ (Theorem \ref{thm:separation}, regarding polysize representations without new variables), we obtain a representation in $\Urefc$ in Theorem \ref{thm:extuc} (for the canonical translation), demonstrating the power of using new variables.

It has been noted in the literature at several places (see \cite{GreenbaumPlaisted1986ClauseFormTrans,JacksonSheridan2004CircuitstoSAT,Een2006Translating}), that one might use only one of the two directions of the equivalences in the Tseitin-translation. Regarding the canonical translation we have the full translation (Definition \ref{def:ct}) versus the reduced translation (Definition \ref{def:ctm}). The full translation yields $\Urefc$ for special inputs (Lemma \ref{lem:hitct}), and has relative hardness $1$ for general DNF (Lemma \ref{lem:relhdct}), however (absolute) hardness for arbitrary DNF-inputs can be arbitrarily high as shown in Lemma \ref{lem:exphddnf}. On the other hand, the reduced translation yields always $\Urefc$ (Lemma \ref{lem:hdctm}). So we have the following explanations why using either both directions or only one direction in the Tseitin translation, in the context of translating DNF's, can perform better than the other form:
\begin{itemize}
\item When using both directions (i.e., the canonical translation), splitting on the auxiliary variables is powerful, which is an advantage over using only one direction (i.e., the reduced canonical translation), where setting an auxiliary variable to false says nothing.
\item On the other hand, the canonical translation, when applied to non-hitting DNFs, can create hard unsatisfiable sub-problems (via partial assignments), which can not happen for the reduced translation.
\end{itemize}
It seems very interesting to us to turn these arguments into theorems (for concrete examples), and also to experimentally evaluate them. In this way we hope that in the future more precise directions can be given when to use which form of the Tseitin translation.

In Subsection \ref{sec:xorclauses} we turn to the translation of ``XOR-clauses''. Section 1.5 of \cite{GwynneKullmann2012Slur,GwynneKullmann2012SlurJ} discusses the translation of the so-called ``Schaefer classes'' into the $\Urefc_k$ hierarchy; see Section 12.2 in \cite{DH09HBSAT} for an introduction, and see \cite{CreignouKolaitisVollmer2008ComplexityConstraints} for an in-depth overview on recent developments. All Schaefer classes except affine equations have natural translations into either $\Urefc_1$ or $\Urefc_2$. The open question is whether systems of XOR-clauses (i.e., affine equations) can be translated into $\Urefc_k$ for some fixed $k$. We consider the most basic questions in a sense. On the positive side, for single XOR clauses, we show in Lemma \ref{lem:1softxor} that the Tseitin translation of a typical XOR summation circuit is in $\Urefc$. On the negative side, in Theorem \ref{thm:2xor}, we show for all $k \ge 3$ that applying this translation piecewise to systems of just two large-enough XOR clauses yields a SAT translation not in $\Urefc_k$. Conjecture \ref{con:xorcls} then hypothesises that, in general, systems of XOR-clauses have no representation of bounded hardness.

\subsection{Experimental results}
\label{sec:introexp}

In Section \ref{sec:experiments} we consider the usage of the class of boolean functions $f$ used for the lower bound as a constraint in a general SAT problem. We have equivalent clause-sets in $\Urefc_k$ for the optimal $k$, as well as short orthogonal DNF representations, which enable us to apply the canonical translation as well as the reduced canonical translation. We complement these three constraint-representations in a fixed way to obtain an unsatisfiable clause-set. The experiments show that for all solver types the optimal $\Urefc_k$ representations performs much better in terms of running time. This yields some evidence to our claim that equivalent representations in $\Urefc_k$ even for higher $k$ (in our experiments we considered $k \le 5$) might outperform representations obtained by introducing new variables, due to using (possibly) much less variables and clauses.

\subsection{Remarks on the term ``hardness''}
\label{sec:remarktermhd}

In general, if one speaks of the ``hardness measure'' $\hardness: \Cls \ra \NNZ$ (Definition \ref{def:hardness}) in context with other measures, then one should call it more specifically \emph{tree-hardness} (``t-hardness''), denoted by $\thardness(F)$, due to its close relation to tree-resolution (and its space complexity). So we have three basic types of hardness-measures, namely t-hardness $\thardness(F)$, the minimum $k$ with $F \in \Urefc_k$, p-hardness $\phardness(F)$, the minimum $k$ with $F \in \Propc_k$, and w-hardness, the minimum $k$ with in $F \in \Wrefc_k$. In this article, since $\thardness(F)$ is still most important here, we denote it by $\hardness(F) = \thardness(F)$.

In what respect is the terminology ``hardness'' appropriate? The hardness measure $\hardness(F)$ has been introduced in \cite{Ku99b,Ku00g}, based on quasi-automatisation of tree-resolution, that is, on a specific algorithmic approach (close to St\r{a}lmarcks approach).\footnote{Using the simplest oracle, on unsatisfiable instances the measure from \cite{Ku99b,Ku00g} yields $\hardness(F)$. But on satisfiable instances the approach of \cite{Ku99b,Ku00g} is very different, namely an algorithmic polynomial-time approach is taken, extending the breadth-first search for tree-resolution refutations in a natural way.} In \cite{AnsoteguiBonetLevyManya2008Hardness}, $\hardness(F)$ for unsatisfiable $F$ was proposed as measure of SAT-solver-hardness in general. This was criticised in \cite{JarvisaloMatsliahNordstromZivny2012ProofCompHardnessSAT} by the argument, that conflict-driven SAT solvers would be closer to dag-resolution (full resolution) than tree-resolution. Due to their heuristical nature, it seems to us that there is no robust measure of SAT-solver-hardness. Instead, our three basic measures, which are robust and mathematically meaningful, measure how good a clause-set $F$ is in representing an underlying boolean function in the following sense:
\begin{itemize}
\item Regarding instantiation we take a worst-case approach, that is, we consider \emph{all} partial assignments $\vp$ and their applications $\vp * F$ (insofar they create unsatisfiability or forced literals).

  A SAT-solver only uses \emph{certain} partial assignments, and thus this worst-case approach is overkill. However when using $F$ in \emph{any} context, then it makes sense to consider all partial assignments.
\item Regarding algorithms, we take a breadth-first approach, that is, the smallest $k$ such that $\rk_k$ or $k$-resolution succeeds. For $k > 1$ a SAT-solver might not find these inferences. In Subsection \ref{sec:introrelSAT} we have discussed the issue of incorporating these reductions into SAT solving. From a theoretical point of view, the maximisation over all partial assignments needs to be complemented with a minimisation (over $k$) in order to yield something interesting.
\end{itemize}

\subsection{Overview on results}

The preliminaries (Section \ref{sec:prelim}) define the basic notions. The classes $\Urefc_k$ and $\Wrefc_k$ are introduced in Section \ref{sec:measurerepcomp}. In Section \ref{sec:minpsdopingcls} we investigate minimal premise sets and doping in general, while in Section \ref{sec:doptreecls} we apply these notions to our source of hard examples. In Section \ref{sec:lowerb} we are then able to show the separation result. In Section \ref{sec:analT} we then turn from lower bounds to upper bounds, and analyse the Tseitin translation. To investigate the hardness of a special case, we present some tools for determining (w-)hardness in Section \ref{sec:hdunion}. Section \ref{sec:experiments} discusses our basic experiments. Finally, in Section \ref{sec:open} one finds many open problems.

The main results on minimal premise sets and doping are:
\begin{enumerate}
\item Theorem \ref{thm:dopedmpsprc} shows the correlation between prime implicates of doped clause-sets and minimal premise-sets of the original (undoped) clause-sets.
\item Theorem \ref{thm:dopedsmumax} characterises unsatisfiable clause-sets where every non-empty sub-clause-set is a minimal premise set.
\item Theorem \ref{thm:sumdsmuo} gives basic characteristics of doped $\Smusati{\delta=1}$-clause-sets.
\end{enumerate}

The main results on size lower bounds for the hardness are:
\begin{enumerate}
\item Theorem \ref{thm:triggersetmethod} introduces the basic method for lower bounding the size of equivalent clause-sets of a given hardness, via the transversal number of ``trigger hypergraphs''.
\item Theorem \ref{thm:nogoodksoft} shows a lower bound on the matching number of the trigger hypergraph of doped ``extremal'' $\Smusati{\delta=1}$-clause-sets.
\item Theorem \ref{thm:separation} shows that for every $k \in \NNZ$ there are polysize clause-sets in $\Urefc_{k+1}$, where every equivalent clause-set in $\Wrefc_k$ is of exponential size.
\end{enumerate}

And regarding upper bounds, that is, short representations (with new variables) with low hardness, we have the following main results:
\begin{enumerate}
\item Lemmas \ref{lem:hitct}, \ref{lem:hdctm} show how the canonical translation can yield results in $\Urefc$.
\item Theorem \ref{thm:extuc} shows that all doped $\Smusati{\delta=1}$-clause-sets (and in fact all doped unsatisfiable hitting clause-sets) have short CNF-representations in $\Urefc$ via the canonical translation.
\item Lemma \ref{lem:1softxor} shows that translating a single XOR-clause to $\Urefc$ is easy, while Theorem \ref{thm:2xor} shows that applying this translation to just two XOR-clauses already yields high hardness.
\end{enumerate}

\section{Preliminaries}
\label{sec:prelim}

We follow the general notations and definitions as outlined in \cite{Kullmann2007HandbuchMU}. We use $\NN = \set{1,2,\dots}$, $\NNZ = \NN \cup \set{0}$, and $\pot(M)$ for the set of subsets of set $M$.

\subsection{Clause-sets}
\label{sec:prelimcls}

Let $\Va$ be the infinite set of variables, and let $\Lit = \Va \cup \set{\ol{v} : v \in \Va}$ be the set of literals, the disjoint union of variables as positive literals and complemented variables as negative literals. We use $\ol{L} := \set{\ol{x} : x \in L}$ to complement a set $L$ of literals. A clause is a finite subset $C \subset \Lit$ which is complement-free, i.e., $C \cap \ol{C} = \es$; the set of all clauses is denoted by $\Cl$. A clause-set is a finite set of clauses, the set of all clause-sets is $\Cls$. By $\var(x) \in \Va$ we denote the underlying variable of a literal $x \in \Lit$, and we extend this via $\var(C) := \set{\var(x) : x \in C} \subset \Va$ for clauses $C$, and via $\var(F) := \bc_{C \in F} \var(C)$ for clause-sets $F$. The possible literals in a clause-set $F$ are denoted by $\lit(F) := \var(F) \cup \ol{\var(F)}$. Measuring clause-sets happens by $n(F) := \abs{\var(F)}$ for the number of variables, $c(F) := \abs{F}$ for the number of clauses, and $\ell(F) := \sum_{C \in F} \abs{C}$ for the number of literal occurrences. A special clause-set is $\top := \es \in \Cls$, the empty clause-set, and a special clause is $\bot := \es \in \Cl$, the empty clause.

A partial assignment is a map $\vp: V \ra \set{0,1}$ for some finite $V \subset \Va$, where we set $\var(\vp) := V$, and where the set of all partial assignments is $\Pass$. For $v \in \var(\vp)$ let $\vp(\ol{v}) := \ol{\vp(v)}$ (with $\ol{0} = 1$ and $\ol{1} = 0$). We construct partial assignments by terms $\pab{x_1 \ra \ve_1, \dots, x_n \ra \ve_n} \in \Pass$ for literals $x_1, \dots, x_n$ with different underlying variables and $\ve_i \in \set{0,1}$. We use $\vp_C := \pab{x \ra 0 : x \in C}$ for the partial assignment setting precisely the literals in clause $C \in \Cl$ to false.

For $\vp \in \Pass$ and $F \in \Cls$ we denote the result of applying $\vp$ to $F$ by $\vp * F$, removing clauses $C \in F$ containing $x \in C$ with $\vp(x) = 1$, and removing literals $x$ with $\vp(x) = 0$ from the remaining clauses. By $\Sat := \set{F \in \Cls \mb \ex\, \vp \in \Pass : \vp * F = \top}$ the set of satisfiable clause-sets is denoted, and by $\Usat := \Cls \sm \Sat$ the set of unsatisfiable clause-sets.

So clausal entailment, that is the relation $F \models C$ for $F \in \Cls$ and $C \in \Cl$, which by definition holds true iff for all $\vp \in \Pass$ with $\vp * F = \top$ we have $\vp * \set{C} = \top$, is equivalent to $\vp_C * F \in \Usat$.

Two clauses $C, D \in \Cl$ are resolvable iff they clash in exactly one literal $x$, that is, $C \cap \ol{D} = \set{x}$, in which case their resolvent is $\bmm{C \res D} := (C \cup D) \sm \set{x,\ol{x}}$ (with resolution literal $x$). A resolution tree is a full binary tree formed by the resolution operation. We write \bmm{T : F \vdash C} if $T$ is a resolution tree with axioms (the clauses at the leaves) all in $F$ and with derived clause (at the root) $C$. A resolution tree $T : F \vdash C$ is regular iff along each path from the root of $T$ to a leaf no resolution-variable is used more than once. In this article we use only resolution \emph{trees}, even when speaking of unrestricted resolution, that is, we always unfold dag-resolution proofs to (full) binary resolution trees. Completeness of resolution means that $F \models C$ (semantic implication) is equivalent to $F \vdash C$, i.e., there is some $C' \sse C$ and some $T$ with $T: F \vdash C'$.

A \emph{prime implicate} of $F \in \Cls$ is a clause $C$ such that a resolution tree $T$ with $T: F \vdash C$ exists, but no $T'$ exists for some $C' \subset C$ with $T': F \vdash C'$; the set of all prime implicates of $F$ is denoted by $\bmm{\primec_0(F)} \in \Cls$. The term ``implicate'' refers to the implicit interpretation of $F$ as a conjunctive normal form (CNF). Considering clauses as combinatorial objects one can speak of ``prime clauses'', and the ``$0$'' in our notation reminds of ``unsatisfiability'', which is characteristic for CNF. Two clause-sets $F, F' \in \Cls$ are equivalent iff $\primec_0(F) = \primec_0(F')$. A clause-set $F$ is unsatisfiable iff $\primec_0(F) = \set{\bot}$. The set of \emph{prime implicants} of a clause-set $F \in \Cls$ is denoted by $\bmm{\primec_1(F)} \in \Cls$, and is the set of all clauses $C \in \Cl$ such that for all $D \in F$ we have $C \cap D \not= \es$, while this holds for no strict subset of $C$.

\subsection{CNF versus DNF}
\label{sec:cnfvdnf}

As we said, the default interpretation of a clause-set $F$ is as a CNF, which we can emphasise by speaking of the ``CNF-clause-set $F$'', that is, the interpretation as a boolean function is
\begin{displaymath}
  F \leadsto \bw_{C \in F} \bv_{x \in C} x.
\end{displaymath}
We might consider $F$ also as a DNF-clause-set, which does not change $F$ itself, but only changes the interpretation of $F$ in considerations regarding the semantics:
\begin{displaymath}
  F \leadsto \bv_{C \in F} \bw_{x \in C} x.
\end{displaymath}
Note that by the de Morgan rules from the CNF-formula we obtain the DNF-formula via negating the whole formula together with negating the literals (in other words, the underlying boolean function of a CNF-clause-set $F$ is the ``dual'' of the underlying boolean function of the DNF-clause-set $F$; see \cite{CramaHammer2011BooleanFunctions}). Thus the logical negation (as CNF) of a clause-set $F$ (as CNF) is obtained from a DNF-clause-set equivalent to $F$ by negating all literals.

\begin{examp}\label{exp:dnfvcnf}
  The clause-set $F = \set{\set{v}}$ has the equivalent DNF-clause-set $F = \set{\set{v}}$ (the underlying boolean function is ``self-dual''; see \cite{CramaHammer2011BooleanFunctions}), while the negation is $\set{\set{\ol{v}}}$. And $F = \set{\set{v,w}}$ has the equivalent DNF-clause-set $\set{\set{v},\set{w}}$, while the negation is $\set{\set{\ol{v}},\set{\ol{w}}}$.
\end{examp}

The above description of the sets $\primec_0(F), \primec_1(F)$ as the set of prime implicates resp.\ implicants holds for the default interpretation of $F$ as CNF, while for the DNF-interpretation $\primec_0(F)$ becomes the set of prime implicants, while $\primec_1(F)$ becomes the set of prime implicates (of the boolean function underlying $F$). A CNF-clause-set $F$ is equivalent to a DNF-clause-set $G$ iff $\primec_0(F) = \primec_1(G)$. The total satisfying assignments of a (CNF-)clause-set $F$ can be identified with the elements of the canonical DNF of $F$, which is defined via the map $\Dnf: \Cls \ra \Cls$, where for $F \in \Cls$ we set $\Dnf(F) := \set{C \in \Cl \mb \var(C) = \var(F) \und \fa\, D \in F : C \cap D \not= \es}$.

While clause-sets and partial assignments themselves are neutral regarding CNF- or DNF-interpretation, the application $\vp * F$ is based on the CNF-interpretation of $F$; if we wish to use the DNF-interpretation of $F$, then we use $\ol{\vp} * F$, where $\ol{\vp} := \pab{v \ra \ol{\vp(v)} : v \in \var(\vp)}$. While $\top$ in the CNF-interpretation stands for ``true'', in the DNF-interpretation it becomes ``false''.

\begin{examp}\label{exp:CNFDNF}
  Consider $F := \set{\set{a},\set{b}} \in \Cls$ (with $n(F) = c(F) = \ell(F) = 2$). Then $\Dnf(F) = \set{\set{a,b}}$, and for $\vp := \pab {a, b \ra 1}$ we have $\vp * F = \top$. This corresponds to the CNF-interpretation $a \und b$ of $F$, which has exactly one satisfying assignment $\vp$. If we consider the DNF-interpretation $a \oder b$ of $F$, then we have three satisfying total assignments for the DNF-clause-set $F$, and for example the satisfying assignment $\psi := \pao a1$ is recognised via $\ol{\psi} * F = \pao a0 * F = \set{\bot, \set{b}}$, where the result as DNF is a tautology, since $\bot$ as a DNF-clause becomes the constant $1$ (as the empty conjunction).
\end{examp}

\subsection{On ``good'' equivalent clause-sets}
\label{sec:prelimgoodeqcls}

A basic problem considered in this article is for a given $F \in \Cls$ to find a ``good'' equivalent $F' \in \Cls$. How ``good'' $F'$ is depends in our context on two factors, which have to be balanced against each other:
\begin{itemize}
\item the size of $F'$: we measure $c(F')$, and the smaller the better;
\item the inference power of $F'$: inference from $F'$ should be ``as easy as possible'', and we consider two measures in this article, (tree-)hardness in Subsection \ref{sec:prelimhdUC}, and width-hardness in Subsection \ref{sec:prelimwhdWC}; the smaller these measures, the easier inference w.r.t.\ tree resolution resp.\ (generalised) width-bounded resolution.
\end{itemize}
The basic size-lower-bound for $F'$ is given by the \textbf{essential prime implicates}, which are those $C \in \primec_0(F)$ such that $\primec_0(F) \sm \set{C}$ is not equivalent to $F$:
\begin{lem}\label{lem:necprcls}
  Consider $F \in \Cls$, and let $P \sse \primec_0(F)$ be the set of essential prime implicates of $F$. Now for every $F' \in \Cls$ equivalent to $F$ there exists an injection $i: P \ra F'$ such that for all $C \in P$ holds $C \sse i(C)$. Thus $c(F') \ge c(P)$.
\end{lem}
\begin{prf}
  For every $C' \in F'$ there exists a $C \in \primec_0(F)$ such that $C \sse C'$; replacing every $C' \in F$ by such a chosen $C$ we obtain $F'' \sse \primec_0(F)$ with $P \sse F''$. \Qed
\end{prf}

Note that Lemma \ref{lem:necprcls} crucially depends on not allowing new variables (see Subsection \ref{sec:cnfrep} for what it means that an $F'$ with new variable ``represents'' $F$) --- when allowing new variable, then we currently do not have any overview on the possibilities for ``better'' $F'$. The most powerful representation regarding inference alone (with or without new variables) is given by the set $\primec_0(F)$ of all prime implicates of $F$, and will have ``hardness'' $0$, as defined in the following section. (The problem is of course that in most cases this representation is too large, and thus higher hardness must be allowed.)

\section{Measuring ``SAT representation complexity''}
\label{sec:measurerepcomp}

In this section we define and discuss the measures $\hardness, \phardness, \whardness: \Cls \ra \NNZ$ and the corresponding classes $\Urefc_k, \Propc_k, \Wrefc_k \subset \Cls$. It is mostly of an expository nature, explaining what we need from \cite{Ku99b,Ku00g,GwynneKullmann2012SlurSOFSEM,GwynneKullmann2012Slur,GwynneKullmann2012SlurJ}, with some additional remarks.

\subsection{Hardness and $\Urefc_k$}
\label{sec:prelimhdUC}

First we turn to the most basic hardness measurement. It can be based on resolution refutation trees, as we do here, but it can also be defined algorithmically, via generalised unit-clause propagation (see Lemma \ref{lem:charachd}).
\begin{defi}\label{def:hthts}
  For a full binary tree $T$ the height $\bmm{\height(T)} \in \NNZ$ and the Horton-Strahler number $\bmm{\hts(T)} \in \NNZ$ are defined as follows:
  \begin{enumerate}
  \item If $T$ is trivial (i.e., $\nnds(T) = 1$), then $\height(T) := 0$ and $\hts(T) := 0$.
  \item Otherwise let $T_1, T_2$ be the two subtrees of $T$:
    \begin{enumerate}
    \item $\height(T) := 1 + \max(\height(T_1), \height(T_2))$
    \item If $\hts(T_1) = \hts(T_2)$, then $\hts(T) := 1 + \max(\hts(T_1), \hts(T_2))$, otherwise $\hts(T) := \max(\hts(T_1), \hts(T_2))$.
    \end{enumerate}
  \end{enumerate}
\end{defi}
Obviously we always have $\hts(T) \le \height(T)$.
\begin{examp}\label{exp:hts}
  For the tree $T$ from Example \ref{exp:dopedsmutree} we have $\height(T) = 3$, $\hts(T) = 2$. The Horton-Strahler numbers of the subtrees are as follows:
    \begin{displaymath}
    \xygraph{
      !{0;/r8ex/:}
        []{2} (
          - [dll]{2} (
            -[dll]{1} (
              -[dl]{0} (),
              -[dr]{0} ()
            ),
            -[drr]{1} (
              -[dl]{0} (),
              -[dr]{0} ()
            )
          ),
          - [drr]{1} (
            -[dl]{0} (),
            -[dr]{0} ()
          )
        )
    }
  \end{displaymath}
\end{examp}

\begin{defi}\label{def:hardness}
  The hardness $\hardness: \Cls \ra \NNZ$ is defined for $F \in \Cls$ as follows:
  \begin{enumerate}
  \item If $F \in \Usat$, then $\hardness(F)$ is the minimum $\hts(T)$ for $T : F \vdash \bot$.
  \item If $F = \top$, then $\hardness(F) := 0$.
  \item If $F \in \Sat \sm \set{\top}$, then $\hardness(F) := \max_{\vp \in \Pass} \set{\hardness(\vp * F) : \vp * F \in \Usat}$.
  \end{enumerate}
\end{defi}

Hardness for unsatisfiable clause-sets was introduced in \cite{Ku99b,Ku00g}, while this generalisation to arbitrary clause-sets was first mentioned in \cite{AnsoteguiBonetLevyManya2008Hardness}, and systematically studied in \cite{GwynneKullmann2012SlurSOFSEM,GwynneKullmann2012Slur,GwynneKullmann2012SlurJ}. Definition \ref{def:hardness} defines hardness proof-theoretically; importantly, it can also be characterised algorithmically via necessary levels of generalised unit-clause propagation (see \cite{GwynneKullmann2012SlurSOFSEM,GwynneKullmann2012Slur,GwynneKullmann2012SlurJ} for the details):
\begin{lem}\label{lem:charachd}
  Consider the reductions $\rk_k: \Cls \ra \Cls$ for $k \in \NNZ$ as introduced in \cite{Ku99b}; it is $\rk_1$ unit-clause propagation, while $\rk_2$ is (full, iterated) failed-literal elimination. Then $\hardness(F)$ for $F \in \Cls$ is the minimal $k \in \NNZ$ such that for all $\vp \in \Pass$ with $\vp * F \in \Usat$ holds $\rk_k(\vp * F) = \set{\bot}$, i.e., the minimal $k$ such that $\rk_k$ detects unsatisfiability of any instantiation.
\end{lem}

We can now define our main hierarchy, the $\Urefc_k$-hierarchy (with ``UC'' for ``unit-refutation complete'') via (tree-)hardness:
\begin{defi}\label{def:UC}
  For $k \in \NNZ$ let $\bmm{\Urefc_k} := \set{F \in \Cls : \hardness(F) \le k}$.
\end{defi}
$\Urefc_1 = \Urefc$ is the class of unit-refutation complete clause-sets, as introduced in \cite{Val1994UnitResolutionComplete}. In \cite{GwynneKullmann2012SlurSOFSEM,GwynneKullmann2012Slur,GwynneKullmann2012SlurJ} we show that $\Urefc = \Slur$, where $\Slur$ is the class of clause-sets solvable via Single Lookahead Unit Resolution (see \cite{FrGe98}). Using \cite{CepekKuceraVlcek2012SLUR} we then obtain (\cite{GwynneKullmann2012SlurSOFSEM,GwynneKullmann2012Slur,GwynneKullmann2012SlurJ}) that membership decision for $\Urefc_k$ ($ = \Slur_k$) is coNP-complete for $k \ge 1$. The class $\Urefc_2$ is the class of all clause-sets where unsatisfiability for any partial assignment is detected by failed-literal reduction (see Section 5.2.1 in \cite{HvM09HBSAT} for the usage of failed literals in SAT solvers).

A basic fact is that the classes $\Urefc_k$ are stable under application of partial assignments, in other words, for $F \in \Cls$ and $\vp \in \Pass$ we have $\hardness(\vp * F) \le \hardness(F)$. For showing lower bounds on the hardness for unsatisfiable clause-sets, we can use the methodology developed in Subsection 3.4.2 of \cite{Ku99b}. A simplified version of Lemma 3.17 from \cite{Ku99b}, sufficient for our purposes, is as follows (with a technical correction, as explained in Example \ref{exp:lbhd}):
\begin{lem}\label{lem:lbhd}
  Consider $\mc{C} \sse \Usat$ and a function $h: \mc{C} \ra \NNZ$. For $k \in \NNZ$ let $\mc{C}_k := \set{F \in \mc{C} : h(F) \ge k}$. Then $\fa\, F \in \mc{C} : \hardness(F) \ge h(F)$ holds if and only if $\Urefc_0 \cap \mc{C}_1 = \es$, and for all $k \in \NN$, $F \in \mc{C}_k$ and $x \in \lit(F)$ there exist clause-sets $F_0, F_1 \in \Cls$ fulfilling the following three conditions:
  \begin{enumerate}
  \item[(i)] $n(F_{\ve}) < n(F)$ for both $\ve \in \set{0,1}$;
  \item[(ii)] $\hardness(F_{\ve}) \le \hardness(\pao x{\ve} * F)$ for both $\ve \in \set{0,1}$;
  \item[(iii)] $F_0 \in \mc{C}_k$ or $F_1 \in \mc{C}_{k-1}$.
  \end{enumerate}
\end{lem}
\begin{prf}
The given conditions are necessary for $\fa\, F \in \mc{C} : \hardness(F) \ge h(F)$, since we can choose $F_{\ve} := \pao v{\ve} * F$ for $\ve \in \set{0,1}$. To see sufficiency, assume for the sake of contradiction that there is $F \in \mc{C}$ with $\hardness(F) < h(F)$, and consider such an $F$ with minimal $n(F)$. If $\hardness(F) = 0$, so $h(F) = 0$ by assumption, and thus $\hardness(F) \ge 1$ would hold. So assume $\hardness(F) \ge 1$. It follows that there is a literal $x \in \lit(F)$ with $\hardness(\pao x1 * F) < \hardness(F)$. Let $k := h(F)$; so $F \in \mc{C}_k$. By assumption there are $F_0, F_1 \in \Cls$ with $\hardness(F_{\ve}) \le \hardness(\pao x{\ve} * F)$ for both $\ve \in \set{0,1}$, and $F_0 \in \mc{C}_k$ or $F_1 \in \mc{C}_{k-1}$. If $F_0 \in \mc{C}_k$, then $\hardness(F_0) \le \hardness(F) < k \le h(F_0)$, while $n(F_0) < n(F)$, contradicting minimality of $F$. And if $F_1 \in \mc{C}_{k-1}$, then $\hardness(F_1) \le \hardness(F) - 1 < k -1 \le h(F_1)$, while $n(F_1) < n(F)$, contradicting again minimality of $F$. \Qed
\end{prf}

Lemma 3.17 in \cite{Ku99b} doesn't state the condition (i) from Lemma \ref{lem:lbhd}. The following example shows that this condition actually needs to be stated (that is, if we just have (ii) and (iii), then $h$ doesn't need to be a lower bound for $\hardness$); fortunately in all applications in \cite{Ku99b} this (natural) condition is fulfilled.
\begin{examp}\label{exp:lbhd}
  Consider $\mc{C} := \Urefc_1 \cap \Usat$. Define $h: \mc{C} \ra \set{0,1,2}$ as $h(F) = 0$ iff $\bot \in F$, and $h(F) = 1$ iff $\bot \notin F$ and there is $v \in \var(F)$ with $\set{v},\set{\ol{v}} \in F$. So we have $h(F) = 2$ if and only if for all literals $x \in \lit(F)$ holds $\hardness(\pao x1 * F) = \hardness(\pao x0 * F) = 1$. By definition we have $\Urefc_0 \cap \mc{C}_1 = \es$. Now consider $k \in \set{1,2}$, $F \in \mc{C}_k$ and $x \in \lit(F)$. If $h(F) = 1$, then let $F_{\ve} := \pao x{\ve} * F$, while otherwise $F_{\ve} := F$ for $\ve \in \set{0,1}$. Now Conditions (ii), (iii) of Lemma \ref{lem:lbhd} are fulfilled (if $h(F) = 1$, then for Condition (iii) always $F_1 \in \mc{C}_{k-1}$ holds, while in case of $h(F) = 2$ we always have $F_0 \in \mc{C}_k$). But by definition $h$ is not a lower bound on $\hardness$.
\end{examp}

Complementary to ``unit-refutation completeness'', there is the notion of ``pro\-pa\-ga\-tion-com\-ple\-te\-ness'' as investigated in \cite{DarwichePipatsrisawat2011ClauseLearnRes,BordeauxMarquesSilva2012KnowledgeCompilation}, yielding the class $\Propc \subset \Urefc$. This was captured and generalised by a measure $\phardness: \Cls \ra \NNZ$ of ``propagation-hardness'' along with the associated hierarchy, defined in \cite{GwynneKullmann2012Slur,GwynneKullmann2012SlurJ} as follows:
\begin{defi}\label{def:phardness}
  For $F \in \Cls$ we define the \textbf{propagation-hardness} (for short ``p-hardness'') $\bmm{\phardness(F)} \in \NNZ$ as the minimal $k \in \NNZ$ such that for all partial assignments $\vp \in \Pass$ we have $\rk_k(\vp * F) = \rki(\vp * F)$, where $\rk_k: \Cls \ra \Cls$ is generalised UCP (\cite{Ku99b,Ku00g}), and $\rki: \Cls \ra \Cls$ applies all forced assignments, and can be defined by $\rki(F) := \rk_{n(F)}(F)$. For $k \in \NNZ$ let $\bmm{\Propc_k} := \set{F \in \Cls : \phardness(F) \le k}$ (the class of \textbf{propagation-complete clause-sets of level $k$}).
\end{defi}
We have $\Propc = \Propc_1$. For $k \in \NNZ$ we have $\Propc_k \subset \Urefc_k \subset \Propc_{k+1}$.

\subsection{W-Hardness and $\Wrefc_k$}
\label{sec:prelimwhdWC}

A basic weakness of the standard notion of width-restricted resolution, which demands that \emph{both} parent clauses must have length at most $k$ for some fixed $k \in \NNZ$ (the ``width''; see \cite{SW98}), is that even Horn clause-sets require unbounded width in this sense. The correct solution, as investigated and discussed in \cite{Ku99b,Ku00g}, is to use the notion of ``$k$-resolution'' as introduced in \cite{Kl93}, where only \emph{one} parent clause needs to have length at most $k$ (thus properly generalising unit-resolution). Nested input-resolution (\cite{Ku99b,Ku00g}) is the proof-theoretic basis of hardness, and approximates tree-resolution. In the same vein, $k$-resolution is the proof-theoretic basis of ``w-hardness'', and approximates dag-resolution (see Theorem 6.12 in \cite{Ku00g}):
\begin{defi}\label{def:whd}
  The w-hardness $\whardness: \Cls \ra \NNZ$ (``width-hardness'') is defined for $F \in \Cls$ as follows:
  \begin{enumerate}
  \item If $F \in \Usat$, then $\whardness(F)$ is the minimum $k \in \NNZ$ such that $k$-resolution refutes $F$, that is, such that $T : F \vdash \bot$ exists where for each resolution step $R = C \res D$ in $T$ we have $\abs{C} \le k$ or $\abs{D} \le k$ (this corresponds to Definition 8.2 in \cite{Ku99b}, and is a special case of $\mr{wid}_{\mc{U}}$ introduced in Subsection 6.1 of \cite{Ku00g}).
  \item If $F = \top$, then $\whardness(F) := 0$.
  \item If $F \in \Sat \sm \set{\top}$, then $\DST \whardness(F) := \max_{\vp \in \Pass} \set{\whardness(\vp * F) : \vp * F \in \Usat}$.
  \end{enumerate}
  For $k \in \NNZ$ let $\bmm{\Wrefc_k} := \set{F \in \Cls : \whardness(F) \le k}$.
\end{defi}
We have $\Wrefc_0 = \Urefc_0$, $\Wrefc_1 = \Urefc_1$, and for all $k \in \NNZ$ holds $\Urefc_k \sse \Wrefc_k$ (this follows by Lemma 6.8 in \cite{Ku00g} for unsatisfiable clause-sets, which extends to satisfiable clause-sets by definition). For unsatisfiable $F$, whether $\whardness(F) = k$ holds for $k \in \set{0,1,2}$ can be decided in polynomial time; this is non-trivial for $k = 2$ (\cite{BuroKleineBuening1996ResolutionShortClauses}) and unknown for $k > 2$. Nevertheless, the clausal entailment problem $F \models C$ for $F \in \Wrefc_k$ and fixed $k \in \NNZ$ is decidable in polynomial time, as shown in Subsection 6.5 of \cite{Ku00g}, by actually using a slight strengthening of $k$-resolution, which combines width-bounded resolution and input resolution. While space-complexity of the decision $F \models C$ for $F \in \Urefc_k$ is linear (for fixed $k$), now for $\Wrefc_k$ space-complexity is $O(\ell(F) \cdot n(F)^{O(k)})$.

As a special case of Theorem 6.12 in \cite{Ku00g} we obtain for $F \in \Usat$, $n(F) \not= 0$, the following general lower bound on resolution complexity:
\begin{displaymath}
  \compr(F) > b^{\frac{\whardness(F)^2}{n(F)}},
\end{displaymath}
where $b := e^{\frac 18} = 1.1331484 \ldots$, while $\compr(F) \in \NN$ is the minimal number of different clauses in a (tree-)resolution refutation of $F$. Similar to Theorem 14 in \cite{GwynneKullmann2012SlurSOFSEM} resp.\ Theorem 5.7 in \cite{GwynneKullmann2012Slur,GwynneKullmann2012SlurJ} we thus obtain:
\begin{lem}\label{lem:suffcritwhd}
  For $F \in \Cls$ and $k \in \NNZ$, such that for every $C \in \primec_0(F)$ with $\abs{C} < n(F)$ there exists a resolution proof of $C$ from $F$ using at most $b^{\frac{(k+1)^2}{n(F)-\abs{C}}}$ different clauses, we have $\whardness(F) \le k$.
\end{lem}

\section{Minimal premise sets and doped clause-sets}
\label{sec:minpsdopingcls}

In this section we study ``minimal premise sets'', ``mps's'' for short, introduced in \cite{Kullmann2007ClausalFormZII}, together with the properties of ``doped'' clause-sets, generalising a construction used in \cite{SloanSzoerenyiTuran2005Primimplikanten_1}. Mps's are generalisations of minimally unsatisfiable clause-sets stronger than irredundant clause-sets, while doping relates prime implicates and sub-mps's.

Recall that a clause-set $F$ is minimally unsatisfiable if $F \in \Usat$, while for all $C \in F$ holds $F \sm \set{C} \in \Sat$. The set of all minimally unsatisfiable clause-sets is $\bmm{\Musat} \subset \Cls$; see \cite{Kullmann2007HandbuchMU} for more information. In other words, for $F \in \Cls$ we have $F \in \Musat$ if and only if $F \models \bot$ and $F$ is minimal regarding this entailment relation. Now an mps is a clause-set $F$ which minimally implies some clause $C$, i.e., $F \models C$, while $F' \not\models C$ for all $F' \subset F$. In Subsection \ref{sec:minimalprems} we study the basic properties of mps's $F$, and determine the unique minimal clause implied by $F$ as $\purec(F)$, the set of pure literals of $F$.

For a clause-set $F$ its doped version $\doping(F) \in \Cls$ receives an additional new (``doping'') variable for each clause. The basic properties are studied in Subsection \ref{sec:Dopingcls}, and in Theorem \ref{thm:dopedmpsprc} we show that the prime implicates of $\doping(F)$ correspond 1-1 to the mps's contained in $F$. In Subsection \ref{sec:hddopedcls} we determine the hardness of doped clause-sets.

\subsection{Minimal premise sets}
\label{sec:minimalprems}

In Section 4.1 in \cite{Kullmann2007ClausalFormZII} basic properties of \emph{minimal premise sets} are considered:
\begin{defi}\label{def:Mps}
  A clause-set $F \in \Cls$ is a \textbf{minimal premise set} (``mps'') \textbf{for a clause $C \in \Cl$} if $F \models C$ and $\fa\, F' \subset F : F' \not\models C$, while $F$ is a \textbf{minimal premise set} if there exists a clause $C$ such that $F$ is a minimal premise set for $C$. The set of all minimal premise (clause-)sets is denoted by \bmm{\Mps}.
\end{defi}
Remarks:
\begin{enumerate}
\item $\top$ is not an mps (since no clause follows from $\top$).
\item An unsatisfiable clause-set is an mps iff it is minimally unsatisfiable, i.e., $\Mps \cap \Usat = \Musat$. In Corollary \ref{cor:mpswp} we will see that the minimally unsatisfiable clause-sets are precisely the mps's without pure literals.
\item Every minimal premise clause-set is irredundant (no clause follows from the other clauses).
\item For a clause-set $F$ and any implicate $F \models C$ there exists a minimal premise sub-clause-set $F' \sse F$ for C.
\item A single clause $C$ yields an mps $\set{C}$.
\item Two clauses $C \not= D$ yield an mps $\set{C,D}$ iff $C, D$ are resolvable.
\item If $F_1, F_2 \in \Mps$ with $\var(F_1) \cap \var(F_2) = \es$, then $F_1 \cup F_2 \notin \Mps$ except in case of $F_1 = F_2 = \set{\bot}$.
\end{enumerate}

\begin{examp}\label{exp:Mps}
  $\set{\set{a},\set{b}}$ for variables $a \not= b$ is irredundant but not an mps.
\end{examp}

With Corollary 4.5 in \cite{Kullmann2007ClausalFormZII} we see that no clause-set can minimally entail more than one clause:
\begin{lem}\label{lem:purelmps}
  For $F \in \Mps$ there exists exactly one $C \in \primec_0(F)$ such that $C$ is a minimal premise set for $C$, and $C$ is the smallest element of the set of clauses for which $F$ is a minimal premise set.
\end{lem}

We remark that Lemma \ref{lem:purelmps} does not mean that $\abs{\primec_0(F)} = 1$ for $F \in \Mps$; indeed, $F$ can have many $F' \subset F$ with $F' \in \Mps$, and each such $F'$ might contribute a prime implicate, as we will see later. We wish now to determine that unique prime implicate $C$ which follows minimally from an mps $F$. It is clear that $C$ must contain all pure literals from $F$, since all clauses of $F$ must be used, and we can not get rid off pure literals.

\begin{defi}\label{def:epc}
  For $F \in \Cls$ the \textbf{pure clause of $F$}, denoted by $\bmm{\purec(F)} \in \Cl$, is the set of pure literals of $F$, that is, $\purec(F) := L \sm (L \cap \ol{L})$, where $L := \bc F$ is the set of literals occurring in $F$.
\end{defi}

\begin{examp}\label{exp:epc}
  For $F = \set{\set{a,b},{\set{\ol{a},\ol{c}}}}$ we have $\purec(F) = \set{b,\ol{c}}$.
\end{examp}

The main observation for determining $C$ is that the conclusion of a regular resolution proof consists precisely of the pure literals of the axioms (this follows by definition):
\begin{lem}\label{lem:regrespurec}
  For a regular resolution proof $T: F \vdash C$, where every clause of $F$ is used in $T$, we have $C = \purec(F)$.
\end{lem}

Due to the completeness of regular resolution we thus see, that $\purec(F)$ is the desired unique prime implicate:
\begin{lem}\label{lem:uniquepurec}
   For $F \in \Mps$ the unique prime implicate $C$, for which $F$ is a minimal premise set (see Lemma \ref{lem:purelmps}), is $C = \purec(F)$.
\end{lem}
\begin{prf}
  Consider a regular resolution proof $T: F \vdash C$ (recall that regular resolution is complete); due to $F \in \Mps$ every clause of $F$ must be used in $T$, and thus the assertion follows by Lemma \ref{lem:regrespurec}. \Qed
\end{prf}

\begin{corol}\label{cor:mpswp}
  If we have $F \in \Mps$ with $\purec(F) = \bot$, then $F \in \Musat$.
\end{corol}

By Lemma 4.4 in \cite{Kullmann2007ClausalFormZII} we get the main characterisation of mps's, namely that after elimination of pure literals they must be minimally unsatisfiable:
\begin{lem}\label{lem:characmps}
  Consider a clause-set $F \in \Cls$. Then $F \in \Mps$ if and only if the following two conditions hold for $\vp := \vp_{\purec(F)}$ (setting precisely the pure literals of $F$ to false):
  \begin{enumerate}
  \item $\vp * F \in \Musat$ (after removing the pure literals we obtain a minimal unsatisfiable clause-sets).
  \item $\vp$ is contraction-free for $F$, that is, for clauses $C, D \in F$ with $C \not= D$ we have $\vp * \set{C} \ne \vp * \set{D}$.
  \end{enumerate}
  These two conditions are equivalent to stating that $\vp * F$ as a multi-clause-set (not contracting equal clauses) is minimally unsatisfiable.
\end{lem}

Thus we obtain all mps's by considering some minimally unsatisfiable clause-sets and adding new variables in the form of pure literals:
\begin{corol}\label{cor:genmps}
  The following process generates precisely the $F' \in \Mps$:
  \begin{enumerate}
  \item Choose $F \in \Musat$.
  \item Choose a clause $P$ with $\var(P) \cap \var(F) = \es$ (``P'' like ``pure'').
  \item Choose a map $e: F \ra \pot(P)$ (``e'' like ``extension'').
  \item Let $F' := \set{C \cup e(C) : C \in F}$.
  \end{enumerate}
\end{corol}

For unsatisfiable clause-sets the set of minimally unsatisfiable sub-clause-sets has been studied extensively in the literature; see \cite{MarquesSilva2012MUS} for a recent overview. The set of subsets which are mps's strengthen this notion (now for all clause-sets):
\begin{defi}\label{def:mps}
  For a clause-set $F \in \Cls$ by $\bmm{\mps(F)} \subset \Cls$ the set of all minimal premise sub-clause-sets is denoted: $\mps(F) := \pot(F) \cap \Mps$.
\end{defi}
We have $\abs{\mps(F)} \le 2^{c(F)}-1$.\footnote{There is a typo in Corollary 4.6 of \cite{Kullmann2007ClausalFormZII}, misplacing the ``$-1$'' into the exponent.} The minimal elements of $\mps(F)$ are $\set{C} \in \mps(F)$ for $C \in F$. Since every prime implicate of a clause-set has some minimal premise sub-clause-set, we get that running through all sub-mps's in a clause-set $F$ and extracting the clauses with the pure literals we obtain at least all prime implicates:
\begin{lem}\label{lem:purecprimec}
  For $F \in \Cls$ the map $F' \in \mps(F) \mapsto \purec(F') \sse \set{C \in \Cl : F \models C}$ covers $\primec_0(F)$ (i.e., its range contains the prime implicates of $F$).
\end{lem}

\begin{examp}\label{exp:moremps}
  Examples where we have more minimal premise sub-clause-sets than prime implicates are given by $F \in \Musat$, where $\primec_0(F) = \set{\bot}$, while in the most extreme case every non-empty subset of $F$ can be a minimal premise sub-clause-set (see Theorem \ref{thm:dopedsmumax}).
\end{examp}

\subsection{Doping clause-sets}
\label{sec:Dopingcls}

``Doping'' is the process of adding a unique new variable to every clause of a clause-set. It enables us to follow the usage of this clause in derivations:
\begin{defi}\label{def:doping}
  For every clause-set $F \in \Cls$ we assume an injection $u^F: F \ra \Va \sm \var(F)$ in the following, assigning to every clause $C$ a different variable $u^F_C$. For a clause $C \in \Cl$ and a clause-set $F \in \Cls$ we then define the \textbf{doping} $\bmm{\doping_F(C)} := C \cup \set{u^F_C} \in \Cl$, while $\bmm{\doping(F)} := \set{\doping_F(C) : C \in F} \in \Cls$.
\end{defi}
Remarks:
\begin{enumerate}
\item In the following we drop the upper index in ``$u^F_C$'', i.e., we just use ``$u_C$''.
\item We have $\doping: \Cls \ra \Sat$.
\item For $F \in \Cls$ we have $n(\doping(F)) = n(F) + c(F)$ and $c(\doping(F)) = c(F)$.
\item For $F \in \Cls$ we have $\purec(\doping(F)) = \purec(F) \cup \set{u_C : C \in F}$.
\end{enumerate}
We are interested in the prime implicates of doped clause-sets. It is easy to see that all doped clauses are themselves essential prime implicates:
\begin{lem}\label{lem:necprcdp}
  For $F \in \Cls$ we have $\doping(F) \sse \primec_0(\doping(F))$, and furthermore all elements of $\doping(F)$ are essential prime implicates.
\end{lem}
\begin{prf}
  Every resolvent of clauses from $\doping(F)$ contains at least two doping variables, and thus the clauses of $\doping(F)$ themselves (which contain only one doping variable) are prime and necessary. \Qed
\end{prf}
Thus by Lemma \ref{lem:necprcls} among all the clause-sets equivalent to $\doping(F)$ this clause-set itself is the smallest. Directly by Lemma \ref{lem:characmps} we get that a clause-set is an mps iff its doped form is an mps:
\begin{lem}\label{lem:mpsdoping}
  For $F \in \Cls$ holds $F \in \Mps \Lra \doping(F) \in \Mps$. Thus the map $F' \in \mps(F) \mapsto \doping(F')$ is a bijection from $\mps(F)$ to $\mps(\doping(F))$.
\end{lem}

For doped clause-sets the surjection of Lemma \ref{lem:purecprimec} is bijective:
\begin{lem}\label{lem:purecprimecdoped}
  Consider a clause-set $F \in \Cls$, and let $G := \doping(F)$.
  \begin{enumerate}
  \item\label{lem:purecprimecdoped1} The map $F' \in \mps(G) \mapsto \purec(F') \in \Cl$ is a bijection from $\mps(G)$ to $\primec_0(G)$.
  \item\label{lem:purecprimecdoped2} The inverse map from $\primec_0(G)$ to $\mps(G)$ obtains from $C \in \primec_0(G)$ the clause-set $F' \in \mps(G)$ with $\purec(F') = C$ as $F' = \set{\doping(D) : D \in F \und u_D \in \var(C)}$.
  \end{enumerate}
\end{lem}
\begin{prf}
  By  Lemma \ref{lem:purecprimec} it remains to show that the map of Part \ref{lem:purecprimecdoped1} is injective and does not have subsumptions in the image. Assume for the sake of contradiction there are $G', G'' \in \mps(G)$, $G' \not= G''$, with $\purec(G') \sse \purec(G'')$. Since every clause of $F$ has a different doping-variable, $G' \subset G''$ must hold. Consider the $F', F'' \in \mps(F)$ with $\doping(F')= G'$ and $\doping(F'') = G''$. We have $F' \subset F''$, and thus $\purec(F') \not\sse \purec(F'')$, since for every $F \in \Mps$ the clause $\purec(F)$ is a prime implicate of $F$. It follows that $\purec(G') \not\sse \purec(G'')$, contradicting the assumption. \Qed
\end{prf}

By Lemma \ref{lem:mpsdoping} and Lemma \ref{lem:purecprimecdoped} we obtain:
\begin{thm}\label{thm:dopedmpsprc}
  Consider $F \in \Cls$. Then the map $F' \in \mps(F) \mapsto \purec(\doping(F')) \in \Cl$ is a bijection from $\mps(F)$ to $\primec_0(\doping(F))$.
\end{thm}
Theorem \ref{thm:dopedmpsprc} together with the description of the inversion map in Lemma \ref{lem:purecprimecdoped} yields computation of the set $\mps(F)$ for $F \in \Cls$ via computation of $\primec_0(\doping(F))$.

\begin{corol}\label{cor:surjprimeFG}
  For $F \in \Cls$ we obtain a map from $\primec_0(\doping(F))$ to the set of implicates of $F$ covering $\primec_0(F)$ by the mapping $C \in \primec_0(\doping(F)) \mapsto C \sm V$ for $V := \set{u_C : C \in F}$.
\end{corol}
\begin{prf}
  The given map can be obtained as a composition as follows: For $C \in \primec_0(\doping(F))$ take (the unique) $F' \in \mps(F)$ with $\purec(\doping(F')) = C$, and we have $C \sm V = \purec(F')$. \Qed
\end{prf}

\subsection{Hardness of doped clause-sets}
\label{sec:hddopedcls}

The hardness of a doped clause-set is the maximal hardness of sub-clause-sets of the original clause-set:
\begin{lem}\label{lem:hddoping}
  For $F \in \Cls$ we have $\hardness(\doping(F)) = \max_{F' \sse F} \hardness(F')$.
\end{lem}
\begin{prf}
  We have $\hardness(F') \le \hardness(\doping(F))$ for all $F' \sse F$, since via applying a suitable partial assignment we obtain $F'$ from $F$, setting the doping-variables in $F'$ to false, and the rest to true. And if we consider an arbitrary partial assignment $\vp$ with $\vp * \doping(F) \in \Usat$, then w.l.o.g.\ all doping variables are set (we can set the doping-variables not used by $\vp$ to true, since these variables are all pure), and then we have a partial assignment making $F'$ unsatisfiable for that $F' \in \Usat$ given by all the doping variables set by $\vp$ to false. \Qed
\end{prf}

\begin{examp}\label{exp:hddot}
  For an example of a clause-set $F \in \Usat$ with $\hardness(\doping(F)) > \hardness(F)$ consider any clause-set $F' \in \Cls$ with $\hardness(F') > 0$, and then take $F := F' \cup \set{\bot}$ (note that $\bot \notin F'$). Thus $\hardness(F) = 0$. And by Part 1 of Lemma 6.5 in \cite{GwynneKullmann2012Slur,GwynneKullmann2012SlurJ}, all $\Urefc_k$ are closed under partial assignments, so for $\vp := \pao{u_\bot}{1} \cup \pab{u_C \ra 0 \mb C \in F'}$ we have $\hardness(\doping(F)) \ge \hardness(\vp * \doping(F)) = \hardness(F') > \hardness(F) = 0$.
\end{examp}

\section{Doping tree clause-sets}
\label{sec:doptreecls}

As explained in Subsection \ref{sec:introstricteq}, we want to construct boolean functions (given by clause-sets) with a large number of prime implicates, and where we have strong control over these prime implicates. For this purpose we dope ``minimally unsatisfiable clause-sets of deficiency $1$'', that is the elements of $\Smusati{\delta=1}$. First we review in Subsection \ref{sec:prelimMU} the background (for more information see \cite{Kullmann2007HandbuchMU}). In Subsection \ref{sec:totalmps} we show that these clause-sets are the core of ``total minimal premise sets'', which have as many minimal-premise sub-clause-sets as possible. In Theorem \ref{thm:dopedsmumax} we show that $F \in \Smusati{\delta=1}$ are precisely the unsatisfiable clause-sets such that every non-empty subset is an mps. Then in Subsection \ref{sec:appSMU1dop} we consider doping of these special clause-sets, and in Theorem \ref{thm:sumdsmuo} we determine basic properties of $\doping(F)$.

\subsection{Preliminaries on minimal unsatisfiability}
\label{sec:prelimMU}

A minimally unsatisfiable $F \in \Musat$ is \emph{saturated minimally unsatisfiable} iff for all clauses $C \in F$ and for every literal $x$ with $\var(x) \notin \var(C)$ the clause-set $(F \sm C) \cup (C \cup \set{x})$ is satisfiable. The set of all saturated minimally unsatisfiable clause-sets is denoted by $\bmm{\Smusat} \subset \Musat$. By \bmm{\Smusati{\delta=k}} we denote the set of $F \in \Smusat$ with $\delta(F) = k$, where the \emph{deficiency} of a clause-set $F$ is given by $\delta(F) := c(F) - n(F)$. In \cite{Ku99dKo} (generalised in \cite{Kullmann2007ClausalFormZII}) it is shown that the elements of $\Smusati{\delta=1}$ are exactly the clause-sets introduced in \cite{Co73}. The details are as follows. For rooted trees $T$ we use \bmm{\nds(T)} for the set of nodes and $\bmm{\lvs(T)} \sse \lvs(T)$ for the set of leaves, and we set $\bmm{\nnds(T)} := \abs{\nds(T)}$ and $\bmm{\nlvs(T)} := \abs{\lvs(T)}$. In our context, the nodes of rooted trees are just determined by their positions, and do not have names themselves. Another useful notation for a tree $T$ and a node $w$ is \bmm{T_w}, which is the sub-tree of $T$ with root $w$; so $\lvs(T) = \set{w \in \nds(T) : \nnds(T_w) = 1}$. Recall that for a full binary tree $T$ (every non-leaf node has two children) we have $\nnds(T) = 2 \nlvs(T) - 1$.

\begin{defi}\label{def:vardisjointreerep}
  Consider a full binary tree $T$ and an injective vertex labelling $u : (\nds(T) \sm \lvs(T)) \ra \Va$ for the inner nodes; the set of all such pairs is denoted by \bmm{\Tsmuo}. The induced edge-labelling assigns to every edge from an inner node $w$ to a child $w'$ the literal $u(w)$ resp.\ $\ol{u(w)}$ for a left resp.\ right child. We define the \textbf{clause-set representation $\bmm{\smuo(T,u)}$} (where ``1'' reminds of deficiency $1$ here; see Lemma \ref{lem:charsmu}) to be $\bmm{\smuo(T,u)} := \set{ C_w : w \in \lvs(T) }$, where clause \bmm{C_w} consists of all the literals (i.e., edge-labels) on the path from the root of $T$ to $w$.
\end{defi}
By Lemma C.5 in \cite{Ku99dKo}:
\begin{lem}\label{lem:charsmu} $\smuo: \Tsmuo \ra \Smusati{\delta=1}$ is a bijection.
\end{lem}
By $\bmm{\tsmuo}: \Smusati{\delta=1} \ra \Tsmuo$ we denote the inversion of $\smuo$. Typically we identify $(T,u) \in \Tsmuo$ with $T$, and let the context determine $u$. So $\tsmuo(F)$ is the full binary tree, where the variable $v$ labelling the root (for $F \not= \set{\bot}$) is the unique variable occurring in every clause of $F$, and the clause-sets determining the left resp.\ right subtree are $\pao v0 * F$ resp.\ $\pao v1 * F$. By $\bmm{w_C}$ for $C \in F$ we denote the leaf $w$ of $\tsmuo(F)$ such that $C_w = C$. Furthermore we identify the literals of $F$ with the edges of $\tsmuo(F)$. Note that $c(F) = \nlvs(\tsmuo(F))$ and $n(F) = \nnds(\tsmuo(F)) - \nlvs(\tsmuo(F))$.

\begin{examp}\label{exp:dopedsmutree}
  Consider the following labelled binary tree $T$:
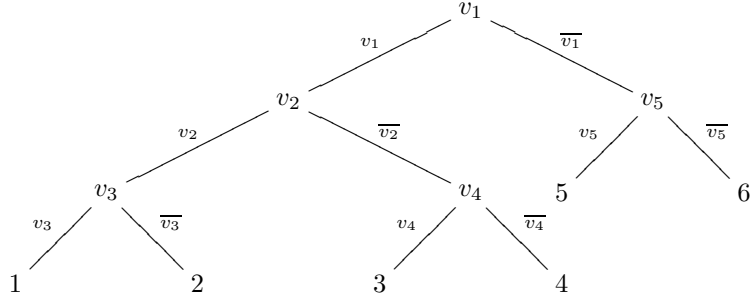
  \begin{displaymath}
    \xygraph{
      !{0;/r8ex/:}
        []{v_1} (
          - [dll]{v_2}_{v_1} (
            -[dll]{v_3}_{v_2} (
              -[dl]{1}_{v_3} (),
              -[dr]{2}^{\ol{v_3}} ()
            ),
            -[drr]{v_4}^{\ol{v_2}} (
              -[dl]{3}_{v_4} (),
              -[dr]{4}^{\ol{v_4}} ()
            )
          ),
          - [drr]{v_5}^{\ol{v_1}} (
            -[dl]{5}_{v_5} (),
            -[dr]{6}^{\ol{v_5}} ()
          )
        )
    }
  \end{displaymath}
  Then $\smuo(T) = \set{ \set{v_1,v_2,v_3}, \set{v_1,v_2,\ol{v_3}}, \set{v_1,\ol{v_2},v_4}, \set{v_1,\ol{v_2},\ol{v_4}}, \set{\ol{v_1},v_5}, \set{v_1,\ol{v_5}} }$, where for example $C_3 = \set{v_1,\ol{v_2},v_4}$ and $w_{\set{v_1,\ol{v_5}}} = 6$.
\end{examp}

We note in passing, that those $\smuo(T)$ with $\hts(T) \le 1$ can be easily characterised as follows. A clause $C \in F$ for $F \in \Cls$ is called \emph{full} if $\var(C) = \var(F)$, that is, $C$ contains all variables of $F$.
\begin{lem}\label{lem:characsmuhts1}
  $F \in \Smusati{\delta=1}$ contains a full clause if and only if $\hts(\tsmuo(F)) \le 1$.
\end{lem}
See Example \ref{exp:sepUC01} for more on these special clause-sets. The effect of applying a partial assignment to some element of $\Smusati{\delta=1}$ is easily described as follows:
\begin{lem}\label{lem:apppasmuo}
  Consider $F \in \Smusati{\delta=1}$ and $x \in \lit(F)$, and let $F' := \pao x1 * F$. We have:
  \begin{enumerate}
  \item $F' \in \Smusati{\delta=1}$.
  \item Let $T := \tsmuo(F)$ and $T' := \tsmuo(F')$. The tree $T'$ is obtained from $T$ as follows:
    \begin{enumerate}
    \item Consider the node $w \in T$ labelled with $\var(x)$. Let $T_x, T_{\ol{x}}$ be the two subtrees hanging at $w$, following the edge labelled with $x$ resp.\ $\ol{x}$.
    \item Now $T'$ is obtained from $T'$ by removing subtree $T_x$, and attaching $T_{\ol{x}}$ directly at position $w$.
    \end{enumerate}
  \end{enumerate}
\end{lem}

\begin{examp}\label{exp:apppasmuo}
  Consider the labelled binary tree $T$ from Example \ref{exp:dopedsmutree} where
  \begin{displaymath}
    \smuo(T) = \set{ \underbrace{\set{v_1,v_2,v_3}}_{\bmm{C_1}}, \underbrace{\set{v_1,v_2,\ol{v_3}}}_{\bmm{C_2}}, \underbrace{\set{v_1,\ol{v_2},v_4}}_{\bmm{C_3}}, \underbrace{\set{v_1,\ol{v_2},\ol{v_4}}}_{\bmm{C_4}}, \underbrace{\set{\ol{v_1},v_5}}_{\bmm{C_5}}, \underbrace{\set{v_1,\ol{v_5}}}_{\bmm{C_6}} }
  \end{displaymath}
  Now consider the application of the partial assignment $\pao {v_2}1$ to $\smuo(T)$:
  \begin{enumerate}
  \item Clauses $C_1$ and $C_2$ are satisfied, and so are removed (both contain $v_2$).
  \item Clauses $C_3$ and $C_4$ both contain $\ol{v_2}$ and so this literal is removed.
  \end{enumerate}
  This yields:
  \begin{displaymath}
    \pao {v_2}1 * \smuo(T) = \set{ \underbrace{\set{v_1,v_4}}_{\bmm{C_3 \sm \set{\ol{v_2}}}}, \underbrace{\set{v_1,\ol{v_4}}}_{\bmm{C_4 \sm \set{\ol{v_2}}}}, \underbrace{\set{\ol{v_1},v_5}}_{\bmm{C_5}}, \underbrace{\set{v_1,\ol{v_5}}}_{\bmm{C_6}} }
  \end{displaymath}
  The satisfaction (removal) of clauses and removal of literals is illustrated directly on $T$ in Figure \ref{fig:dopedsmutreepass} with dotted and dashed lines for clause and literal removal respectively. The tree corresponding to $\pao {v_2}1 * \smuo(T)$ is illustrated in Figure \ref{fig:dopedsmutreepassafter}.
  \begin{figure}[ht]
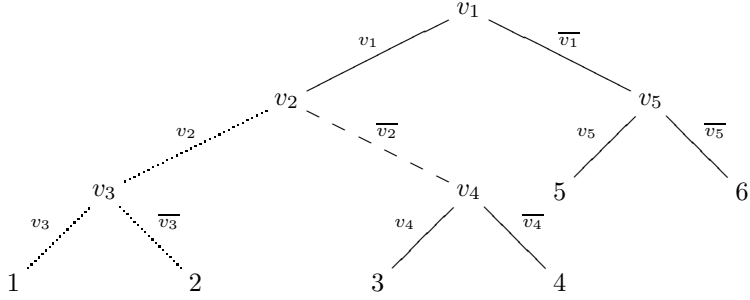

    \begin{displaymath}
    \xygraph{
      !{0;/r8ex/:}
        []{v_1} (
          - [dll]{v_2}_{v_1} (
            -@{..}[dll]{v_3}_{v_2} (
              -@{..}[dl]{1}_{v_3} (),
              -@{..}[dr]{2}^{\ol{v_3}} ()
            ),
            -@{--}[drr]{v_4}^{\ol{v_2}} (
              -[dl]{3}_{v_4} (),
              -[dr]{4}^{\ol{v_4}} ()
            )
          ),
          - [drr]{v_5}^{\ol{v_1}} (
            -[dl]{5}_{v_5} (),
            -[dr]{6}^{\ol{v_5}} ()
          )
        )
    }
  \end{displaymath}
  \caption{Illustration of application of $\pao {v_2}1$ to $\smuo(T)$. Dotted lines indicate that the clauses corresponding to the effected leaves are satisfied; dashed lines indicate that the corresponding literal is falsified and therefore removed from all clauses.}
  \label{fig:dopedsmutreepass}
  \end{figure}
  \begin{figure}
    \begin{displaymath}
    \xygraph{
      !{0;/r8ex/:}
        []{v_1} (
          - [dll]{v_4}_{v_1} (
            -[dl]{3}_{v_4} (),
            -[dr]{4}^{\ol{v_4}} ()
          ),
          - [drr]{v_5}^{\ol{v_1}} (
            -[dl]{5}_{v_5} (),
            -[dr]{6}^{\ol{v_5}} ()
          )
        )
    }
  \end{displaymath}
  \caption{Tree associated with $\pao {v_2}1 * \smuo(T)$.}
  \label{fig:dopedsmutreepassafter}
  \end{figure}
\end{examp}

\begin{corol}\label{cor:SMU1stable}
  $\Smusati{\delta=1}$ is stable under application of partial assignments, that is, for $F \in \Smusati{\delta=1}$ and $\vp \in \Pass$ holds $\vp * F \in \Smusati{\delta=1}$.
\end{corol}

From Lemma \ref{lem:charsmu} follows $\Smusati{\delta=1} \subset \Uclash$, where $\bmm{\Clash} \subset \Cls$ is the set of \emph{hitting clause-sets}, that is, those $F \in \Cls$ where every two clauses clash in at least one literal, i.e., for all $C, D \in F$, $C \not= D$, we have $\abs{C \cap \ol{D}} \ge 1$, and $\bmm{\Uclash} := \Clash \cap \Usat$. It is well-known that $\Uclash \subset \Smusat$ holds (for a proof see Lemma 2 in \cite{KullmannZhao2012Confluence}).

\subsection{Total minimal premise sets}
\label{sec:totalmps}

We are interested in clause-sets which have as many sub-mps's as possible:
\begin{defi}\label{def:tmps}
  A clause-set $F \not= \top$ is a \textbf{total mps} if $\mps(F) = \pot(F) \sm \set{\top}$.
\end{defi}
Every total mps is an mps.
\begin{examp}\label{exp:totalmps}
  $\set{\set{a,b},\set{\ol{a},b},\set{\ol{b}}}$ is a total mps, while $\set{\set{a,b},\set{\ol{a}},\set{\ol{b}}}$ is an mps (since minimally unsatisfiable), but not a total mps.
\end{examp}

To determine all total mps's, the central task to determine the minimally unsatisfiable total mps's. Before we can prove that these are precisely the saturated minimally unsatisfiable clause-sets of deficiency $1$, we need to state a basic property of these clause-sets, which follows by definition of $\tsmuo(F)$ for $F \in \Smusati{\delta=1}$ (recall Subsection \ref{sec:prelimMU}):
\begin{lem}\label{lem:detpureSMU1}
  Consider $F \in \Smusati{\delta=1}$ and $F' \sse F$. Let $T := \tsmuo(F)$. The set $\purec(F')$ of pure literals of $F'$ can be determined as follows:
  \begin{enumerate}
  \item Let $W_{F'} := \set{w_C : C \in F'} \sse \lvs(T)$ be the set of leaves corresponding to the clauses of $F'$.
  \item For a literal $x \in \lit(F)$ let $w \in \nds(T)$ be the node labelled with $\var(x)$, and let $T_x$ the the subtree of $w$ reached by $x$, and let $T_{\ol{x}}$ be the subtree of $w$ reached by $\ol{x}$.
  \item Now $x \in \purec(F')$ if and only if $W_{F'} \cap \lvs(T_x) \not= \es$ and $W_{F'} \cap \lvs(T_{\ol{x}}) = \es$.
  \end{enumerate}
\end{lem}

\begin{examp}\label{exp:purecSMU1}
  Consider the clause-set
  \begin{multline*}
    F := \setb{ \underbrace{\set{v_1,v_2,v_3}}_{\bmm{C_1}}, \underbrace{\set{v_1,v_2,\ol{v_3}}}_{\DST C_2}, \underbrace{\set{v_1,\ol{v_2}, v_4}}_{\bmm{C_3}}, \underbrace{\set{v_1, \ol{v_2}, \ol{v_4}}}_{\bmm{C_4}},\\
      \underbrace{\set{\ol{v_1}, v_5, v_6}}_{\DST C_5}, \underbrace{\set{\ol{v_1}, v_5, \ol{v_6}}}_{\DST C_6}, \underbrace{\set{\ol{v_1}, \ol{v_5}}}_{\bmm{C_7}} }
  \end{multline*}
  and the subset $F' := \set{C_1, C_3, C_4, C_7}$. The tree $\tsmuo(F)$ is as follows, with the dashed edges representing literals not in $\bc F' = \set{v_1,v_2,v_3,v_4,\ol{v_1},\ol{v_2},\ol{v_4},\ol{v_5}}$:
  \begin{displaymath}
    \xygraph{
      !{0;/r7ex/:}
        []{v_1} (
          - [dlll]{v_2}_{v_1} (
            -[dll]{v_3}_{v_2} (
              -[dl]{\bmm{1}}_{v_3} (),
              -@{--}[dr]{2}^{\ol{v_3}} ()
            ),
            -[drr]{v_4}^{\ol{v_2}} (
              -[dl]{\bmm{3}}_{v_4} (),
              -[dr]{\bmm{4}}^{\ol{v_4}} ()
            )
          ),
          - [drrr]{v_5}^{\ol{v_1}} (
            -@{--}[dl]{v_6}_{v_5} (
              -@{--}[dl]{5}_{v_6} (),
              -@{--}[dr]{6}^{\ol{v_6}} ()
            ),
            -[dr]{\bmm{7}}^{\ol{v_5}} ()
          )
        )
    }
  \end{displaymath}
  We have $W_{F'} = \set{1,3,4,7}$ and
  \begin{displaymath}
    \purec(F') = \bc F' \sm \set{\underbrace{v_2,\ol{v_2}}_{\bmm{C_1, C_3}}, \underbrace{v_1,\ol{v_1}}_{\bmm{C_1, C_7}},\underbrace{v_4,\ol{v_4}}_{\bmm{C_3, C_4}}} = \set{v_3, \ol{v_5}}.
  \end{displaymath}
  Now consider $x \in \lit(F)$:
  \begin{enumerate}
  \item For $x = v_3$ holds $\lvs(T_{v_3}) \cap W_{F'} = \set{1}$ and $T_{\ol{v_3}} \cap W_{F'} = \es$, thus $v_3 \in \purec(F')$.
  \item For $x = \ol{v_5}$ holds $\lvs(T_{\ol{v_5}}) \cap W_{F'} = \set{7}$ and $T_{v_5} \cap W_{F'} = \es$, thus $\ol{v_5} \in \purec(F')$.
  \item Considering for example $x = v_1$, we have $\lvs(T_{v_1}) \cap W_{F'} = \set{1,3}$ and $\lvs(T_{\ol{v_1}}) \cap W_{F'} = \set{7}$, thus $v_1 \notin \purec(F')$, while for $x = v_6$ we have $\lvs(T_{v_6}) \cap W_{F'} = \es$ and $\lvs(T_{\ol{v_6}}) \cap W_{F'} = \es$, thus $v_6 \notin \purec(F')$.
  \end{enumerate}
\end{examp}

\begin{thm}\label{thm:dopedsmumax}
  An unsatisfiable clause-set $F \in \Usat$ is a total mps if and only if $F \in \Smusati{\delta=1}$.
\end{thm}
\begin{prf}
  First assume that $F$ is a total mps. Then every two clauses $C, D \in F$, $C \not= D$, clash in exactly one literal (otherwise $\set{C,D} \notin \Mps$). In \cite{Ku2003e}, Corollary 34, it was shown that that an unsatisfiable clause-sets $F$ has precisely one clash between any pair of different clause-sets iff $F \in \Smusati{\delta=1}$ holds (an alternative proof was found in \cite{SloanSzoerenyiTuran2005Primimplikanten_1}).\footnote{In \cite{Ku2003e} the notation ``$\Uclash$'' was used to denote ``uniform hitting clause-sets'', which is now more appropriately called ``(conflict-)regular hitting clause-sets'', while ``U'' now stands for ``unsatisfiable''.} Now assume $F \in \Smusati{\delta=1}$, and we have to show that $F$ is a total mps. So consider $F' \in \pot(F) \sm \set{\top}$, and let $C := \purec(F)$, $\vp := \vp_C$. Since $F'$ is a hitting clause-set, $\vp$ is contraction-free for $F'$, and according to Lemma \ref{lem:characmps} it remains to show that $F'' := \vp * F'$ is unsatisfiable (recall that hitting clause-sets are irredundant). Assume that $F''$ is satisfiable, and consider a partial assignment $\psi$ with $\psi * F'' = \top$ and $\var(\psi) \cap \var(\vp) = \es$. We show that then $\vp \cup \psi$ would be a satisfying assignment for $F$, contradicting the assumption. To this end it suffices to show that for all $D \in F \sm F'$ holds $\ol{C} \cap D \not= \es$. Consider $T := \tsmuo(F)$, and let $W_{F'}$ be defined as in Lemma \ref{lem:detpureSMU1}. Starting from the leaf $w_D$, let $w$ be the first node on the path to the root of $T$ such that one of the two subtrees of $w$ contains a leaf of $W_{F'}$. Let $\ol{x}$ be the literal at $w$ on the path to $w_D$. So by Lemma \ref{lem:detpureSMU1} we have $x \in C$, while by definition $\ol{x} \in D$. \Qed
\end{prf}

\begin{corol}\label{cor:alltotmps}
  For a clause-set $F \in \Cls$ the following properties are equivalent:
  \begin{enumerate}
  \item $F$ is a total mps.
  \item $\vp_{\purec(F)} * F \in \Smusati{\delta=1}$, and $\vp_{\purec(F)}$ is contraction-free for $F$.
  \end{enumerate}
\end{corol}
\begin{prf}
Let $F' := \vp_{\purec(F)} * F$. If $F$ is a total mps, then by Lemma \ref{lem:characmps} follows $F' \in \Musat$, where $\vp_{\purec(F)}$ is contraction-free for $F$. Also by Lemma \ref{lem:characmps} follows then, that $F' \in \Mps$, and thus by Theorem \ref{thm:dopedsmumax} we obtain $F' \in \Smusati{\delta=1}$. For the other direction, if $F' \in \Smusati{\delta=1}$ holds, where $\vp_{\purec(F)}$ is contraction-free for $F$, then by Theorem \ref{thm:dopedsmumax} follows that $F'$ is a total mps, which by Lemma \ref{lem:characmps} yields that $F$ is a total mps.  \Qed
\end{prf}

Thus we can precisely construct all total mps's, if we start the process described in Corollary \ref{cor:genmps} not with an arbitrary $F \in \Musat$, but with an $F \in \Smusati{\delta=1}$.

\begin{examp}\label{exp:countertmps}
  That every $2$-element sub-clause-set of $F \in \Cls$ is an mps, that is, every two (different) clauses of $F$ clash in precisely one literal, says that $F$ is $1$-regular hitting in the terminology of \cite{Kullmann2007ClausalFormZII}, Section 6. For $F \in \Usat$ the proof of Theorem \ref{thm:dopedsmumax} shows, that $F$ is a total mps iff $F$ is $1$-regular hitting. However for $F \in \Sat$ this is not true, and the simplest example is $F := \set{\set{\ol{a},b},\set{\ol{b},c},\set{\ol{c},a}}$: $F$ is $1$-regular hitting, but has no pure literal and is satisfiable, and thus $F \notin \Mps$. In this case we have $\delta(F) = 0$. For an interesting example with deficiency $1$ see Section 5 in \cite{Ku2003e}.
\end{examp}
We arrive at a simple and perspicuous proof of the main result of \cite{SloanSzoerenyiTuran2005Primimplikanten_1}, that the clause-sets $F$ with $\abs{\primec_0(F)} = 2^{c(F)} - 1$ are precisely the clause-sets $\doping(F)$ for $F \in \Smusati{\delta=1}$ when allowing to replace the single doping variable of a clause by any non-empty set of new (pure) literals:
\begin{lem}\label{lem:characmaxnprimeimpl}
  For $F \in \Cls \sm \set{\top}$ holds $\abs{\primec_0(F)} = 2^{c(F)} - 1$ if and only if the following two conditions hold:
  \begin{enumerate}
  \item $F$ is a total mps.
  \item For every clause $C \in F$ there is $x \in C$ such that $\var(x) \notin \var(F \sm \set{C})$.
  \end{enumerate}
\end{lem}
\begin{prf}
First assume $\abs{\primec_0(F)} = 2^{c(F)} - 1$. Thus the map $F' \in \mps(F) \mapsto \purec(F') \sse \set{C \in \Cl : F \models C}$, which according to Lemma \ref{lem:purecprimec} covers $\primec_0(F)$, must indeed be a bijection from $\mps(F)$ to $\primec_0(F)$, and hence $F$ is a total mps (here we need $F \ne \top$). If there would be $C \in F$ such that for all $x \in C$ we have $\var(x) \in \var(F \sm \set{C})$, then $\purec(F) \sse \purec(F \sm \set{C})$, and thus $F \sm \set{C}$ could not yield a prime implicate different from the prime implicate obtained from $F$.

The inverse direction follows by the observation, that the existence of the unique ``doping literals'' $x \in C$ has the consequence, that for $\top \subset F', F'' \sse F$ with $F' \ne F''$ we get $\purec(F') \ne \purec(F'')$, since these doping literals make a difference. \Qed
\end{prf}

\subsection{Doping $\Smusati{\delta=1}$}
\label{sec:appSMU1dop}

We are turning now our attention to a closer understanding of the prime implicates $C$ of doped $F \in \Smusati{\delta=1}$. We start with their identification with non-empty sub-clause-sets $F'$ of $\doping(F)$:
\begin{lem}\label{lem:dopedsmumax2}
  Consider a clause-set $F \in \Smusati{\delta=1}$. By Theorem \ref{thm:dopedsmumax} each non-empty subset yields a minimal premise set. Thus by Theorem \ref{thm:dopedmpsprc} we have:
  \begin{enumerate}
  \item $\primec_0(\doping(F)) = \set{ \purec(F') \mb \top \not= F' \sse \doping(F) }$.
  \item $\abs{\primec_0(\doping(F))} = 2^{c(F)} - 1$.
  \end{enumerate}
\end{lem}
Since the clauses of $\doping(F)$ can be identified with leaves of the tree $\tsmuo(F)$, we obtain a bijection between non-empty sets $V$ of leaves of the tree $\tsmuo(F)$ and prime implicates of $\doping(F)$:
\begin{defi}\label{def:clausesDSMU1}
  For $F \in \Smusati{\delta = 1}$ and $\es \not= V \sse \lvs(\tsmuo(F))$ the clause \bmm{C_V} is the prime implicate $\purec(\set{C_w \in F \mb w \in V})$ of $\doping(F)$ according to Lemma \ref{lem:dopedsmumax2}. For $w \in \lvs(\tsmuo(F))$ we furthermore set $\bmm{u_w} := u_{C_w}$.
\end{defi}

By Lemma \ref{lem:dopedsmumax2}:
\begin{lem}\label{lem:dopedsmuprimetreechar}
  For $F \in \Smusati{\delta = 1}$ holds $\primec_0(\doping(F)) = \set{ C_V \mb \es \not= V \sse \lvs(\tsmuo(F)) }$.
\end{lem}

How precisely from $V \sse \lvs(\tsmuo(F))$ the prime implicate $C_V$ is constructed shows the following lemma:
\begin{lem}\label{lem:dopedsmuprimechar}
  Consider $F \in \Smusat_{\delta=1}$ and $\es \not= V \sse \lvs(\tsmuo(F))$. We have $C_V = U_V \cup P_V$, $U_V \cap P_V = \es$, where
  \begin{enumerate}
  \item $U_V := \set{u_w \mb w \in V}$, and
  \item $P_V := \purec(F')$ for $F' := \set{C_w : w \in V}$ as given in Lemma \ref{lem:detpureSMU1}, that is, $P_V$ is the set of literals $x$ such that $V \cap \lvs(T_x) \not= \es$ and $V \cap \lvs(T_{\ol{x}}) = \es$.
  \end{enumerate}
\end{lem}

\begin{examp}\label{exp:CV}
  Consider the clause-set
  \begin{displaymath}
    F := \set{ \set{v_1,v_2}, \set{v_1,\ol{v_2}}, \set{\ol{v_1},v_3}, \set{\ol{v_1},\ol{v_3}} } \in \Smusat_{\delta=1}
  \end{displaymath}
  corresponding to the tree
  \begin{displaymath}
    \xygraph{
      !{0;/r8ex/:}
        []{v_1} (
          - [dll]{v_2}_{v_1} (
            -[dl]{\bmm{1}}_{v_2} (),
            -[dr]{2}^{\ol{v_2}} ()
          ),
          - [drr]{v_3}^{\ol{v_1}} (
            -[dl]{\bmm{3}}_{v_3} (),
            -[dr]{4}^{\ol{v_3}} ()
          )
        )
    }
  \end{displaymath}
  with the doped clause-set
  \begin{displaymath}
    \doping(F) = \set{ \set{v_1,v_2,u_1}, \set{v_1,\ol{v_2},u_2}, \set{\ol{v_1},v_3,u_3}, \set{\ol{v_1},\ol{v_3}, u_4 }}.
  \end{displaymath}
  Now consider the set $V := \set{1,3}$. According to Definition \ref{def:clausesDSMU1} we have that $C_V = \purec(\set{ \set{v_1,v_2,u_1}, \set{\ol{v_1},v_3,u_3} }) = \set{v_2,v_3,u_1,u_3}$. By Lemma \ref{lem:dopedsmuprimechar} we have that $C_V = U_V \cup P_V$, where $U_V = \set{u_1, u_3}$ and $P_V = \purec( \set{ \set{v_1,v_2}, \set{\ol{v_1},v_3} } = \set{v_2,v_3}$. Note that for both $x \in \set{v_2,v_3} = P_V$ we have that $\lvs(T_{x}) \cap V \not= \es$ and $\lvs(T_{\ol{x}}) \cap V = \es$, but we do not have this for $x \in \lit(F) \sm \set{v_2,v_3}$.
\end{examp}

The hardness of $F$ as well as $\doping(F)$ is the Horton-Strahler number of $\tsmuo(F)$:
\begin{lem}\label{lem:hdsmu1}
  Consider $F \in \Smusati{\delta=1}$, and let $k := \hts(\tsmuo(F))$. Then we have $\hardness(F) = \hardness(\doping(F)) = k$.
\end{lem}
\begin{prf}
Let $T := \tsmuo(F)$. First we show $\hardness(F) = k$. We have $\hardness(F) \le k$, since $T$ is by definition of $F = \smuo(T)$ already a resolution tree (when extending the labelling of leaves to all nodes), deriving $\bot$ from $F$. To show $\hardness(F) \ge k$, we use Lemma \ref{lem:lbhd} with $\mc{C} := \Smusati{\delta=1}$ and $h(F) := \hts(\tsmuo(F))$. Based on Lemma \ref{lem:apppasmuo}, we consider the effect on the Horton-Strahler number of assigning a truth value to one variable $v \in \var(F)$. Let $w \in \nds(T)$ be the (inner) node labelled with $v$, and let $T^w_0, T^w_1$ be the left resp.\ right subtree hanging at $w$. Now the effect of assigning $\ve \in \set{0,1}$ to $v$ is to replace $T_w$ with $T^w_{\ve}$. Let $T_{\ve}$ be the (whole) tree obtained by assigning $\ve$ to $v$, that is, $T_{\ve} := \tsmuo(\pao {v}{\ve} * F)$. If $\hts(T^w_0) = \hts(T^w_1)$, then we have $\hts(T_{\ve}) \ge k-1$, since at most one increase of the Horton-Strahler number for subtrees is missed out now. Otherwise we have $\hts(T_0) = \hts(T)$ or $\hts(T_1) = \hts(T)$, since removal of the subtree with the smaller Horton-Strahler number has no influence on the Horton-Strahler number of the whole tree. So altogether Lemma \ref{lem:lbhd} is applicable, which concludes the proof of $\hardness(F) = k$.

For showing $\hardness(\doping(F)) = k$ we use Lemma \ref{lem:hddoping}: so consider $F' \sse F$ and $\vp \in \Pass$ with $\vp * F' \in \Usat$, let $F'' := \vp * F'$, and we have to show $\hardness(F'') \le k$. W.l.o.g.\ $\var(\vp) \sse \var(F')$. By Corollary \ref{cor:SMU1stable} we have that $\vp *F \in \Smusati{\delta=1}$, and thus $\vp *F = F''$ must hold, and $\hardness(F'') = \hts(\tsmuo(F''))$ (by the first part). By Lemma \ref{lem:apppasmuo}, $\tsmuo(F'')$ results from $T$ by a sequence of removing subtrees, and it is easy to see, that thus $\hts(\tsmuo(F'')) \le k$ holds. \Qed
\end{prf}

We summarise what we have learned about $\doping(F)$ for $F \in \Smusati{\delta=1}$:
\begin{thm}\label{thm:sumdsmuo}
  Consider $F \in \Smusati{\delta=1}$.
  \begin{enumerate}
  \item For each clause-set $F'$ equivalent to $\doping(F)$ there is an injection $i: \doping(F) \ra F'$ with $\fa\, C \in \doping(F) : C \sse i(C)$ (by Lemma \ref{lem:necprcdp}).
  \item $\doping(F)$ is a total mps (by Corollary \ref{cor:alltotmps}).
  \item The prime implicates of $\doping(F)$ are given by Lemmas \ref{lem:dopedsmuprimetreechar}, \ref{lem:dopedsmuprimechar}.
  \item $\hardness(\doping(F)) = \hts(\tsmuo(F))$ (by Lemma \ref{lem:hdsmu1}).
  \end{enumerate}
\end{thm}

\section{Lower bounds}
\label{sec:lowerb}

This section proves the main result of this article, Theorem \ref{thm:separation}, which exhibits for every $k \ge 0$ sequences $(F^k_h)_{h \in \NN}$ of small clause-sets of hardness $k+1$, where every equivalent clause-set of hardness $k$ (indeed of w-hardness $k$) is of exponential size. In this way we show that the $\Urefc_k$ hierarchy is useful, i.e., equivalent clause-sets with higher hardness can be substantially shorter. These $F^k_h$ are doped versions of clause-sets from $\Smusati{\delta=1}$ (recall Theorem \ref{thm:sumdsmuo}), which are ``extremal'', that is, their underlying trees $\tsmuo(F^k_h)$ are for given Horton-Strahler number $k+1$ and height $h$ as large as possible.

The organisation of this section is as follows: In Subsection \ref{sec:triggerhyp} the main tool for showing size-lower-bounds for equivalent clause-sets of a given (w-)hardness is established in Theorem \ref{thm:triggersetmethod}. Subsection \ref{sec:est} introduces the ``extremal trees''. Subsection \ref{sec:hierbadrep} shows the main lower bound in Theorem \ref{thm:nogoodksoft}, and applies it to show the separation Theorem \ref{thm:separation}.

\subsection{Trigger hypergraphs}
\label{sec:triggerhyp}

Our goal is to construct clause-sets $F^k_h$ of hardness $k+1$, which have no short equivalent clause-set $F$ with $\whardness(F) \le k$, where w.l.o.g.\ $F \sse \primec_0(F^k_h) = \primec_0(F)$. This subsection is about the general lower-bound method. How are we going to find a lower bound on the number of clauses of $F$ ? The property $\whardness(F) \le k$ means, that for every $C \in \primec_0(F)$ the unsatisfiable clause-set $\vp_C * F$ can be refuted by $k$-resolution. In order for $k$-resolution to have a chance, there must be at least one clause of length at most $k$ in $\vp_C * F$ --- and this necessary condition is all what we consider. So our strategy is to show that every $F \sse \primec_0(F^k_h)$, such that for all $C \in \primec_0(F^k_h)$ there is a clause of length at most $k$ in $\vp_C * F$, is big.

It is useful to phrase this approach in hypergraph terminology. Recall that a hypergraph is a pair $G = (V,E)$, where $V$ is a set (of ``vertices'') and $E \sse \pot(V)$ (the set of hyperedges), where one uses $V(G) := V$ and $E(G) := E$. A \emph{transversal} of $G$ is a set $T \sse V(G)$ such that for all $E \in E(G)$ holds $T \cap E \not= \es$. The minimum size of a transversal is denoted by \bmm{\tau(G)}, the \textbf{transversal number}.
\begin{defi}\label{def:triggerhypergraph}
  Consider $k \in \NNZ$ and $F \in \Cls$. The \textbf{trigger hypergraph} $\trighyp{F}{k}$ is the hypergraph with the prime implicates of $F$ as its vertices, and for every prime implicate $C$ of $F$ a hyperedge $E^k_C$. The hyperedge $E^k_C$ contains all prime implicates $C' \in \primec_0(F)$ which are not satisfied by $\vp_C$ and yield a clause of size at most $k$ under $\vp_C$. That is,
  \begin{enumerate}
  \item $V(\trighyp{F}{k}) := \primec_0(F)$, and
  \item $E(\trighyp{F}{k}) := \set{E^k_C \mb C \in \primec_0(F)}$,
  \end{enumerate}
  where $E^k_C := \set{C' \in \primec_0(F) \mb C' \cap \ol{C} = \es \und \abs{C' \sm C} \le k}$.
\end{defi}
Note that the trigger hypergraph of $F \in \Cls$ depends only on the underlying boolean function of $F$, and thus for every equivalent $F'$ we have $\trighyp{F'}{k} = \trighyp{F}{k}$.
\begin{examp}\label{exp:trighyp}
  Consider the clause-set
  \begin{displaymath}
    F := \setb{\underbrace{\set{v_1,\ol{v_3},\ol{v_4}}}_{C_1},\underbrace{\set{v_2,v_3,\ol{v_4}}}_{C_2},\underbrace{\set{v_2,\ol{v_3},v_4}}_{C_3},
               \underbrace{\set{\ol{v_2},v_3,v_4}}_{C_4},\underbrace{\set{v_1,v_3,v_4}}_{C_5},\underbrace{\set{v_1,v_2}}_{C_6}}.
  \end{displaymath}
  As shown in Example 8.2 of \cite{GwynneKullmann2012Slur,GwynneKullmann2012SlurJ} we have $\primec_0(F) = F$. The trigger hypergraph $\trighyp{F}{0}$ is (as always) the hypergraph with all singleton sets, i.e., $E(\trighyp{F}{0}) = \setb{\set{C_1},\dots,\set{C_6}}$. The hypergraphs $\trighyp{F}{k}$ for $k \in \set{1,2}$ are represented by Figures \ref{fig:T1F}, \ref{fig:T2F}.

  \begin{figure}[ht]
    \begin{minipage}[b]{0.45\linewidth}
      \centering
      $\xymatrix{
        C_1 \ar@(u,l) \ar@/^/@{.>}[dr] & C_2 \ar@(u,r) \ar@/^/@{.>}[d] & C_5  \ar@(u,r) \ar[d] \ar@/^/@{.>}[dl] \\
        C_3 \ar@(d,l) \ar@/^/@{.>}[r] & C_6 \ar@(r,d) \ar[ul] \ar[u] \ar[ur] \ar[l]  & C_4 \ar@(d,r) \ar[u]
      }$
      \vspace{0.2cm}
      \caption{$\trighyp{F}{1}$}
      \label{fig:T1F}
    \end{minipage}
    \hspace{0.5cm}
    \begin{minipage}[b]{0.45\linewidth}
      \centering
      $\xymatrix{
        C_1 \ar@(u,l) \ar[dr] & C_2 \ar@(u,r) \ar[d] & C_5  \ar@(u,r) \ar[d] \ar[dl] \\
        C_3 \ar@(d,l) \ar[r] & C_6 \ar@(r,d) \ar[ul] \ar[u] \ar[ur] \ar[l]  & C_4 \ar@(d,r) \ar[u]
      }$
      \vspace{0.2cm}
      \caption{$\trighyp{F}{2}$}
      \label{fig:T2F}
    \end{minipage}
  \end{figure}

  To interpret the diagrams:
  \begin{enumerate}
  \item An arrow from a clause $C$ to a clause $D$ represents that $C \in E^k_D$.
  \item A dotted arrow from $C$ to $D$ represents that $\abs{D \sm C} > k$ (so $C \notin E^k_D$), but $C \cap \ol{D} = \es$, and thus for some large enough $k' > k$ we will have $C \in E^{k'}_D$.
  \item No arrow between $C$ and $D$ indicates that $C \cap \ol{D} \not= \es$ (i.e., for all $k'$ we have $C \notin E^k_D$ and $D \notin E^k_C$).
  \item The size of a hyperedge $E^k_D$ is the in-degree of the vertex $D$.
  \end{enumerate}
  Consider $E^1_{C_6} = \set{C_6}$ and $E^2_{C_6} = \set{C_1,C_2,C_3,C_5,C_6}$. As we will see in Lemma \ref{lem:transreptrighyp}, therefore every $F' \sse F$ equivalent to $F$ such that $F' \in \Urefc_1$ must have $C_6 \in F'$. However, $E^2_{C_6}$ contains more clauses than $E^1_{C_6}$, and for example $F \sm \set{C_6} \in \Urefc_2 \sm \Urefc_1$ as shown in Example 8.2 of \cite{GwynneKullmann2012Slur,GwynneKullmann2012SlurJ}. Using the above diagrammatic notation, we can also see that for all $k' \ge 2$ we have $\trighyp{F}{k'} = \trighyp{F}{2}$, as there are no dotted lines for $\trighyp{F}{2}$ (i.e., no clauses $C$ and $D$ such that $\abs{D \sm C} > 2$ but $C \cap \ol{D} = \es$).
\end{examp}

The point of the trigger hypergraph $\trighyp{F}{k}$ is, that every clause-set equivalent to $F$ and of w-hardness at most $k$ must be a transversal of it:
\begin{lem}\label{lem:transreptrighyp}
  Consider $k \in \NNZ$ and $F \in \Cls$ with $\whardness(F) \le k$. Then there is a clause-set $F'$ such that
  \begin{enumerate}
  \item\label{lem:transreptrighyp1} $F' \sse \primec_0(F)$ and $F'$ is equivalent to $F$;
  \item\label{lem:transreptrighyp2} there is an injection $i: F' \ra F$ such that $\fa\, C \in F' : C \sse i(C)$;
  \item\label{lem:transreptrighyp3} $\whardness(F') \le k$;
  \item\label{lem:transreptrighyp4} $F'$ is a transversal of $\trighyp{F}{k}$.
  \end{enumerate}
\end{lem}
\begin{prf}
  Obtain $F'$ from $F$ by choosing for every $C \in F$ some $C' \in \primec_0(F)$ with $C' \sse C$. Then the first two properties are obvious, while Property \ref{lem:transreptrighyp3} follows from Part 1 of Lemma 6.1 in \cite{Ku00g}. Assume that $F'$ is not a transversal of $\trighyp{F}{k}$, that is, there is $C \in \primec_0(F)$ with $F' \cap E^k_C = \es$. Then $\vp_C * F' \in \Usat$, but every clause has length strictly greater than $k$, and thus $k$-resolution does not derive $\bot$ from $\vp_C * F'$, contradicting $\whardness(F') \le k$. \Qed
\end{prf}

Our lower bound method is now captured by the following theorem, which directly follows from Lemma \ref{lem:transreptrighyp}:
\begin{thm}\label{thm:triggersetmethod}
  For $k \in \NNZ$ and $F \in \Wrefc_k$ we have $c(F) \ge \tau(\trighyp{F}{k})$.
\end{thm}
Instead of lower-bounding the transversal number of $\trighyp{F}{k}$, we use that every transversal has to have at least as many elements as there are disjoint hyperedges. So let \bmm{\nu(G)} be the \textbf{matching number} of hypergraph $G$, the maximum number of pairwise disjoint hyperedges; we have $\tau(G) \ge \nu(G)$ for all hypergraphs $G$. So we have to show that there is a set $S \sse \primec_0(F^k_h)$ of exponential size, such that the hyperedges $E^k_C$ for $C \in S$ are pairwise disjoint. For $F^k_h$ we use the doped clause-set $\doping(\smuo(T))$ as considered in Subsection \ref{sec:appSMU1dop}, where the special trees $T$ are constructed in the subsequent subsection.

\subsection{Extremal trees}
\label{sec:est}

For a given hardness $k \ge 1$ we need to construct (full binary) trees which are as large as possible; this is achieved by specifying the height, and using trees which are ``filled up'' completely for the given parameter values:
\begin{defi}\label{def:exstrahltree}
  A pair $(k, h) \in \NNZ^2$ with $h \ge k$ and $k=0 \Ra h=0$ is called an \textbf{allowed parameter pair}. For an allowed parameter pair $(k,h)$ a full binary tree $T$ is called an \textbf{extremal tree of Horton-Strahler number \bmm{k} and height \bmm{h}} if
  \begin{enumerate}
  \item $\hts(T) = k$, $\height(T) = h$;
  \item for all $T'$ with $\hts(T') \le k$ and $\height(T') \le h$ we have $\nds(T') \le \nds(T)$.
  \end{enumerate}
   We denote the set of all extremal trees with Horton-Strahler number $k$ and height $h$ by \bmm{\exhst{k}{h}}.
\end{defi}
Note that for allowed parameter pairs $(k,h)$ we have $k = 0 \Lra h = 0$. Extremal trees are easily characterised and constructed as follows:
\begin{enumerate}
\item $\exhst 00$ contains only the trivial tree (with one node).
\item $\exhst 1h$ for $h \in \NN$ consists exactly of the full binary trees $T$ with $\hts(T) = 1$ and $\height(T) = h$, which can also be characterised as those full binary trees $T$ with $\height(T) = h$ such that every node has at least one child which is a leaf.
\item For $k \ge 2$ and $h \ge k$ we have $T \in \exhst{k}{h}$ iff $T$ has the left subtree $T_0$ and the right subtree $T_1$, and there is $\ve \in \set{0,1}$ with $T_\ve \in \exhst{k-1}{h-1}$ and $T_{1-\ve} \in \exhst{\min(k,h-1)}{h-1}$.
\end{enumerate}
\begin{lem}\label{lem:existextr}
  For all allowed parameter pair $(k,h)$ we have $\exhst{k}{h} \not= \es$.
\end{lem}
The unique elements of $\exhst kk$ for $k \in \NNZ$ are the perfect binary trees of height $k$, which are the smallest binary trees of Horton-Strahler number $k$.

\begin{lem}\label{lem:numexleaves}
 For an allowed parameter pair $(k, h)$ and for $T \in \exhst{k}{h}$ we have $\nlvs(T) = \bmm{\exstrahlersize{k}{h}} := \sum_{i=0}^k \binom{h}{i}$. We have $\alpha(k,h) = \Theta(h^k)$ for fixed $k$.
\end{lem}
\begin{prf}
For $k \le 1$ we have $\alpha(0,0) = 1$ and $\alpha(1,h) = 1 + h$. which are obviously correct. Now consider $k \ge 2$. By induction hypothesis we get
\begin{displaymath}
  \nnds(T) = \alpha(k-1,h-1) + \alpha(\min(k,h-1),h-1).
\end{displaymath}
If $h = k$, then $\alpha(k,h) = 2^k$ (for all $k$), and we get $\nnds(T) = \alpha(k-1,k-1) + \alpha(k-1,k-1) = 2 \cdot 2^{k-1} = 2^k = \alpha(k,k)$. Otherwise we have
\begin{multline*}
  \nnds(T) = \alpha(k-1,h-1) + \alpha(k,h-1) =\\
  \sum_{i=0}^{k-1} \binom{h-1}{i} + \sum_{i=0}^{k} \binom{h-1}{i} = \binom{h-1}{0} + \sum_{i = 1}^k \binom{h-1}{i-1} + \binom{h-1}{i} =\\
             \binom{h-1}{0} + \sum_{i = 1}^k \binom{h}{i} = \sum_{i=0}^{k} \binom{h}{i} = \alpha(k,h).
\end{multline*}
\Qed
\end{prf}

\begin{examp}\label{exp:extt}
  Consider the following labelled binary tree $T$:
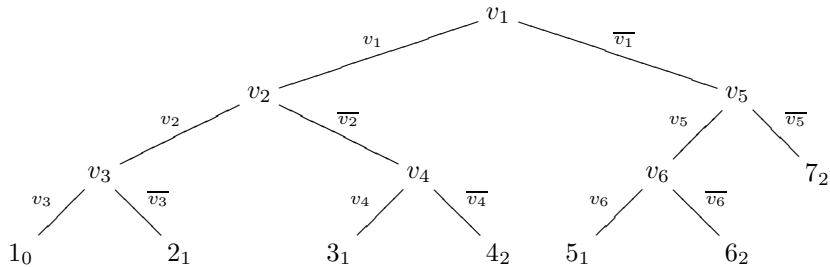
  \begin{displaymath}
    \xygraph{
      !{0;/r7ex/:}
        []{v_1} (
          - [dlll]{v_2}_{v_1} (
            -[dll]{v_3}_{v_2} (
              -[dl]{1_0}_{v_3} (),
              -[dr]{2_1}^{\ol{v_3}} ()
            ),
            -[drr]{v_4}^{\ol{v_2}} (
              -[dl]{3_1}_{v_4} (),
              -[dr]{4_2}^{\ol{v_4}} ()
            )
          ),
          - [drrr]{v_5}^{\ol{v_1}} (
            -[dl]{v_6}_{v_5} (
              -[dl]{5_1}_{v_6} (),
              -[dr]{6_2}^{\ol{v_6}} ()
            ),
            -[dr]{7_2}^{\ol{v_5}} ()
          )
        )
    }
  \end{displaymath}
  Applying the recursive construction/characterisation we see $T \in \exhst{2}{3}$. By simple counting we see that $T$ has $7$ leaves, in agreement with Lemma \ref{lem:numexleaves}, i.e.,  $\sum_{j=0}^2 \binom{3}{j} = \binom{3}{0} + \binom{3}{1} + \binom{3}{2} = 1 + 3 + 3 = 7$. Assuming that of the two subtrees at an inner node, the left subtree has Horton-Strahler numbers as least as big as the right subtree, the idea is that the sum runs over the \emph{number $j$ of right turns in a path from the root to the leaves}. In the above tree $T$, the number of right turns is indicated as an index to the leaf-name. If the Horton-Strahler number is $k$, with at most $k$ right-turns we must be able to reach every leaf.
\end{examp}

We summarise the additional knowledge over Theorem \ref{thm:sumdsmuo} (using additionally that most leaves of $T \in \exhst kh$ have depth precisely $h$):
\begin{lem}\label{lem:sizeex}
  Consider an allowed parameter pair $(k, h)$ and $T \in \exhst kh$, and let $F := \smuo(T)$.
  \begin{enumerate}
  \item $n(\doping(F)) = 2\cdot\alpha(k,h) - 1$ ($ = \Theta(h^k)$ for fixed $k$).
  \item $c(\doping(F)) = \alpha(k,h)$ ($ = \Theta(h^k)$ for fixed $k$).
  \item $\ell(\doping(F)) \le h \cdot \alpha(k,h)$ ($ = \Theta(h^{k+1})$ for fixed $k$).
  \item $\doping(F) \in \Urefc_k \sm \Urefc_{k-1}$ (for $k \ge 1$).
  \end{enumerate}
\end{lem}
In Theorem \ref{thm:separation} we will see that these $\doping(F)$ from Lemma \ref{lem:sizeex} do not have short equivalent clause-sets of hardness $k-1$. A simple example demonstrates the separation between $\Urefc_0$ and $\Urefc_1$ (similar to \cite{Val1994UnitResolutionComplete}, Example 2, which uses Example 6.1 from \cite{KeanTsiknis1990IncrementalPrimeImp}):
\begin{examp}\label{exp:sepUC01}
  The strongest separation is obtained by using $F_h := \doping(\smuo(T))$ for $T \in \exhst 1h$ and $h \in \NN$:
  \begin{enumerate}
  \item $\smuo(T)$, when considering all possible $T$, covers precisely the saturated minimally unsatisfiable renamable Horn clause-set with $h$ variables, which is up to isomorphism equal to $\set{\set{v_1},\set{\ol{v_1},v_2},\dots,\set{\ol{v_1},\dots,\ol{v_{h-1}},v_h}, \set{\ol{v_1},\dots,\ol{v_h}}}$. By Lemma \ref{lem:characsmuhts1} these are precisely those $F \in \Smusati{\delta=1}$ with $n(F) \ge 1$ which contain a full clause.
  \item $n(F_h) = 2h + 1$, $c(F_h) = h+1$, and $\hardness(F_h) = 1$.
  \item $\abs{\primec_0(F_h)} = 2^{h+1}-1$.
  \end{enumerate}
  Considering $G_n := \set{\set{v_1},\dots,\set{v_n},\set{\ol{v_1},\dots,\ol{v_n}}}$ for $n \ge 2$ and $F_n := \doping(G_n)$ we obtain an example similar (but simpler) to Example 6.1 from \cite{KeanTsiknis1990IncrementalPrimeImp}:
  \begin{enumerate}
  \item $n(G_n) = n$ and $c(G_n) = n+1$.
  \item $G_n \in \Musati{\delta=1} \sm \Smusati{\delta=1}$. The above clause-sets $\smuo(T)$ are obtained precisely as saturations of the $G_n$ (due to Lemma \ref{lem:characsmuhts1}; a saturation adds literal occurrences until we obtain a saturated minimally unsatisfiable clause-set).
  \item $\mps(G_n)$ consists precisely of the subsets of $G_n$ containing the negative clause, plus the singleton-subsets given by the unit-clauses.
  \item Thus $\abs{\mps(G_n)} = 2^n + n$.
  \item $n(F_n) = 2n+1$, $c(F_n) = n+1$, and $\hardness(F_n) = 1$.
  \item $\abs{\primec_0(F_n)} = 2^n + n$.
  \end{enumerate}
\end{examp}

\subsection{The exponential lower bound}
\label{sec:hierbadrep}

The task is to find many disjoint hyperedges in $\trighyp{F^k_h}{k}$, where $F^k_h := \doping(\smuo(T))$ for $T \in \exhst {k+1}h$. Our method for this is to show that there are many ``incomparable'' subsets of leaves in $T$ in the following sense. The \emph{depth} of a node $w$ in a rooted tree $T$, denoted by $\bmm{\depth_T(w)} \in \NNZ$, is the length of the path from the root of $T$ to $w$. Recall that two sets $A, B$ are \emph{incomparable} iff $A \not\sse B$ and $B \not\sse A$. Furthermore we call two sets $A, B$ \emph{incomparable on a set $C$} if the sets $A \cap C$ and $B \cap C$ are incomparable.
\begin{defi}\label{def:depthkincomp}
  Consider a full binary tree $T$, where every leaf has depth at least $k+1$. Consider furthermore $\es \subset V, V' \sse \lvs(T)$. Then $V$ and $V'$ are \textbf{depth-\bmm{k}-incomparable for \bmm{T}} if $V$ and $V'$ are incomparable on $\lvs(T_w)$ for all $w \in \nds(T)$ with $\depth_T(w) = k$.
\end{defi}
Note that for all allowed parameter pairs $(k,h)$ and $T \in \exhst kh$ every leaf has depth at least $k$.

\begin{lem}\label{lem:depthksephyperedge}
  Consider $k \in \NNZ$, $T \in \Tsmuo$, and $\es \not= V_0, V_1 \sse \lvs(T)$ which are depth-$k$-incomparable for $T$. Let $F := \smuo(T)$ and consider $\trighyp Fk$ (recall Definition \ref{def:triggerhypergraph}).  Then the hyperedges $E^k_{C_{V_0}}$, $E^k_{C_{V_1}}$ are disjoint (recall Definition \ref{def:clausesDSMU1}).
\end{lem}
\begin{prf}
Assume that $E^k_{C_{V_0}}$, $E^k_{C_{V_1}}$ are not disjoint; thus there is $\es \not= V \sse \lvs(T)$ with $C_V \in E^k_{C_{V_0}} \cap E^k_{C_{V_1}}$. We will show that there is $\ve \in \set{0,1}$ with $\abs{C_V \sm C_{V_\ve}} \ge k+1$, which contradicts the definition of $\trighyp Fk$.

Since $V \not= \es$, there is $w \in V$. Consider the first $k+1$ nodes $w_1,\dots,w_{k+1}$ on the path from the root to $w$. Let $w_i'$ be the child of $w_{i-1}$ different from $w_i$ for $i \in \tb 2{k+1}$, and let $T_i := T_{w_{i+1}'}$ for $i \in \tb 1k$, while $T_{k+1} := T_{w_{k+1}}$; see Figure \ref{fig:depthksep}. We show that each of $T_1, \dots, T_{k+1}$ contributes at least two unique literals to $\abs{C_V \sm C_{V_0}} + \abs{C_V \sm C_{V_1}}$, so that we get $\abs{C_V \sm C_{V_0}} + \abs{C_V \sm C_{V_1}} \ge (k+1) \cdot 2$, from which follows that there is $\ve \in \set{0,1}$ with $\abs{C_V \sm C_{V_\ve}} \ge k+1$ as claimed.

\begin{figure}[h]
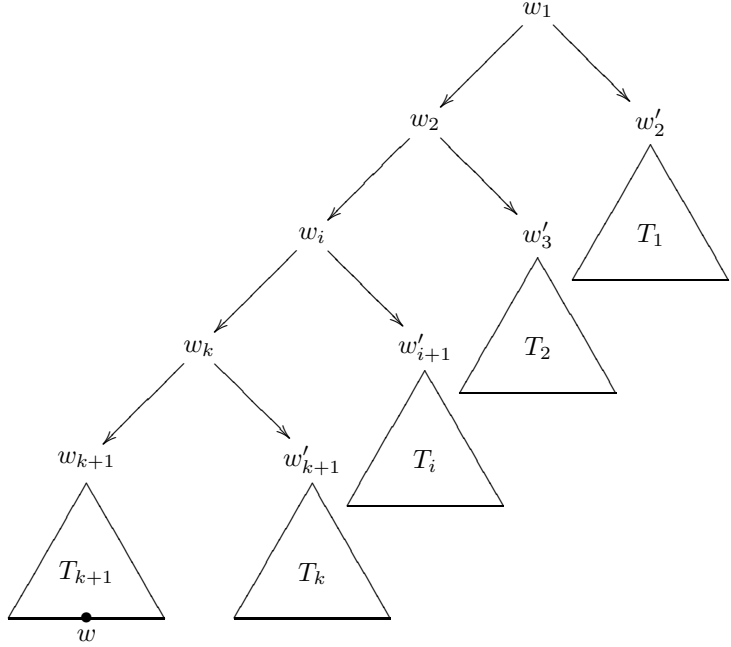

  \[\xygraph{
    !{(-4,-5.5 )}*+{}
    !{0;/r1.5cm/:}
    !{(0,0) }*+{w_1}="w1"
    !{(-1,-1) }*+{w_2}="w2"
    !{(1,-1) }*+{w_2'}="w1p"
    !{(-2,-2) }*+{w_i}="wi"
    !{(0,-2) }*+{w_3'}="w2p"
    !{(-3,-3)}*+{w_k}="wk"
    !{(-1,-3) }*+{w_{i+1}'}="wip"
    !{(-4,-4) }*+{w_{k+1}}="wkpp"
    !{(-2,-4) }*+{w_{k+1}'}="wkp"
    !{(-2,-2.5) }*+{\xypolygon3"tkpd"{~:{(0.8,0):}}}="wkpptree"
    !{(-1,-2.5) }*+{\xypolygon3"tkd"{~:{(0.8,0):}}}="wkptree"
    !{(-0.5,-2.0) }*+{\xypolygon3"tid"{~:{(0.8,0):}}}="wiptree"
    !{(0,-1.5) }*+{\xypolygon3"t2d"{~:{(0.8,0):}}}="w2ptree"
    !{(0.5,-1) }*+{\xypolygon3"t1d"{~:{(0.8,0):}}}="w1ptree"
    !{(-4,-5) }*+{T_{k+1}}="tkp"
    !{(-2,-5) }*+{T_k}="tk"
    !{(-1,-4) }*+{T_i}="ti"
    !{(0,-3) }*+{T_2}="t2"
    !{(1,-2) }*+{T_1}="t1"
    !{(-4,-5.4)}*+{\bullet}="wb"
    !{(-4,-5.55)}*+{w}="w"
    "w1" : "w2"
    "w1" : "w1p"
    "w2" : "wi"
    "w2" : "w2p"
    "wi" : "wk"
    "wi" : "wip"
    "wk" : "wkpp"
    "wk" : "wkp"
    }\]
  \caption{Illustration of sub-trees $T_1,\dots,T_{k+1}$.}
  \label{fig:depthksep}
\end{figure}

Due to the depth-k-incomparability of $V, V'$, for each $i \in \tb 1{k+1}$ and each $\ve \in \set{0,1}$ there are nodes $v_i^{\ve}$ with $v_i^{\ve} \in (\lvs(T_i) \cap V_{\ve}) \sm V_{\ol{\ve}}$. We have two cases now:
\begin{enumerate}
\item[I] If $v_i^{\ve} \in V$, then $u_{v_i^{\ve}} \in C_V \sm C_{V_{\ol{\ve}}}$.
\item[II] If $v_i^{\ve} \notin V$, then consider the first node $v$ on the path from $v_i^{\ve}$ to the root such that for the other child $v'$ of $v$, not on that path to the root, holds $\lvs(T_{v'}) \cap V \not= \es$: now for the literal $x$ labelling the edge from $v$ to $v'$ we have $x \in C_V \sm C_{V_{\ve}}$. Note that $v$ is below or equal to $w_i$ (due to $w \in V$).
\end{enumerate}
For each $\ve \in \set{0,1}$, the literals collected in $C_V \sm C_{V_{\ve}}$ from these $k+1$ sources do not coincide, due to the pairwise node-disjointness of the trees $T_1, \dots, T_{k+1}$. \Qed
\end{prf}

\begin{thm}\label{thm:nogoodksoft}
  Consider $k \in \NNZ$, $h \ge k+1$, and $T \in \exhst {k+1}h$; let $F := \doping(\smuo(T))$ and $m := \exstrahlersize{1}{h-k} = 1 + h - k$. We have
  \begin{displaymath}
    \nu(\trighyp{F}{k}) \ge \binom{m}{\floor{\frac{m}{2}}} > \frac{1}{\sqrt{2}} \frac{2^m}{\sqrt{m}} = \Theta(\frac{2^h}{\sqrt{h}}),
  \end{displaymath}
  where the second inequality assumes $h \ge k+5$, while the $\Theta$-estimation assumes fixed $k$.
\end{thm}
\begin{prf}
  For every $S \sse \pot(\lvs(T))$ with $\es \notin S$, such that every two different elements of $S$ are depth-$k$-incomparable for $T$, we have $\nu(\trighyp{F}{k}) \ge \abs{S}$ by Lemma \ref{lem:depthksephyperedge}. We can actually determine the maximal size of such an $S$, which is $M := \binom{m}{m'}$, where $m' := \floor{\frac{m}{2}}$, as follows. Let $\TT := \set{T_w : w \in \nds(T) \und \depth_T(w) = k}$; note that for $T', T'' \in \TT$ with $T' \not= T''$ we have $\lvs(T') \cap \lvs(T'') = \es$. Choose $T_0 \in \TT$ with minimal $\nlvs(T_0)$; by Lemma \ref{lem:numexleaves} we have $\nlvs(T_0) = m$. Let $S_0 := \set{V \cap \lvs(T_0) : V \in S}$. Then $S_0$ is an antichain (i.e., the elements of $S_0$ are pairwise incomparable) and $\abs{S_0} = \abs{S}$. By Sperner's Theorem (\cite{Sperner1928SubsetsFiniteSets}) holds $\abs{S_0} \le M$, and this upper bound $M$ is realised, just observing the antichain-condition, by choosing for $S_0$ the set $\binom{\lvs(T_0)}{m'}$ of subsets of $\lvs(T_0)$ of size $m'$. This construction of $S_0$ can be extended to a construction of $S$ (of the same size) by choosing for each $T' \in \TT$ an injection $j_{T'}: S_0 \ra \binom{\lvs(T')}{m'}$ and defining $S := \set{\bc_{T' \in \TT} j_{T'}(V)}_{V \in S_0}$. The given estimation of $M$ follows from Stirling's approximation. \Qed
\end{prf}

We are now able to state the main result of this article, proving Conjecture 1.1 from \cite{GwynneKullmann2012Slur,GwynneKullmann2012SlurJ} that $\Urefc_k$, and indeed also $\Wrefc_k$, is a proper hierarchy of boolean functions regarding polysize representations without new variables (see Subsection \ref{sec:cnfrep} for a discussion of ``representations'' in general):
\begin{thm}\label{thm:separation}
  Consider $k \in \NNZ$. For $h \ge k+1$ choose one $T_h \in \exhst {k+1}h$ (note there is up to left-right swaps exactly one element in $\exhst {k+1}h$), and let $F_h := \doping(T_h)$. Consider the sequence $(F_h)_{h \ge k+1}$.
  \begin{enumerate}
  \item By Lemma \ref{lem:sizeex} we have $n(F_h) = \Theta(h^{k+1})$ as well as $c(F_h) = \Theta(h^{k+1})$, and $F_h \in \Urefc_{k+1}$.
  \item Consider a sequence $(F_h')_{h \ge k+1}$ of clause-sets with $F_h'$ equivalent to $F_h$, such that $F_h' \in \Wrefc_k$. By Theorems \ref{thm:nogoodksoft}, \ref{thm:triggersetmethod} we have $c(F_h') = \Omega(\frac{2^h}{\sqrt{h}})$.
  \end{enumerate}
\end{thm}

We conjecture that Theorem \ref{thm:separation} can be strengthened by including the PC-hierarchy in the following way:
\begin{conj}\label{con:sepsharp}
  For every $k \in \NNZ$ there exists a sequence $(F_n)_{n \in \NN}$ of clause-sets in $\Propc_{k+1}$, where for convenience we assume $n(F_n) = n$ for all $n$, such that $(\ell(F_n))_{n \in \NN}$ is polynomially bounded, and such that for every sequence $(F_n')_{n \in \NN}$ in $\Wrefc_k$, where for all $n \in \NN$ holds that $F_n'$ is equivalent to $F_n$, the sequence $(\ell(F_n'))_{n \in \NN}$ is not polynomially bounded.
\end{conj}

\section{Analysing the Tseitin translation}
\label{sec:analT}

We now turn to upper bounds, investigating cases where the Tseitin translation yields representations in $\Urefc$. We consider two main cases: translating a DNF into a CNF, or translating an XOR-circuit. In Subsection \ref{sec:cnfrep} we discuss the general notion of ``CNF representation''. In Subsection \ref{sec:cantrans} we discuss translating DNF into CNF, which we consider as a map from $\Cls$ to $\Cls$, and which we call the ``canonical translation''. Lemma \ref{lem:exphddnf} shows that the hardness of canonical translation results can be arbitrarily high. On the other hand, Lemma \ref{lem:hitct} shows that for hitting DNF the canonical translation result is in $\Urefc$, and Theorem \ref{thm:extuc} applies this to our lower bound examples, in contrast to Theorem \ref{thm:separation} (so we see that new variables here help). Finally by using only the necessary direction of the equivalences in the Tseitin translation, in Lemma \ref{lem:hdctm} we see that for this ``reduced canonical translation'' the result is always in $\Urefc$. We conclude by discussing representations of XOR-clause-sets in Subsection \ref{sec:xorclauses}.

\subsection{CNF-representations}
\label{sec:cnfrep}

In Subsections 1.4 and 9.2 of \cite{GwynneKullmann2012Slur,GwynneKullmann2012SlurJ} we discussed representations of boolean functions in general. The most general notion useful in the SAT-context seems to allow existentially quantified new variables, which yields the following basic definition:
\begin{defi}\label{def:cnfrep}
  A \textbf{CNF-representation} of $F \in \Cls$ (as CNF) is a clause-set $F' \in \Cls$ with $\var(F) \sse \var(F')$ such that the satisfying assignments of $F'$ (as CNF) projected to $\var(F)$ are precisely the satisfying assignments of $F$.
\end{defi}

\begin{examp}\label{exp:cnfrep}
  Consider $F := \set{\set{a,b}}$. Then $F' := F \cup \set{\set{v,a}}$ is a CNF-representation of $F$, since the satisfying assignments of $F$ can be extended to satisfying assignments of $F'$ by assigning $v \ra 1$, while no new satisfying assignments are present, since $F'$ is a superset of $F$. Also $F \cup \set{\set{v}}$ is a CNF-representation of $F$, but $F \cup \set{\set{a}}$ is not, since the satisfying assignment $\pab{a \ra 0, b \ra 1}$ of $F$ would be lost. Also $\set{\set{v,a,b}}$ is not a CNF-representation of $F$, since here now we would obtain a new satisfying assignment for $F$, namely $\pab{a, b \ra 0}$.
\end{examp}
The CNF-representations $F'$ of $F$ \emph{without new variables}, that is, with $\var(F') = \var(F)$, are precisely the clause-sets $F'$ equivalent to $F$ with $\var(F') = \var(F)$. We have conjectured in \cite{GwynneKullmann2012Slur,GwynneKullmann2012SlurJ} (Conjecture 9.4) that Theorem \ref{thm:separation} (and Conjecture \ref{con:sepsharp}) also holds when allowing new variables, which in this context we can rephrase as follows, also extending the conjecture by including $\Wrefc_k$ (see Conjecture \ref{con:newvnhhd} for a further strengthening):
\begin{conj}\label{con:sepext}
  For every $k \in \NNZ$ there exists a sequence $(F_n)_{n \in \NN}$ of clause-sets, such that there is a sequence $(F'_n)_{n \in \NN}$, where
  \begin{itemize}
  \item each $F'_n$ is a CNF-representation of $F_n$,
  \item $\ell(F'_n)$ is polynomial in $n$,
  \item and we have $F'_n \in \Propc_{k+1}$,
  \end{itemize}
  but where there is no such sequence $(F''_n)_{n \in \NN}$ with $F''_n \in \Wrefc_k$.
\end{conj}

Our basic condition for a ``good'' representation $F'$ of $F \in \Cls$ is that $F' \in \Urefc_k$ holds for some ``low'' $k$ (a constant if $F$ depends on parameters). This is what we call the \textbf{absolute condition} --- regarding the requirement of detecting unsatisfiability of $\vp * F'$ for some partial assignment $\vp$ we do not distinguish between original variables (those in $\var(F)$) and new variables (those in $\var(F') \sm \var(F)$), that is, $\var(\vp) \sse \var(F')$ is considered. If we consider only $\var(\vp) \sse \var(F)$, then we obtain the \textbf{relative condition}:
\begin{defi}\label{def:relhd}
  For $F \in \Cls$ and $V \sse \Va$ the \textbf{relative hardness} $\bmm{\hardness^V(F)} \in \NNZ$ is defined as the minimum $k \in \NNZ$ such that for all partial assignments $\vp \in \Pass$ with $\var(\vp) \sse V$ and $\vp * F \in \Usat$ we have $\rk_k(\vp * F) = \set{\bot}$. And the  \textbf{relative w-hardness} $\bmm{\whardness^V(F)} \in \NNZ$ is defined as the minimum $k \in \NNZ$ such that for all partial assignments $\vp \in \Pass$ with $\var(\vp) \sse V$ and $\vp * F \in \Usat$ we have that $k$-resolution derives $\bot$ from $\vp * F$.
\end{defi}
Obviously $\hardness^V(F) \le \hardness(F)$ and $\hardness^{\var(F)}(F) = \hardness(F)$, as well as $\whardness^V(F) \le \whardness(F)$ and $\whardness^{\var(F)}(F) = \whardness(F)$. Having a representation $F'$ of $F$ with $\hardness^{\var(F)}(F') \le 1$ is closely related to what is typically called ``maintaining arc consistency''; it would be precisely that if we would use p-hardness instead of hardness, while using (only) hardness is a certain weakening. Having $\hardness^{\var(F)}(F') = 0$ here is equivalent to $\primec_0(F) \sse F'$, and thus for hardness $0$ new variables are not helpful, neither for the relative nor the absolute condition.

Conjecture \ref{con:sepext} is false for relative hardness, since regarding relative hardness the hierarchy collapses to the first level: we will present the details in a future paper, but they are not difficult --- since there are no conditions on the new variables, the $\rk_k$-computations for $k > 1$ can be encoded into CNF, only relying on $\rk_1$. Such an encoding is an extension of Theorem 1 in \cite{BKNW2009CircuitComplexity}, using similar techniques. More involved is the collapse of the $\Wrefc_k$-hierarchy to the first level regarding relative hardness; we believe we can also show this, but we better formulate it explicitly as a conjecture:
\begin{conj}\label{con:collapseWrefc}
  For every $k \ge 1$ there is a polytime function $t(F,V)$, which takes a clause-set $F$ and a finite set $V$ of variables as arguments, such that in case of $\whardness^V(F) \le k$ the output $t(F,V)$ is a representation of $F$ with $\whardness^V(t(F)) \le 1$.
\end{conj}
Note that for all $F \in \Cls$ and $V \sse \Va$ holds $\whardness^V(F) \le 1 \Lra \hardness^V(F) \le 1$. The collapse of all considered hierarchies to their first level, when considering the relative condition, is for us a major argument in favour of the absolute condition: Within the class of representations of relative hardness at most $1$ (when using new variables) there is a lot of structure, and many representations fulfil absolute conditions; some basic examples follow in the remainder of this section.

\subsection{The canonical translation}
\label{sec:cantrans}

If for the $F \in \Cls$ to be represented we have an equivalent DNF $G \in \Cls$, then we can apply the Tseitin translation, using one new variable $v$ to express one DNF-clause, i.e., using $\primec_0(v \lra \bw_{x \in C} x)$ for $C \in G$. The details are as follows.

We assume that an injection $\vcan: \set{(F,C) \mb F \in \Cls \und C \in F}  \ra \Va$ is given, yielding the \ul{v}ariables of the \ul{c}anonical \ul{t}ranslation, such that these variables are new for $F$, that is, $\var(F) \cap \set{\vcan(F,C)}_{C \in F} = \es$ holds for all $F \in \Cls$. We write $\bmm{\vcan_F^C} := \vcan(F,C)$.

\begin{defi}\label{def:ct}
  The map $\bmm{\cant}: \Cls \ra \Cls$ is defined for $F \in \Cls$ as
  \begin{multline*}
    \cant(F) := \setb{\set{\ol{\vcan_F^C}, x} : C \in F \und x \in C} \cup \setb{\set{\vcan_F^C} \cup \ol{C} : C \in F} \cup \\
    \setb{\set{\vcan_F^C}_{C \in F}}.
  \end{multline*}
\end{defi}
The first two types of clauses are the prime implicates of the boolean functions $\vcan_F^C \lra \bw_{x \in C} x$, while the last type (a long, single clause) says that one of the (DNF-)clauses from $F$ must be true. To emphasise: the map $\cant$ is a map from clause-sets to clause-sets, where the (implicit) interpretation of the input and the output is different: the input $F \in \Cls$ is interpreted as DNF, while the output $\cant(F) \in \Cls$ is interpreted as CNF. Some basic properties of the canonical translation:
\begin{enumerate}
\item The basic measures of the canonical translation for $F \in \Cls$ are given by
  \begin{enumerate}
  \item $n(\cant(F)) = n(F) + c(F)$
  \item $c(\cant(F)) = 1 + c(F) + \ell(F)$ for $F \not= \set{\bot}$.
  \item $\ell(\cant(F)) = 2 c(F) + 3 \ell(F)$ for $F \not= \set{\bot}$.
  \end{enumerate}
\item $\cant(\top) = \set{\bot}$ and $\cant(\set{\bot}) = \set{\set{\vcan_{\set{\bot}}^{\bot}}}$.
\item Consider $\vp \in \Pass$ with $\var(\vp) \sse \var(F)$, and treat $F$ as a multi-clause-set, that is, if application of $\vp$ to different non-satisfied clauses from $F$ makes these clauses equal, then no contractions are performed. Then the canonical translation behaves homomorphic regarding application of partial assignments in the sense that $\cant(\ol{\vp} *F)$ (recall that we need to treat $F$ here as a DNF) is isomorphic to $(\vp \cup \psi) * \cant(F)$, where $\psi$ sets those $\vcan_F^C$ to $0$ for which there is $x \in C$ with $\vp(x) = 0$.
\end{enumerate}

\begin{examp}\label{exp:cant}
  We give some simple examples for canonical translations.
  \begin{enumerate}
  \item\label{exp:cant1} For $F :=  \set{\underbrace{\set{v_1}}_{\bmm{C_1}}, \bot}$ we have
    \begin{displaymath}
      \cant(F) = \set{ \underbrace{\set{\ol{\vcan_F^{C_1}},v_1}, \set{\vcan_F^{C_1},\ol{v_1}}}_{\bmm{v_1 \lra \vcan_F^{C_1}}}, \underbrace{\set{\vcan_F^{\bot}}}_{\bmm{1 \lra \vcan_F^{\bot}}}, \underbrace{\set{\vcan_F^{C_1},\vcan_F^{\bot}}}_{\bmm{\vcan_F^{C_1} \oder \vcan_F^\bot}}}.
    \end{displaymath}
  \item \label{exp:cant2} For $F := \set{ \underbrace{\set{v_1,v_2,v_3}}_{\bmm{C_1}}, \underbrace{\set{v_1,v_2,v_4}}_{\bmm{C_2}}}$ we have
    \begin{multline*}
      \cant(F) = \set{
        \underbrace{\set{\ol{\vcan_F^{C_1}},v_1}, \set{\ol{\vcan_F^{C_1}}, v_2}, \set{\ol{\vcan_F^{C_1}}, v_3}, \set{\vcan_F^{C_1}, \ol{v_1}, \ol{v_2}, \ol{v_3}}}_{\bmm{(v_1 \und v_2 \und v_3) \lra \vcan_F^{C_1}}},\\
        \underbrace{\set{\ol{\vcan_F^{C_2}},v_1}, \set{\ol{\vcan_F^{C_2}}, v_2}, \set{\ol{\vcan_F^{C_2}}, v_4}, \set{\vcan_F^{C_2}, \ol{v_1},\ol{v_2},\ol{v_4}}}_{\bmm{(v_1 \und v_2 \und v_4) \lra \vcan_F^{C_2}}},
        \underbrace{\set{\vcan_F^{C_1}, \vcan_F^{C_2}}}_{\bmm{\vcan_F^{C_1} \oder \vcan_F^{C_2}}} }.
    \end{multline*}
  \item\label{exp:cant3} Applying $\vp := \pab{v_3 \ra 1, v_4 \ra 1}$ to the last example (Case \ref{exp:cant2}) yields
    \begin{multline*}
      \vp * \cant(F) = \set{
        \underbrace{\set{\ol{\vcan_F^{C_1}},v_1}, \set{\ol{\vcan_F^{C_1}}, v_2}, \set{\vcan_F^{C_1}, \ol{v_1}, \ol{v_2}}}_{\bmm{(v_1 \und v_2) \lra \vcan_F^{C_1}}},\\
        \underbrace{\set{\ol{\vcan_F^{C_2}},v_1}, \set{\ol{\vcan_F^{C_2}}, v_2}, \set{\vcan_F^{C_2}, \ol{v_1},\ol{v_2}}}_{\bmm{(v_1 \und v_2) \lra \vcan_F^{C_2}}}, \underbrace{\set{\vcan_F^{C_1}, \vcan_F^{C_2}}}_{\bmm{\vcan_F^{C_1} \oder \vcan_F^{C_2}}} }.
    \end{multline*}
  \item\label{exp:cant4} Applying $\vp := \pab{v_3 \ra 0}$ to Case \ref{exp:cant2} yields
    \begin{multline*}
      \vp * \cant(F) = \set{
        \underbrace{\set{\ol{\vcan_F^{C_1}},v_1}, \set{\ol{\vcan_F^{C_1}}, v_2}, \set{\ol{\vcan_F^{C_1}}}}_{\bmm{\ol{\vcan_F^{C_1}}}},\\
        \underbrace{\set{\ol{\vcan_F^{C_2}},v_1}, \set{\ol{\vcan_F^{C_2}}, v_2}, \set{\ol{\vcan_F^{C_2}}, v_4}, \set{\vcan_F^{C_2}, \ol{v_1},\ol{v_2},\ol{v_4}}}_{\bmm{(v_1 \und v_2 \und v_4) \lra \vcan_F^{C_2}}},
        \underbrace{\set{\vcan_F^{C_1}, \vcan_F^{C_2}}}_{\bmm{\vcan_F^{C_1} \oder \vcan_F^{C_2}}} }.
    \end{multline*}
  \item\label{exp:cant5} While applying $\vp := \pab{v_3 \ra 0}$ and $\psi := \set{\vcan_F^{C_1} \ra 0}$ to Case \ref{exp:cant2} yields
    \begin{multline*}
      (\vp \cup \psi) * \cant(F) =\\
      \set{
        \underbrace{\set{\ol{\vcan_F^{C_2}},v_1}, \set{\ol{\vcan_F^{C_2}}, v_2}, \set{\ol{\vcan_F^{C_2}}, v_4}, \set{\vcan_F^{C_2}, \ol{v_1},\ol{v_2},\ol{v_4}}}_{\bmm{(v_1 \und v_2 \und v_4) \lra \vcan_F^{C_2}}},
        \underbrace{\set{\vcan_F^{C_2}}}_{\bmm{\vcan_F^{C_2}}} }.
    \end{multline*}
  \end{enumerate}
  In Case \ref{exp:cant3} we see an example of why for the canonical translation to have the homomorphism property we must consider $F$ as a multi-clause-set. That is, $\ol{\vp} * F = \set{\set{v_1,v_2}}$, and so $\vp * \cant(F) \not= \cant(\ol{\vp} * F)$: the clause $\set{v_1, v_2}$ is represented by two separate new variables in $\vp * \cant(F)$ compared to only one in $\cant(\ol{\vp} * F)$.

  In Case \ref{exp:cant4} we see an example where for the homomorphism property of the canonical translation not just renaming, but also some unit-clause elimination is needed. These unit-clauses are added in Case \ref{exp:cant5}, extending the assignment to falsify the new variable $\vcan_F^{C_1}$ corresponding to falsified DNF-clause $C_1$.
\end{examp}

\begin{lem}\label{lem:ctrep}
  Consider $F \in \Cls$ (as CNF) and an equivalent DNF-clause-set $G \in \Cls$. Then $\cant(G)$ is a CNF-representation of $F$.
\end{lem}
\begin{prf}
  $\cant(F)$ is true iff at least one of its $\vcan$-variables is set to true, which is precisely the case iff at least one of DNF-clauses of $G$ is satisfied, where the (DNF-)clauses of $G$ cover precisely the satisfying assignments of $F$. \Qed
\end{prf}

\begin{lem}\label{lem:relhdct}
  For $F \in \Cls$ we have $\hardness^{\var(F)}(\cant(F)) \le 1$ (recall Definition \ref{def:relhd}).
\end{lem}
\begin{prf}
  Consider $\vp \in \Pass$ with $\var(\vp) \sse \var(F)$ and $\vp * \cant(F) \in \Usat$. Then all DNF-clauses of $F$ are falsified, which yields via UCP that all $\vcan$-variables are set to false, and thus $\rk_1(\vp * \cant(F)) = \set{\bot}$. \Qed
\end{prf}
In \cite{BarahomaJungKatsirelosWalsh2008EncodingDNNF} a more general version of Lemma \ref{lem:relhdct} is proven, showing that for all ``smooth'' DNNFs (Disjoint Negation Normal Form) the Tseitin translation yields a clause-set which maintains arc-consistency via UCP (a somewhat stronger property than relative hardness $\le 1$ as in Lemma \ref{lem:relhdct}).\footnote{There is a mistake in \cite{BarahomaJungKatsirelosWalsh2008EncodingDNNF} in that it claims that the Tseitin translation of \emph{all} DNNFs maintain arc-consistency via UCP, however this is shown only for smooth DNNFs as confirmed by George Katirelos via e-mail in January 2012.} That Lemma \ref{lem:relhdct} only establishes the relative condition, and not the absolute one, is due to the fact that setting $\vcan$-variables to $0$ can pose arbitrarily hard conditions; a concrete example follows, while a more drastic general construction is given in Lemma \ref{lem:exphddnf}. However the difficulties can be overcome, by just removing them: In Lemma \ref{lem:hdctm} we will see that when dropping the part of the canonical translation which gives meaning to setting $\vcan$-variables to $0$, that then we actually can establish the absolute condition.

\begin{examp}\label{exp:relhdct}
  Consider the following clause-set with variables $x_1, \dots, x_5$:
  \begin{displaymath}
    F := \setb{ \underbrace{\set{x_1, x_2, x_3}}_{\bmm{C_1}}, \underbrace{\set{x_1, x_2, x_4}}_{\bmm{C_2}}, \underbrace{\set{x_1,x_2,x_5}}_{\bmm{C_3}} }.
  \end{displaymath}
  The canonical translation is
  \begin{eqnarray*}
    \cant(F) &=& \underbrace{\set{\set{x_1,\ol{\vcan_F^{C_1}}}, \set{x_2,\ol{\vcan_F^{C_1}}}, \set{x_3,\ol{\vcan_F^{C_1}}}}, \set{\ol{x_1}, \ol{x_2}, \ol{x_3}, \vcan_F^{C_1}}}_{\bmm{\vcan_F^{C_1} \lra (x_1 \und x_2 \und x_3)}} \cup\\
             &  & \underbrace{\set{\set{x_1,\ol{\vcan_F^{C_2}}}, \set{x_2,\ol{\vcan_F^{C_2}}}, \set{x_4,\ol{\vcan_F^{C_2}}}}, \set{\ol{x_1}, \ol{x_2}, \ol{x_4}, \vcan_F^{C_2}}}_{\bmm{\vcan_F^{C_2} \lra (x_1 \und x_2 \und x_4)}} \cup\\
             &  & \underbrace{\set{\set{x_1,\ol{\vcan_F^{C_3}}}, \set{x_2,\ol{\vcan_F^{C_3}}}, \set{x_5,\ol{\vcan_F^{C_3}}}}, \set{\ol{x_1}, \ol{x_2}, \ol{x_5}, \vcan_F^{C_3}}}_{\bmm{\vcan_F^{C_3} \lra (x_1 \und x_2 \und x_5)}} \cup\\
             &  & \underbrace{\set{\set{\vcan_F^{C_1}, \vcan_F^{C_2}, \vcan_F^{C_3}}}}_{\bmm{(\vcan_F^{C_1} \oder \vcan_F^{C_2} \oder \vcan_F^{C_3})}}.
  \end{eqnarray*}
  Applying the partial assignment $\vp := \pab{x_3 \ra 1, x_4 \ra 1, x_5 \ra 1, \vcan_F^{C_3} \ra 0}$ yields\vspace{-2ex}
  \begin{eqnarray*}
    F' := \vp * \cant(F) &=& \underbrace{\set{\set{x_1,\ol{\vcan_F^{C_1}}}, \set{x_2,\ol{\vcan_F^{C_1}}}}, \set{\ol{x_1}, \ol{x_2}, \vcan_F^{C_1}}}_{\bmm{\vcan_F^{C_1} \lra (x_1 \und x_2)}} \cup\\
             &  & \underbrace{\set{\set{x_1,\ol{\vcan_F^{C_2}}}, \set{x_2,\ol{\vcan_F^{C_2}}}}, \set{\ol{x_1}, \ol{x_2}, \vcan_F^{C_2}}}_{\bmm{\vcan_F^{C_2} \lra (x_1 \und x_2)}} \cup\\
             &  & \underbrace{\set{\set{\ol{x_1}, \ol{x_2}}}}_{\bmm{\neg (x_1 \und x_2)}} \cup \underbrace{\set{\set{\vcan_F^{C_1}, \vcan_F^{C_2}}}}_{\bmm{(\vcan_F^{C_1} \oder \vcan_F^{C_2})}}.
  \end{eqnarray*}
  We have $F' \in \Usat$, where $F'$ has no unit-clauses, whence $\hardness(F') \ge 2$, and so $\cant(F) \notin \Urefc_1$.
\end{examp}

For general input-DNFs, the hardness of the canonical translation can be arbitrary high:
\begin{lem}\label{lem:exphddnf}
  Consider $F \in \Cls$. Let $v \in \Va \sm \var(F)$ and $F' := F \cup \set{\set{v}}$. Then $\hardness(\cant(F')) \ge \hardness(F)$.
\end{lem}
\begin{prf}
  Let $\vp := \pab{\vcan_{F'}^C \ra 0 : C \in F} \cup \pab{v, \vcan_{F'}^{\set{v}} \ra 1}$. Then $\vp * \cant(F') = F'' := \set{\ol{C} : C \in F}$, where $\hardness(\cant(F')) \ge \hardness(F'') = \hardness(F)$. \Qed
\end{prf}

If we do not have just a DNF, but a ``disjoint'' or ``orthogonal'' DNF (see Section 1.6 and Chapter 7 in \cite{CramaHammer2011BooleanFunctions}), which are as clause-sets precisely the hitting clause-sets, then we obtain absolute hardness $1$:
\begin{lem}\label{lem:hitct}
  For $F \in \Clash$ we have $\cant(F) \in \Urefc$, where $\cant(F)$ is a representation of the DNF-clause-set $F$.
\end{lem}
\begin{prf}
  Consider a partial assignment $\vp$ such that $\vp * \cant(F)$ is unsatisfiable. Since $\Clash$ is stable under application of partial assignments, and furthermore here no contractions take place, w.l.o.g.\ we can assume that $\var(\vp) \cap \var(F) = \es$. If $\vp$ sets two or more $\vcan$-variables to true, then UCP yields a contradiction, since any two clauses from $F$ clash. If $\vp$ would set precisely one $\vcan$-variable to true, then we had $\vp * \cant(F) = \top$. So assume that $\vp$ sets no $\vcan$-variable to true. Now $\vp$ must set all $\vcan$-variables to false, since, as already mentioned, just setting one $\vcan$-variable to true satisfies $\cant(F)$. And thus $\bot \in \cant(F)$. \Qed
\end{prf}

We now want to show that via the canonical translation we can obtain representations of $\doping(F)$ for $F \in \Uclash$. For this we show first that all such $\doping(F)$ have short hitting DNF clause-sets. For $F \in \Cls$ let $\bmm{\nsat(F)} \in \NNZ$ denote the number of satisfying assignments for $F$, that is, $\nsat(F) = \abs{\Dnf(F)}$.
\begin{lem}\label{lem:exstrahlerdnf}
  Consider $F \in \Uclash$, and let $m := n(F) + c(F)$.
  \begin{enumerate}
  \item\label{lem:exstrahlerdnf1} $\nsat(\doping(F)) = 2^{m-1}$.
  \item\label{lem:exstrahlerdnf2} Let $F' := \setb{ \ol{C} \cup \set{u_C} \mb C \in F}$; by definition we have $F' \in \Clash$. Furthermore $\nsat(F') = 2^{m-1}$.
  \item\label{lem:exstrahlerdnf3} $F'$ as a DNF-clause-set is equivalent to the CNF-clause-set $\doping(F)$.
  \end{enumerate}
\end{lem}
\begin{prf}
We have $\sum_{C \in F} 2^{-\abs{C}} = 1$ (see \cite{Kullmann2007HandbuchMU}). Thus $\sum_{C \in \doping(F)} 2^{-\abs{C}} = \frac 12$, which proves Part \ref{lem:exstrahlerdnf1} (note $m = n(\doping(F))$ and $\doping(F) \in \Clash$). Part \ref{lem:exstrahlerdnf2} follows from Part \ref{lem:exstrahlerdnf1}, since $F'$ results from $\doping(F)$ by flipping literals. Finally we consider Part \ref{lem:exstrahlerdnf3}. All elements of $F'$, as DNF-clauses (i.e., conjunctions of literals), represent satisfying assignments for $\doping(F)$, that is, for all $C \in F'$ and $D \in \doping(F)$ we have $C \cap D \not= \es$. By Part \ref{lem:exstrahlerdnf2}, precisely half of the total assignments of DNF-clause-set $F'$ are falsifying, and thus precisely half of the total assignments are satisfying: since this is the same number as the satisfying assignments of $\doping(F)$, we obtain that the DNF-clause-set $F'$ is equivalent to the CNF-clause-set $F$. \Qed
\end{prf}

An alternative line of argumentation is that for $F \in \Uclash$ the (logical) negation of $\doping(F)$ (as a CNF) is given by $\doping(F)'$, which is obtained from $\doping(F)$ by flipping all doping literals, i.e., replacing clauses $C \cup \set{u_C}$ by $C \cup \set{\ol{u_C}}$. That this is indeed the negation, follows from the two facts, that $\doping(F) \cup \doping(F)' \in \Clash$ by definition, and that $\doping(F) \cup \doping(F)'$ results from $F$ by replacing each clause $C$ with the two clauses $C \cup \set{u_C}, C \cup \set{\ol{u_C}}$, which are together equivalent to $C$.

By Lemma \ref{lem:exstrahlerdnf} and Lemma \ref{lem:hitct} we obtain now that doped unsatisfiable hitting clause-sets have good representations via the canonical translation:
\begin{thm}\label{thm:extuc}
  For $F \in \Uclash$ there is a short CNF-representation (using new variables) of $\doping(F)$ in $\Urefc$, namely $F' := \cant(\set{\ol{C} \cup \set{u_C} : C \in F}) \in \Urefc$, where:
  \begin{enumerate}
  \item $n(F') = n(F) + 2 c(F)$.
  \item $c(F') = 1 + 2 c(F) + \ell(F)$.
  \end{enumerate}
  This applies especially for $F \in \Smusati{\delta=1} \subset \Uclash$.
\end{thm}

Finally we show that when relaxing the canonical translation, using only the necessary direction of the constitutive equivalences, then we actually obtain representations in $\Urefc$ for every DNF-clause-set:
\begin{defi}\label{def:ctm}
  The map $\bmm{\cantm}: \Cls \ra \Cls$ (``reduced canonical translation'') is defined for $F \in \Cls$ as $\cantm(F) := \set{\set{\ol{\vcan_F^C}, x} : C \in F \und x \in C} \cup \set{\set{\vcan_F^C}_{C \in F}}$.
\end{defi}
Note that all clauses of $\cantm(F)$ are binary except of the long clause stating that one of the $\vcan$-variables must become true. And also note that in case of $\bot \notin F$ the additional clauses of $\cant(F)$, that is, the $C \in \cant(F) \sm \cantm(F)$, are all blocked for $\cant(F)$ (see \cite{Ku97b,Ku96c}), since $C$ can not be resolved on the $\vcan$-variable in it. We have $\var(\cantm(F)) = \var(\cant(F))$, while the basic measure for $F \in \Cls$ are given as follows:
\begin{enumerate}
\item $n(\cantm(F)) = n(F)+c(F)$
\item $c(\cantm(F)) = 1 + \ell(F)$
\item $\ell(\cantm(F)) = c(F) + 2 \ell(F)$.
\end{enumerate}
With the same proof as Lemma \ref{lem:ctrep} we get:
\begin{lem}\label{lem:ctmrep}
  Consider $F \in \Cls$ (as CNF) and an equivalent DNF-clause-set $G \in \Cls$. Then $\cantm(G)$ is a CNF-representation of $F$.
\end{lem}

We show now that dropping the additional (blocked) clauses, present in the full form $\cant(F)$, actually leads to the hardness dropping to $1$ for arbitrary input-DNFs (recall Lemma \ref{lem:exphddnf}), exploiting that now there are less possibilities for making $\cantm(F)$ unsatisfiable by instantiation:
\begin{lem}\label{lem:hdctm}
  For $F \in \Cls$ we have $\cantm(F) \in \Urefc$ (i.e., $\cantm: \Cls \ra \Urefc$).
\end{lem}
\begin{prf}
  For the sake of contradiction consider a partial assignment $\vp$ such that $F' := \rk_1(\vp * \cantm(F)) \in \Usat$ but $F' \not= \set{\bot}$. Note that $F'$ contains neither $\bot$ nor a unit-clause, and thus $F'$ is a subset of $\cantm(F)$ except of the possibly shortened or satisfied long $\vcan$-clause. If $F'$ contains no new variables, then thus $F' = \top$, a contradiction. So there exists some $C \in F$ such that $\vcan_F^C$ occurs in $F'$. Consider the assignment $\vp'$, which sets $\vcan_F^C$ and all $x \in C$ to true, while setting all other (remaining) new variables to false: $\vp'$ satisfies $F'$, a contradiction. \Qed
\end{prf}

\begin{examp}\label{exp:uep}
  We conclude our basic considerations of ``canonical translations'' by discussing ``unique extension properties''. A representation $F'$ of $F$ has the \emph{unique extension property} (``uep'') if for every total satisfying assignment of $F$ there is exactly one extension to a satisfying assignment of $F'$. For every $F \in \Cls$ the representation $\cant(F)$ of $F$ has the uep, since a variable $\vcan_F^C$ must be set to $1$ precisely for those $C \in F$ which are satisfied by $\vp$ in the DNF-sense (i.e., $\ol{\vp} * \set{C} = \set{\bot}$). On the other hand, the representation $\cantm(F)$ of $F$ in general has not the uep: The total satisfying assignments for $\cant(F)$ extending $\vp$ are exactly those which set \emph{at least} one of the variables $\vcan_F^C$ true for those $C \in F$ which are satisfied in the DNF-sense.

  A representation $F'$ of $F$ has the \emph{strong unique extension property} if for every partial assignment $\vp$ with $\bot \in \ol{\vp} * F$ (i.e., $\vp$ satisfied at least one of the DNF-clauses) there is exactly one extension on the new variables (alone) to a satisfying assignment of $F'$. For $F \in \Clash$ the representation $\cant(F)$ of $F$ has the strong uep, since the satisfying assignments given by the clauses of $F$ are inconsistent with each other.
\end{examp}

\subsection{XOR-clauses}
\label{sec:xorclauses}

For the $n$-bit parity function $x_1 \oplus \dots \oplus x_n = 0$ the unique equivalent clause-set $\primec_0(x_1 \oplus \dots \oplus x_n = 0)$ (unique since the prime implicates are not resolvable) has $2^{n-1}$ clauses. We now show that a typical SAT translation of the $n$-bit parity function, using new variables $y_i$ (for $i \in \tb2{n-1}$) to compute the xor of the first $i$ bits, is in $\Urefc$.
\begin{lem}\label{lem:1softxor}
  Consider $n \ge 3$, literals $x_1,\dots,x_n$ with different underlying variables, and new variables $y_2, \dots,y_{n-1}$. Let $F := P_2 \cup \left(\bc_{i = 3}^{n-1} P_i \right) \cup P_n$, where
  \begin{enumerate}
  \item $P_2 := \primec_0(x_1 \oplus x_2 = y_2) = \set{\set{\ol{x_1},x_2,y_2}, \set{x_1,\ol{x_2},y_2}, \set{x_1,x_2,\ol{y_2}}, \set{\ol{x_1},\ol{x_2},\ol{y_2}}}$
  \item $P_i := \primec_0(y_{i-1} \oplus x_i = y_i) = \set{\set{\ol{y_{i-1}},x_i,y_i}, \set{y_{i-1},\ol{x_i}, y_i}, \set{y_{i-1},x_i,\ol{y_i}}, \\ \set{\ol{y_{i-1}},\ol{x_i},\ol{y_i}}}$
  \item $P_n := \primec_0(y_{n-1} \oplus x_n = 0) = \set{\set{y_{n-1},\ol{x_n}}, \set{\ol{y_{n-1}},x_n}}$.
  \end{enumerate}
  We have $F \in \Urefc$, and $F$ represents $\primec_0(x_1 \oplus \dots \oplus x_n = 0)$.
\end{lem}
\begin{prf}
  Assume for the sake of contradiction that $F \notin \Urefc$. Thus there exists a partial assignment $\vp$ such that for $F' := \rk_1(\vp * F)$ we have $F' \in \Usat$, but $F' \not= \set{\bot}$. By definition $F'$ has no clauses of size $\le 1$ and is non-empty. Observe that setting any variable in $P_i$ for $i \in \tb2{n-1}$ yields a pair of binary clauses representing an equivalence or anti-equivalence between the two remaining variables. Also if $P_i \cap F' \not= \es$ for some $i \in \tb2{n-1}$, then we have $P_i \sse F'$, since all clauses of $P_i$ contain all variables of $P_i$. Therefore we have $F' = E \cup \bc_{i \in I} P_i$ for some subset $I \sse \tb2{n-1}$, where $E$ is a set of clauses representing a chain of equalities and inequalities. Consider the assignment $\vp' := \pab{x_i \ra 0 : i \in I}$. We have $\vp' * P_i = \vp' * \primec_0(y_{i-1} + x_i = y_i) = \primec_0(y_{i-1} = y_i)$; note that $x_i$ is \emph{only} in $P_i$, and so $\pao{x_i}{1}$ only touches $P_i$. So $\vp' * F'$ now contains only variable-disjoint chains of equivalences and anti-equivalences, each trivially satisfiable, yielding a contradiction. \Qed
\end{prf}
\begin{examp}\label{exp:xortrans}
  For $n = 3$ we get
  \begin{displaymath}
    F = \setb{ \underbrace{\set{x_1,x_2,\ol{y_2}}, \set{x_1,\ol{x_2},y_2}, \set{\ol{x_1},x_2, y_2}, \set{\ol{x_1},\ol{x_2},\ol{y_2}}}_{\bmm{x_1 \oplus x_2 = y_2}},\underbrace{\set{y_2,\ol{x_3}}, \set{\ol{y_2},x_3} }_{\bmm{y_2 \oplus x_3 = 0}} }.
  \end{displaymath}
\end{examp}

A very interesting question is how much the (simple) Lemma \ref{lem:1softxor} can be extended, towards representing arbitrary systems of linear equations. It seems to us, that we do not have polysize representations with bounded hardness in the UC-framework:
\begin{conj}\label{con:xorcls}
  As usual, an ``XOR-clause'' is a (boolean) constraint of the form $x_1 \oplus \dots \oplus x_n = 0$ for literals $x_1, \dots, x_n$, which we just represent by the clause $\set{x_1,\dots,x_n} \in \Cl$. An ``XOR-clause-set'' $F$ is a set of XOR-clauses, which is just represented by an ordinary clause-set $F \in \Cls$ (with an alternative interpretation). The conjecture now is that XOR-clause-sets do not have good representations with bounded hardness, not even when using relative hardness. That is, there is no $k \in \NNZ$ and no polynomial $p(x)$ such that for all clause-sets $F \in \Cls$ there exists a CNF-representation $F' \in \Cls$ (possibly using new variables), taking $F$ as an XOR-clause-set, with $\ell(F') \le p(\ell(F))$ and $\hardness^{\var(F)}(F') \le k$.
\end{conj}
Basic results for showing such a lower bound are obtained in \cite{BKNW2009CircuitComplexity}. As we have already remarked (after Definition \ref{def:relhd}), regarding relative hardness only $k \in \set{0,1}$ are of relevance (because we allow new variables), while regarding absolute hardness we conjecture that also with new variables we have a proper hierarchy (Conjecture \ref{con:sepext}).

 We conclude now our initial study on ``good representations'' by the basic observations regarding the naive approach for translating XOR-clause-sets.

\section{Hardness under union}
\label{sec:hdunion}

When applied piecewise to a system of linear equations (with different auxiliary variables for each single equation), the translation from Lemma \ref{lem:1softxor} does not yield a clause-set in $\Urefc$, as we show in Theorem \ref{thm:2xor}. To facilitate the precise computation of the hardness of the union of two such XOR-clause-translations, we present two general tools for upper bounds on hardness and one for lower bounds.

\begin{lem}\label{lem:hdpassub}
  Consider $F \in \Cls$ and $V \sse \var(F)$. Let $P$ be the set of partial assignments $\psi$ with $\var(\psi) = V$. Then $\hardness(F) \le \abs{V} + \max_{\psi \in P} \hardness(\psi * F)$.
\end{lem}
\begin{prf}
Consider a partial assignment $\vp$ with $\vp * F \in \Usat$; we have to show $\hardness(\vp * F) \le \abs{V} + \max_{\psi \in P} \hardness(\psi * F)$. Build a resolution refutation of $\vp * F$ by first creating a splitting tree (possibly degenerated) on the variables of $V$; this splitting tree (a perfect binary tree) has height $\abs{V}$, and at each of its leaves we have a clause-set $\vp * (\psi * F)$ for some appropriate $\psi \in P$. Thus at each leaf we can attach a splitting tree of Horton-Strahler number of hardness at most $\max_{\psi \in P} \hardness(\psi * F)$, and from that (via the well-known correspondence of splitting trees and resolution trees; see \cite{Ku99b,Ku00g} for details) we obtain a resolution tree fulfilling the desired hardness bound. \Qed
\end{prf}

We obtain an upper bound on the hardness of the union of two clause-sets:
\begin{corol}\label{cor:hdunion}
  For $F_1, F_2 \in \Cls$ holds $\hardness(F_1 \cup F_2) \le \max(\hardness(F_1), \hardness(F_2)) + \abs{\var(F_1) \cap \var(F_2)}$.
\end{corol}
\begin{prf}
  Apply Lemma \ref{lem:hdpassub} with $F := F_1 \cup F_2$ and $V := \var(F_1) \cap \var(F_2)$, and apply the general upper bound $\hardness(F_1 \cup F_2) \le \max(\hardness(F_1), \hardness(F_2))$ for variable-disjoint $F_1, F_2$ (Lemma 15 in \cite{GwynneKullmann2012SlurSOFSEM}). \Qed
\end{prf}

Substitution of literals can not increase (w-)hardness:
\begin{lem}\label{lem:eqhardness}
  Consider a clause-set $F \in \Usat$ and (arbitrary) literals $x,y$. Denote by $F_{x \la y} \in \Usat$ the result of replacing $x$ by $y$ and $\ol{x}$ by $\ol{y}$ in $F$, followed by removing clauses containing complementary literals. Then we have $\hardness(F_{x \la y}) \le \hardness(F)$ and $\whardness(F_{x \la y}) \le \whardness(F)$.
\end{lem}
\begin{prf}
Consider $T: F \vdash \bot$. It is a well-known fact (and a simply exercise), that the substitution of $y$ into $x$ can be performed in $T$, obtaining $T_{x \la y}: F_{x \la y} \vdash \bot$. This is easiest to see by performing first the substitution with $T$ itself, obtaining a tree $T'$ which as a binary tree is identical to $T$, using ``pseudo-clauses'' with (possibly) complementary literals; the resolution rule for sets $C, D$ of literals with $x \in C$ and $\ol{x} \in D$ allows to derive the clause $(C \sm \set{x}) \cup (D \sm \set{\ol{x}})$, where the resolution-variables are taken over from $T$. Now ``tautological'' clauses (containing complementary literals) can be cut off from $T'$: from the root (labelled with $\bot$) go to a first resolution step where the resolvent is non-tautological, while one of the parent clauses is tautological (note that not both parent clauses can be tautological) --- the subtree with the tautological clause can now be cut off, obtaining a new pseudo-resolution tree where clauses only got (possibly) shorter (see Lemma 6.1, part 1, in \cite{Ku00g}). Repeating this process we obtain $T_{x \la y}$ as required. Obviously $\hts(T_{x \la y}) \le \hts(T)$, and if in $T$ for every resolution step at least one of the parent clauses has length at most $k$ for some fixed (otherwise arbitrary) $k \in \NNZ$, then this also holds for $T_{x \la y}$. \Qed
\end{prf}

\begin{examp}\label{exp:substhd}
  The simplest example showing that for satisfiable clause-sets $F$ (w-)hardness can be increased by substitution is given by $F := \set{\set{x},\set{\ol{y}}}$ for $\var(x) \not= \var(y)$. Here $\hardness(F) = 0$, while $F_{x \la y} = \set{\set{y},\set{\ol{y}}}$, and thus $\hardness(F_{x \la y}) = 1$.
\end{examp}

Now we are ready to determine the (high) hardness of the union of the (piecewise) translation of two XOR-clauses for a basic special case:
\begin{thm}\label{thm:2xor}
  For $n \ge 3$ consider the system
  \begin{eqnarray*}
    x_1 \oplus x_2 \oplus \dots \oplus x_n &=& 0 \\
    x_1 \oplus x_2 \oplus \dots \oplus \ol{x_n} &=& 0.
  \end{eqnarray*}
  Let $F := F_1 \cup F_2$, where $F_1$ is the translation of the first equation by Lemma \ref{lem:1softxor}, and $F_2$ is the translation of the second equation, using different auxiliary variables (so $n(F) = 2 \cdot (n + (n-2)) - n = 3n - 4$). We have $F \in \Usat$ with $\hardness(F) = n$.
\end{thm}
\begin{prf}
From Corollary \ref{cor:hdunion} and Lemma \ref{lem:1softxor} we obtain $\hardness(F) \le n+1$. Better is to apply Lemma \ref{lem:hdpassub} with $V := \var(\set{x_2,\dots,x_{n-1}})$. By definition we see that $\psi * F \in \Pcls{2}$ (i.e., all clauses have length at most two) for $\psi$ with $\var(\psi) = V$. By Lemma 19 in \cite{GwynneKullmann2012SlurSOFSEM} we have $\hardness(\psi * F) \le 2$, and thus $\hardness(F) \le (n-2) + 2 = n$. The lower bound is obtained by an application of Lemma \ref{lem:lbhd}. Consider any literal $x \in \lit(F_n)$, where the subscript in $F_n = F$ makes explicit the dependency on $n$. Setting $x$ to true or false results either in an equivalence or in an anti-equivalence. Propagating this (anti-)equivalence yields a clause-set $F'$ isomorphic to $F_{n-1}$, where by Lemma \ref{lem:eqhardness} this propagation does not increase hardness, so we have $\hardness(\pao{x}{1} * F_n) \ge \hardness(F') = \hardness(F_{n-1})$. The argumentation can be trivially extended for $n \in \set{0,1,2}$, and so by Lemma \ref{lem:lbhd} we get $\hardness(F) \ge n$. \Qed
\end{prf}
If $F_1, F_2$ in Theorem \ref{thm:2xor} were the direct translations (with $\hardness(F_1) = \hardness(F_2) = 0$), then $\hardness(F) = n$ would follow easily with Lemma 3.18 in \cite{Ku99b}, since then $F$ would be simply the clause-set with all $2^n$ clauses of length $n$. Of course, regarding a good translation of the system from Theorem \ref{thm:2xor} we can just use $\set{\bot}$, easily computed by preprocessing --- however the content of Conjecture \ref{con:xorcls} is, that no preprocessing is powerful enough to handle arbitrary (satisfiable!) systems of linear equations (over the two-element field).

\section{Basic experiments}
\label{sec:experiments}

In this section we perform some experiments on the use of the three different mechanisms for representing boolean functions $f$ studied in this article:
\begin{enumerate}
\item clause-sets $F$ equivalent to $f$ with $F \in \Urefc_k$ where $k$ is as low as feasible;
\item the canonical translation $\cant(G)$ for a DNF-clause-set $G$ equivalent to $f$;
\item and the reduced canonical translation $\cantm(G)$.
\end{enumerate}
The instances are described in Subsection \ref{sec:expinst}, while the experimental results are discussed in Subsection \ref{sec:expsolver}. Our focus is on gaining a better understanding of the interaction between solver behaviour and problem representation, and so we consider various representative complete SAT solvers.

\subsection{The instances}
\label{sec:expinst}

For our experiments we want to take a boolean function $f_{k,h}$ as a constraint in a bigger SAT problem $G_{k,h}$. The ``optimal'' equivalent representation $F_{k,h}$ of $f_{k,h}$ shall have hardness $k$, and $f_{k,h}$ should also have a small equivalent DNF, so that the canonical and reduced canonical translation are available.

For $f_{k,h}$ we take the boolean functions from Theorem \ref{thm:nogoodksoft}, which have the short CNF's (without new variables) $F_{k,h} := \doping(\smuo(T_{k,h}))$ for $k \ge 2$ and $h \ge k+1$, where $T_{k,h} \in \exhst {k}h$. So $F_{k,h}$ has hardness $k$, while every equivalent clause-set of hardness at most $k-1$ contains at least $b(m) := \binom{m}{\floor{\frac{m}{2}}}$ many clauses for $m := h - k$.

For the ``completion'' to $G_{k,h}$ let $F'_{k,h}$ be the negation of $F_{k,h}$ according to the remarks to Lemma \ref{lem:exstrahlerdnf}, that is, $F'_{k,h}$ is obtained from $F_{k,h}$ by complementing the doping literals in each clause. Let $\ol{F} := \set{\ol{C} : C \in F}$ for $F \in \Cls$. Note that $\ol{F'_{k,h}}$ is the DNF for $F_{k,h}$. We define $G_{k,h}^i$ for $i = 1,2,3$ as always including $F'_{k,h}$, and additionally
\begin{enumerate}
\item $G_{k,h}^1$ uses $F_{k,h}$, i.e., $G_{k,h}^1 := F'_{k,h} \cup F_{k,h}$.
\item $G_{k,h}^2$ uses the canonical translation according to Theorem \ref{thm:extuc} (and Lemma \ref{lem:exstrahlerdnf}), i.e., $G_{k,h}^2 := F'_{k,h} \cup \cant(\ol{F'_{k,h}})$.
\item $G_{k,h}^3$ uses the reduced canonical translation according to Lemma \ref{lem:hdctm} (and Lemma \ref{lem:exstrahlerdnf}), i.e., $G_{k,h}^3 := F'_{k,h} \cup \cantm(\ol{F'_{k,h}})$,
\end{enumerate}

The sizes of the $G_{k,h}^i$ are determined as follows:
\begin{itemize}
\item By Lemma \ref{lem:sizeex} we have $c(F_{k,h}) = \alpha(k,h)$, while $n(F_{k,h}) = 2 c(F_{k,h})-1 = 2 \alpha(k,h) - 1$.
\item The size of $F'_{k,h}$ is precisely the same.
\item So $n(G_{k,h}^1) = 2 \alpha(k,h) - 1$ and $c(G_{k,h}^1) = 2 \alpha(k,h)$, while $\ell(G_{k,h}^1) = 2 \ell(F_{k,h})$.
\item $n(G_{k,h}^2) = n(G_{k,h}^3) = 3 \alpha(k,h) - 1$.
\item $c(G_{k,h}^3) = 1 + \alpha(k,h) + \ell(F_{k,h})$ and $\ell(G_{k,h}^3) = \alpha(k,h) + 3 \ell(F_{k,h})$.
\item $c(G_{k,h}^2) = 1 + 2 \alpha(k,h) + \ell(F_{k,h})$ and $\ell(G_{k,h}^3) = 2 \alpha(k,h) + 4 \ell(F_{k,h})$.
\end{itemize}
See Figure \ref{fig:inststat} for the numerical data. The lower bounds $b(h-k)$ there for the number of clauses in any clause-set $F$ equivalent to $F_{k,h}$ and with $F \in \Wrefc_{k-1}$ show that these representations are infeasible here. As an amusing fact one can note here that the number of clauses in $F \in \Wrefc_0$ would be (precisely) $2^c - 1$, which even for the smallest example considered is a rather astronomical number. We can determine the hardness of the $G_{k,h}^i$ precisely; first an auxiliary lemma:
\begin{lem}\label{lem:hdunioncomplct}
  For $F \in \Cls$ and $F' \in \set{\cant(\ol{F}), \cantm(\ol{F})}$ holds $F \cup F' \in \Usat$ and $\hardness(F \cup F') \le 2$.
\end{lem}
\begin{prf}
  Due to $\cantm(\ol{F}) \sse \cant(\ol{F})$ w.l.o.g.\ $F' = \cantm(\ol{F})$, since $\Urefc_2 \cap \Usat$ is closed under formation of super-clause-sets by Lemma 6.7 in \cite{GwynneKullmann2012SlurJ}. For all $C \in F$ and $x \in \ol{C}$ the binary clause $\vcan^{\ol{C}} \ra x$ is in $F'$. Thus setting $\vcan^{\ol{C}}$ to $1$ in $F \cup F'$ results in setting $x$ to $1$ via $\rk_1$, which altogether falsifies $C \in F$. Thus $\rk_2$ applied to $F \cup F'$ sets all $\vcan^{\ol{C}}$ to $0$, which falsifies $\set{\vcan^{\ol{C}} : C \in F} \in F'$. \Qed
\end{prf}

Now the (total) hardness of the unsatisfiable SAT problems $G_{k,h}$ is as follows:
\begin{lem}\label{lem:hduhitex}
  Consider $k, h \in \NN$ with $k \ge 2$ and $h \ge k + 1$. We have:
  \begin{enumerate}
  \item $\hardness(G_{k,h}^1) = k+1$.
  \item $\hardness(G_{k,h}^2) = \hardness(G_{k,h}^3) = 2$.
  \end{enumerate}
\end{lem}
\begin{prf}
  $\hardness(G_{k,h}^1) = k+1$ follows from the fact, that by definition $G_{k,h}^1 \in \Smusat_{\delta=1}$ holds, where the corresponding tree $T := \tsmuo(G_{k,j}^1)$ has Horton-Strahler number $k+1$ (recall Lemma \ref{lem:hdsmu1}): $T$ is obtained from the underlying $T_{k,h}$ by replacing each leaf with the full binary tree with three nodes. $G^2_{k,h}$ and $G^3_{k,h}$ have hardness at least $2$ since they are unsatisfiable and contain no unit-clauses. The remaining assertions follow by Lemma \ref{lem:hdunioncomplct}. \Qed
\end{prf}

\subsection{Solver performances}
\label{sec:expsolver}

For the experiments we used a 64-bit workstation with 32 GB RAM and Intel i5-2320 CPUs (6144 KB cache) running with 3 GHz, where we only employed a single CPU. To emphasise again, the aim of these experiments is to obtain a qualitative picture of the behaviour of a range of contemporary SAT solvers, and not to find out which solver is ``fastest''. For our experimentation we use the following solvers:
\begin{enumerate}
\item \OKsolver{} (\cite{Ku2002h}): a look-ahead solver, used as a ``theoretical'' solver.
\item \texttt{MiniSat}-family:
  \begin{enumerate}
  \item \texttt{MiniSat}, version 2.2 (see \cite{Sorensson2010Minisat22}).
  \item \texttt{CryptoMiniSat}, version 2.9.6 (see \cite{Soos2010CMSDesc}).
  \item \texttt{Glucose}, version 2.0 (see \cite{AudemardSimon2009PredictingLearnt}).
  \end{enumerate}
\item \texttt{Lingeling}-family:
  \begin{enumerate}
  \item \texttt{PicoSAT}, version 913 (see\cite{Biere2008picosat,Biere2010PicoSATLingelingDesc}).
  \item \texttt{PrecoSAT}, version 570 (\cite{Biere2009Precosat}).
  \item \texttt{Lingeling}, version \texttt{ala-b02aa1a-121013} (see \cite{Biere2010PicoSATLingelingDesc,Biere2012LingelingDesc}).
  \end{enumerate}
\end{enumerate}

First we consider the \OKsolver{} (see Figure \ref{fig:inststat}), as a look-ahead solver (see \cite{HvM09HBSAT,Kullmann2007HandbuchTau} for the general concepts), as well as a solver with a ``clean'' behaviour, due to the minimisation of the use of heuristical shortcuts. For example, the \OKsolver{} seems to be the only SAT-solver computing $\rk_2$, while all other solvers (recall the discussion in Subsection \ref{sec:introrelSAT}) only test selected literals for failed literals. We see that the $G_{k,h}^1$ are far easier than the $G_{k,h}^{2,3}$, although they require branching. Indeed, the straightforward heuristics choosing a variable occurring most often will find a backtracking tree of optimal, i.e., linear size (note that all $F \in \Smusati{\delta=1}$ have exactly one variable occurring in every clause, and splitting on this variable creates two variable-disjoint instances). In conformance with this, linear regression shows with high correlation that the instances $G_{k,h}^1$ are solved by the \OKsolver{} in linear time, i.e., $O(\ell)$. Considering now $G_{k,h}^{2,3}$, recall that by Lemma \ref{lem:hduhitex} all these instances have hardness $2$, that is, they can be solved without branching. And in fact the number of $\rk_2$-reductions of the \OKsolver{} for these instances is precisely $c(F_{k,h})-1$, in accordance with Lemma \ref{lem:hdunioncomplct}. The worst-case running time for $\rk_2$ is $O(n^2 \cdot \ell)$, but in this case going once through the list of all variables is sufficient to find the contradiction. Again in conformance with this, linear regression shows with high correlation that the instances $G_{k,h}^{2,3}$ are solved in time $O(n \cdot \ell)$. We note here that the \OKsolver{} is actually the fastest solver on these instances, for all three types, though this is not the focus of these experiments.

Other look-ahead solvers performed badly on these instances. \texttt{satz} performs very badly even on the very easy ones: on $G_{2,22}^1$ it needed $4.3$ sec, and on $G_{2,32}^1$ already $2$ hours. Thus it was not considered further. \texttt{march\_pl} performed somewhat better, but was also not able to complete even the easier instances $G_{k,h}^1$; furthermore it crashed on various instances, and was thus also not considered further.

\begin{figure}[H]
  \tabcolsep 2pt
  \small
  \centering
  \begin{tabular}{|c|c|c||l|l|l||l|l||l|l|l|}
    \cline{4-11}
    \multicolumn{3}{c|}{} & \multicolumn{5}{c||}{Instance statistics} & \multicolumn{3}{c|}{\OKsolver{} statistics}\\
    \cline{1-11}
    $k$ & $h$ & $i$ &  $n$       & $c$        & $\ell$     & $\alpha(k,h)$ & $b(h-k)$  & $t$ (sec) & nds & $\rk_2$\\
    \hline\hline
    2   & 22  & 1   &  507       & 508        & 8604      & 254           & $1.8 \cdot 10^6$    & 0.0     & 43     & 232   \\
        &     & 2   &  761       & 4811      & 17716     &               &                     & 0.0     & 1      & 253   \\
        &     & 3   &  761       & 4557      & 13160     &               &                     & 0.0     & 1      & 253   \\
        & 32  & 1   &  1057     & 1058      & 24994     & 529           & $1.6 \cdot 10^9$    & 0.0     & 63     & 497   \\
        &     & 2   &  1586     & 13556     & 51046     &               &                     & 0.0     & 1      & 528   \\
        &     & 3   &  1586     & 13027     & 38020     &               &                     & 0.0     & 1      & 528   \\
        & 42  & 1   &  1807     & 1808      & 54784     & 904           & $1.4 \cdot 10^{12}$  & 0.0     & 83     & 862   \\
        &     & 2   &  2711     & 29201     & 111,376    &               &                     & 0.1     & 1      & 903   \\
        &     & 3   &  2711     & 28297     & 83080     &               &                     & 0.1     & 1      & 903   \\
        & 52  & 1   &  2757     & 2758      & 101,974    & 1379         & $1.3 \cdot 10^{15}$  & 0.1     & 103    & 1327 \\
        &     & 2   &  4136     & 53746     & 206,706    &               &                     & 0.4     & 1      & 1378 \\
        &     & 3   &  4136     & 52367     & 154,340    &               &                     & 0.2     & 1      & 1378 \\
        & 62  & 1   &  3907     & 3908      & 170,564    & 1954         & $1.2 \cdot 10^{18}$  & 0.2     & 123    & 1892 \\
        &     & 2   &  5861     & 89191     & 345,036    &               &                     & 1.0     & 1      & 1953 \\
        &     & 3   &  5861     & 87237     & 257,800    &               &                     & 0.5     & 1      & 1953 \\
        & 72  & 1   &  5257     & 5258      & 264,554    & 2629         & $1.1 \cdot 10^{21}$  & 0.4     & 143    & 2557 \\
        &     & 2   &  7886     & 137,536    & 534,366    &               &                     & 4.0     & 1      & 2628 \\
        &     & 3   &  7886     & 134,907    & 399,460    &               &                     & 1.0     & 1      & 2628 \\
\hline
    3   & 23  & 1   &  4095     & 4096      & 80594     & 2048         & $1.8 \cdot 10^6$    & 0.0     & 507    & 1794    \\
        &     & 2   &  6143     & 44394     & 165,284    &               &                     & 0.2     & 1      & 2047    \\
        &     & 3   &  6143     & 42346     & 122,939    &               &                     & 0.1     & 1      & 2047    \\
        & 33  & 1   &  12035    & 12036     & 327,384    & 6018         & $1.6 \cdot 10^9$    & 0.2     & 1057  & 5489    \\
        &     & 2   &  18053    & 175,729    & 666,804    &               &                     & 4.8     & 1      & 6017    \\
        &     & 3   &  18053    & 169,711    & 497,094    &               &                     & 1.6     & 1      & 6017    \\
        & 43  & 1   &  26575    & 26576     & 922,524    & 13288        & $1.4 \cdot 10^{12}$  & 1.0     & 1807  & 12384   \\
        &     & 2   &  39863    & 487,839    & 1,871,624  &               &                     & 82.6    & 1      & 13287   \\
        &     & 3   &  39863    & 474,551    & 1,397,074  &               &                     & 28.9    & 1      & 13287   \\
\hline
    4   & 24  & 1   &  25901    & 25902     & 562,542    & 12951        & $1.8 \cdot 10^6$    & 0.4     & 4095   & 10903 \\
        &     & 2   &  38852    & 307,174    & 1,150,986  &               &                     & 15.4    & 1      & 12950  \\
        &     & 3   &  38852    & 294,223    & 856,764    &               &                     & 4.5     & 1      & 12950  \\
        & 34  & 1   &  105,911   & 105,912    & 3,150,408  & 52,956        & $1.6 \cdot 10^9$    & 3.3     & 12035  & 46938 \\
        &     & 2   &  158,867   & 1,681,117  & 6,406,728  &               &                     & 843.4   & 1      & 52955  \\
        &     & 3   &  158,867   & 1,628,161  & 4,778,568  &               &                     & 410.8   & 1      & 52955  \\
        & 44  & 1   &  299,971   & 299,972    & 11,326,724 & 149,986       & $1.4 \cdot 10^{12}$  & 16.5    & 26575  & 136,698 \\
        &     & 2   &  449,957   & 5,963,335  & 22,953,420 &               &                     & 10233 & 1      & 149,985  \\
        &     & 3   &  449,957   & 5,813,349  & 17,140,072 &               &                     & 5296  & 1      & 149,985  \\
\hline
    5   & 25  & 1   &  136,811   & 136,812    & 3,202,912  & 68406        & $1.8 \cdot 10^6$    & 2.7     & 25901  & 55455  \\
        &     & 2   &  205,217   & 1,738,269  & 6,542,636  &               &                     & 664.6   & 1      & 68405   \\
        &     & 3   &  205,217   & 1,669,863  & 4,872,774  &               &                     & 348.7   & 1      & 68405   \\
        & 35  & 1   &  768,335   & 768,336    & 24,413,776 & 384,168       & $1.6 \cdot 10^9$    & 31.2    & 105,911 & 331,212 \\
        &     & 2   &  1,152,503 & 12,975,225 & 49,595,888 &               &                     & 36743 & 1      & 384,167   \\
        &     & 3   &  1,152,503 & 12,591,057 & 37,004,832 &               &                     & 20062 & 1      & 384,167   \\
    \hline
  \end{tabular}
  \caption{Instance statistics for $G_{k,h}^i$, and solver statistics for the \OKsolver{} with option ``no-tree-pruning'', turning off the intelligent backtracking, which consumes too much memory for the larger instances. ``nds'' is the number of nodes of the backtracking tree, while ``$\rk_2$'' is the number of $\rk_2$-reductions $F \leadsto \rk_2 (\rk_1(\pao x1 * F))$ in case of $\rk_1(\pao x0 * F) = \set{\bot}$.}
  \label{fig:inststat}
\end{figure}

Now we turn to the conflict-driven solvers (see \cite{MSLM09HBSAT} for a general introduction), where we consider the \texttt{MiniSat}-family (Figure \ref{fig:minisatstats}) and the \texttt{Lingeling}-family (Figure \ref{fig:lingelingstats}). Considering $G_{k,h}^1$, we note that \texttt{MiniSat} as well as \texttt{PrecoSAT} always solve these instances by preprocessing. And actually \texttt{MiniSat -no-pre} (without preprocessing) solves these instances faster (by branching) than with preprocessing. While \texttt{PicoSAT}, which also does not use preprocessing, is not much slower than \texttt{PrecoSAT}. With the largest instance $G_{5,35}^1$, all solvers except of \texttt{PrecoSAT} have considerable difficulties, but all can handle it (only \texttt{PicoSAT} aborts, likely due to a bug). That \OKsolver{} is much faster here we believe is due to the fact, that in general look-ahead solvers should be better than conflict-driven solvers on unsatisfiable instances, where the shortest refutations are (close to) tree-like (and in this case the tree-like refutation of $F \in \Smusati{\delta=1}$ given by the underlying tree $\tsmuo(F)$ is the shortest possible overall).

\enlargethispage{2ex}

\begin{figure}[H]
  \tabcolsep 0.5pt
  \footnotesize
  \centering
  \hspace*{-6em}\begin{tabular}{|c|c|c||l|l|l||l|l|l||l|l|l||l|l|l|}
    \cline{4-15}
    \multicolumn{3}{c|}{} & \multicolumn{3}{c||}{\texttt{MiniSat}}  & \multicolumn{3}{c||}{\texttt{MiniSat -no-pre}}  & \multicolumn{3}{c||}{\texttt{CryptoMiniSat}} & \multicolumn{3}{c|}{\texttt{Glucose}}  \\
    \hline
    $k$ & $h$ & $i$ & $t$ (sec)       & decisions  & confl & $t$ (sec)       & decisions      & confl & $t$ (sec)  & decisions      & confl & $t$ (sec)       & decisions      & confl \\
    \hline\hline
2       & 22  & 1   & 0.0             & 0          & 0         & 0.0             & 10227          & 365       & 0.0        & 1832           & 20        & 0.0             & 11763          & 350       \\
        &     & 2   & 0.0             & 1134       & 136       & 0.0             & 6706           & 416       & 0.0        & 0              & 0         & 0.0             & 6433           & 383       \\
        &     & 3   & 0.0             & 1134       & 136       & 0.0             & 6706           & 416       & 0.0        & 0              & 0         & 0.0             & 6442           & 400       \\
        & 32  & 1   & 0.0             & 0          & 0         & 0.0             & 36646          & 795       & 0.0        & 2165           & 25        & 0.0             & 27529          & 853       \\
        &     & 2   & 0.0             & 12816      & 740       & 0.0             & 18301          & 905       & 0.0        & 0              & 0         & 0.0             & 19317          & 934       \\
        &     & 3   & 0.0             & 12816      & 740       & 0.0             & 18301          & 905       & 0.0        & 0              & 0         & 0.0             & 21288          & 929       \\
        & 42  & 1   & 0.1             & 0          & 0         & 0.0             & 133,105        & 1366      & 0.0        & 5798           & 53        & 0.0             & 113,533        & 1371      \\
        &     & 2   & 0.2             & 29334      & 1529      & 0.1             & 32008          & 1563      & 0.0        & 0              & 0         & 0.1             & 32674          & 1594      \\
        &     & 3   & 0.1             & 29334      & 1529      & 0.1             & 32008          & 1563      & 0.0        & 0              & 0         & 0.1             & 32472          & 1544      \\
        & 52  & 1   & 0.2             & 0          & 0         & 0.0             & 206,925        & 1962      & 0.1        & 12291          & 50        & 0.0             & 204,314        & 2017      \\
        &     & 2   & 0.6             & 65019      & 2778      & 0.2             & 79259          & 2496      & 0.0        & 0              & 0         & 0.2             & 44778          & 2694      \\
        &     & 3   & 0.6             & 65019      & 2778      & 0.2             & 79259          & 2496      & 0.0        & 0              & 0         & 0.2             & 44535          & 2588      \\
        & 62  & 1   & 0.5             & 0          & 0         & 0.1             & 482,733        & 2861      & 0.2        & 21874          & 55        & 0.1             & 256,891        & 3003      \\
        &     & 2   & 1.1             & 129,523    & 3887      & 0.4             & 158,975        & 3697      & 0.1        & 0              & 0         & 0.5             & 79563          & 3786      \\
        &     & 3   & 1.0             & 129,523    & 3887      & 0.4             & 158,975        & 3697      & 0.1        & 0              & 0         & 0.5             & 79604          & 3800      \\
        & 72  & 1   & 1.1             & 0          & 0         & 0.1             & 533,500        & 3963      & 0.5        & 24866          & 56        & 0.2             & 1,200,842      & 3969      \\
        &     & 2   & 2.7             & 165,596    & 5417      & 1.4             & 137,582        & 4981      & 0.1        & 0              & 0         & 1.0             & 152,734        & 6127      \\
        &     & 3   & 2.4             & 165,596    & 5417      & 1.3             & 137,582        & 4981      & 0.1        & 0              & 0         & 1.0             & 152,734        & 6127      \\
\hline
3       & 23  & 1   & 0.2             & 0          & 0         & 0.1             & 726,328        & 3344      & 0.1        & 66885          & 349       & 0.1             & 293,599        & 3642      \\
        &     & 2   & 0.1             & 34343      & 1276      & 0.5             & 87038          & 2719      & 0.1        & 0              & 0         & 0.4             & 50429          & 2601      \\
        &     & 3   & 0.1             & 34343      & 1276      & 0.5             & 84922          & 2683      & 0.1        & 0              & 0         & 0.4             & 50883          & 2626      \\
        & 33  & 1   & 2.1             & 0          & 0         & 0.6             & 6,024,786      & 10163     & 0.7        & 767,860        & 1426      & 0.5             & 2,573,291      & 9976      \\
        &     & 2   & 15.0            & 245,555    & 8333      & 10.4            & 293,488        & 8213      & 0.4        & 0              & 0         & 19.0            & 310,569        & 8770      \\
        &     & 3   & 14.8            & 244,410    & 8272      & 10.5            & 303,064        & 8217      & 0.4        & 0              & 0         & 13.0            & 264,088        & 8633      \\
        & 43  & 1   & 14.6            & 0          & 0         & 3.3             & 30,413,289     & 23355     & 4.8        & 11,673,409     & 12547     & 2.5             & 12,387,073     & 21567     \\
        &     & 2   & 132.5           & 886,834    & 20033     & 89.5            & 764,994        & 20101     & 1.0        & 0              & 0         & 98.7            & 834,345        & 22453     \\
        &     & 3   & 134.7           & 837,910    & 19939     & 91.0            & 808,130        & 19817     & 1.2        & 0              & 0         & 107.9           & 894,762        & 22331     \\
\hline
4       & 24  & 1   & 5.5             & 0          & 0         & 3.4             & 26,310,775     & 23307     & 2.5        & 10,823,044     & 14335     & 3.0             & 13,739,340     & 23265     \\
        &     & 2   & 10.1            & 351,753    & 10427     & 71.8            & 603,915        & 15761     & 1.1        & 0              & 0         & 93.4            & 746,936        & 16170     \\
        &     & 3   & 9.9             & 351,468    & 10330     & 62.2            & 603,915        & 15761     & 1.4        & 0              & 0         & 73.6            & 624,916        & 16463     \\
        & 34  & 1   & 149.5           & 0          & 0         & 62.2            & 510,575,547    & 88280     & 57.8       & 121,886,023    & 65608     & 73.1            & 404,205,131    & 92344     \\
        &     & 2   & 6381            & 3,851,979  & 72123     & 5376            & 4,762,174      & 69651     & 706.2      & 1,080,246      & 30501     & 5889            & 3,856,879      & 73007     \\
        &     & 3   & 6894            & 4,265,009  & 70250     & 4749            & 4,762,174      & 69651     & 614.8      & 1,165,228      & 30500     & 5557            & 3,857,144      & 75795     \\
        & 44  & 1   & 2117            & 0          & 0         & 475.7           & 4,225,934,440  & 232,867   & 538.6      & 1,756,703,536  & 332,497   & 539.0           & 3,658,524,320  & 287,335   \\
        &     & 2   & \textbf{A}17749 & 10,905,675 & 62092     & \textbf{A}50777 & 16,691,952     & 192,830   & 34461      & 5,708,264      & 114,958   & \textbf{A}32100 &                &           \\
        &     & 3   & \textbf{S}      &            &           & \textbf{A}31985 & 14,856,654     & 155,899   & 34850      & 4,565,988      & 105,312   & \textbf{A}31080 &                &           \\
\hline
5       & 25  & 1   & 143.3           & 0          & 0         & 74.9            & 702,026,588    & 109,898   & 76.5       & 168,438,898    & 66235     & 102.2           & 731,691,363    & 129,523   \\
        &     & 2   & 3282            & 3,391,255  & 67344     & 4044            & 5,413,350      & 82751     & 1323       & 1,336,804      & 30561     & 12922           & 4,993,251      & 83431     \\
        &     & 3   & 3209            & 3,202,774  & 66739     & 4058            & 5,413,350      & 82751     & 1283       & 1,281,716      & 30500     & 11711           & 5,333,175      & 83959     \\
        & 35  & 1   & 11636           & 0          & 0         & 2633            & 30,154,061,700 & 608,180   & 4440       & 16,767,014,292 & 942,020   & 4250            & 30,080,297,160 & 816,139   \\
        &     & 2   & \textbf{A}90649 & 9,481,265  & 68589     & \textbf{S}      &                &           & \textbf{L} &                &           & \textbf{A}40440 &                &           \\
        &     & 3   & \textbf{A}60657 & 8,729,650  & 52968     & \textbf{A}36000 &                &           & \textbf{L} &                &           & \textbf{A}32280 &                &           \\
\hline
  \end{tabular}
  \caption{Solver times for $G_{k,h}^i$ for default \texttt{MiniSat} and \texttt{MiniSat -no-pre} (turning off pre-processing), \texttt{CryptoMiniSat}, and \texttt{Glucose}. ``S'' marks a segmentation fault of the solver, ``L'' marks that the solver failed due to ``too long clauses'', and ``A'' marks a user-abortion.}
  \label{fig:minisatstats}
\end{figure}

\begin{figure}[H]
  \tabcolsep 1pt
  \footnotesize
  \centering
  \hspace*{-0.5em}\begin{tabular}{|c|c|c||l|l|l||l|l|l||l|l|l|}
    \cline{4-12}
    \multicolumn{3}{c|}{} & \multicolumn{3}{c||}{\texttt{Lingeling}} & \multicolumn{3}{c||}{\texttt{PrecoSAT}} & \multicolumn{3}{c|}{\texttt{PicoSAT}}\\
    \hline
    $k$ & $h$ & $i$  & $t$ (sec)      & decisions            & confl &  $t$ (sec)      & decisions            & confl & $t$ (sec)      & decisions            & confl    \\
    \hline\hline
2       & 22  & 1    & 0.0       & 31414         & 100     & 0.0     & 0          & 1     & 0.0       & 5832            & 254   \\
        &     & 2    & 0.0       & 972            & 100     & 0.0     & 20         & 16    & 0.0       & 0                & 254   \\
        &     & 3    & 0.0       & 972            & 100     & 0.0     & 20         & 16    & 0.0       & 0                & 254   \\
        & 32  & 1    & 0.2       & 55014         & 100     & 0.0     & 0          & 1     & 0.0       & 21843           & 585   \\
        &     & 2    & 0.0       & 1593          & 100     & 0.1     & 18         & 20    & 0.0       & 0                & 529   \\
        &     & 3    & 0.0       & 1593          & 100     & 0.1     & 18         & 20    & 0.0       & 0                & 529   \\
        & 42  & 1    & 0.3       & 106,962        & 100     & 0.0     & 0          & 1     & 0.0       & 39980           & 964   \\
        &     & 2    & 0.0       & 2413          & 100     & 0.4     & 39798     & 560   & 0.0       & 0                & 904   \\
        &     & 3    & 0.0       & 2413          & 100     & 0.4     & 39798     & 560   & 0.0       & 0                & 904   \\
        & 52  & 1    & 0.5       & 195,342        & 100     & 0.1     & 0          & 1     & 0.0       & 87623           & 1411 \\
        &     & 2    & 0.1       & 3432          & 100     & 0.9     & 135,771    & 1438 & 0.1       & 0                & 1379 \\
        &     & 3    & 0.1       & 3432          & 100     & 1.0     & 135,771    & 1438 & 0.0       & 0                & 1379 \\
        & 62  & 1    & 2.1       & 2,908,253      & 1528   & 0.1     & 0          & 1     & 0.1       & 195,811          & 2023 \\
        &     & 2    & 1.4       & 9993          & 338     & 2.0     & 268,652    & 2398 & 0.1       & 0                & 1954 \\
        &     & 3    & 1.2       & 8754          & 343     & 2.2     & 268,652    & 2398 & 0.1       & 0                & 1954 \\
        & 72  & 1    & 3.8       & 5,780,521      & 2100   & 0.2     & 0          & 1     & 0.1       & 358,396          & 2689 \\
        &     & 2    & 3.4       & 41069         & 835     & 4.1     & 452,024    & 3493 & 0.2       & 0                & 2629 \\
        &     & 3    & 1.6       & 17563         & 745     & 4.4     & 452,024    & 3493 & 0.1       & 0                & 2629 \\
\hline
3       & 23  & 1    & 0.9       & 772,664        & 655     & 0.0     & 0          & 1     & 0.1       & 373,029          & 2217 \\
        &     & 2    & 0.3       & 4730          & 100     & 0.9     & 21         & 17    & 0.1       & 0                & 2048 \\
        &     & 3    & 0.3       & 4730          & 100     & 0.9     & 21         & 17    & 0.1       & 0                & 2048 \\
        & 33  & 1    & 7.0       & 11,494,104     & 4470   & 0.2     & 0          & 1     & 0.4       & 1,832,220        & 6261 \\
        &     & 2    & 15.6      & 133,145        & 4822   & 8.8     & 9954      & 209   & 3.6       & 301,757          & 7808 \\
        &     & 3    & 34.8      & 187,408        & 4941   & 8.7     & 9954      & 209   & 3.8       & 397,594          & 7774 \\
        & 43  & 1    & 54.6      & 103,646,649    & 13585  & 1.1     & 0          & 1     & 1.5       & 6,710,296        & 13635\\
        &     & 2    & 834.3     & 1,058,591      & 28616  & 53.9    & 2,137,226  & 17295& 125.1     & 2,259,244        & 20567\\
        &     & 3    & 683.6     & 920,917        & 29862  & 54.7    & 2,137,226  & 17295& 125.2     & 2,378,109        & 20760\\
\hline
4       & 24  & 1    & 33.2      & 61,516,109     & 13324           & 0.5     & 0          & 1      & 1.3       & 7,197,271        & 13337      \\
        &     & 2    & 201.0     & 420,199        & 19113           & 30.4    & 87937     & 857    & 44.5      & 730,302          & 16283      \\
        &     & 3    & 411.0     & 813,880        & 20978           & 29.9    & 87937     & 857    & 53.5      & 899,721          & 16270      \\
        & 34  & 1    & 389.4     & 736,985,317    & 54187           & 9.2     & 0          & 1      & 15.9      & 77,852,480       & 54002      \\
        &     & 2    & 25004  & 4,431,011      & 103,069          & 751.9   & 37,282,690 & 64688 & 5110   & 4,952,348        & 73501      \\
        &     & 3    & 18593  & 5,206,665      & 119,524          & 735.8   & 37,282,690 & 64688 & 5822   & 6,354,378        & 73540      \\
        & 44  & 1    & 3139  & 5,980,879,353  & 152,934          & 94.2    & 0          & 1      & 135.7     & 524,180,945      & 152,931     \\
        &     & 2    & \textbf{A}94270  & 7,284,838      & 72027  & 35356 & 1,035,463,259 & 410,510 & \textbf{M} &&\\
        &     & 3    & \textbf{A}60882  & 7,688,765      & 71408  & 44808 & 1,035,463,259 & 410,510 & \textbf{M} &&\\
\hline
5       & 25  & 1    & 479.6     & 903,741,154    & 70177     & 10.1    & 0          & 1      & 25.6      & 120,756,190      & 69336      \\
        &     & 2    & 37201  & 5,026,759      & 124,208    & 3636 & 31,539,722 & 31092 & 5523   & 4,436,819        & 83821      \\
        &     & 3    & 19148  & 3,958,185      & 117,605    & 3484 & 31,539,722 & 31092 & 6540   & 5,328,658        & 83829      \\
        & 35  & 1    & 14845  & 28,147,090,014 & 392,047    & 389.0   & 0          & 1      & \textbf{F}478.5  &&\\
        &     & 2    & \textbf{A}687,866 & 11,495,987     & 39217  & \textbf{F}49440 &&& \textbf{M} &&\\
        &     & 3    & \textbf{A}94779  & 6,939,217      & 30146  & \textbf{F}49527 &&& \textbf{M} &&\\
        \hline
  \end{tabular}
  \caption{Solver times for $G_{k,h}^i$ for \texttt{Lingeling}, \texttt{PrecoSAT}, and \texttt{PicoSAT}. ``M'' marks a failure of the solver due to ``out of memory'', ``F'' marks a self-declared failure of the solver, and ``A'' marks a user-abortion.}
  \label{fig:lingelingstats}
\end{figure}

 Turning to $G_{k,h}^{2,3}$, we see that \texttt{CryptoMiniSat} as well as \texttt{PicoSAT} solve the easier instances with failed-literal elimination (without branching). Most of the time these instances are harder than their $G_{k,h}^1$ counterparts, and for $k \in \set{4,5}$ much more so, and actually no solver here was able to handle $k=5$ with $h=35$. There seems to be no essential difference between $G_{k,h}^2$ and $G_{k,h}^3$ (different from the \OKsolver{}, whose running time was proportionally larger for $G_{k,h}^2$, according to the bigger size).

\section{Conclusion and open problems}
\label{sec:open}

We have discussed three hierarchies $\Propc_k$, $\Urefc_k$ and $\Wrefc_k$ of target classes for ``good'' SAT representations. We showed that each level of $\Urefc_{k+1}$ contains clause-sets without equivalent short clause-sets in $\Wrefc_k$. And we showed conditions under which the Tseitin translation produces translations in $\Urefc$. We conclude by directions for future research.

\subsection{Strictness of hierarchies}
\label{sec:conclstrict}

A fundamental question is the strictness of the hierarchies $\Propc_k$, $\Urefc_k$ and $\Wrefc_k$ in each of the dimensions. In Theorem \ref{thm:separation} we have shown w.r.t.\ logical equivalence (i.e., without new variables) that the $\Urefc_k$ and $\Wrefc_k$ hierarchies are strict. It follows that for $\Propc_k$ at least every second level yields an advance regarding logical equivalence (and polysize). This offer evidence that these hierarchies are useful, for example using failed literal reduction can allow one to use exponentially smaller SAT translations. Open are the questions of strictness for the hierarchies allowing new variables. To summarise, the main conjectures are:
\begin{enumerate}
\item Conjecture \ref{con:sepsharp} strengthens Theorem \ref{thm:separation} by taking the PC-hierarchy into account.
\item Conjecture \ref{con:sepext} roughly says that all of $\Propc_k$, $\Urefc_k$ and $\Wrefc_k$ are strict (similar to Theorem \ref{thm:separation}), when allowing new variables under the \emph{absolute} condition.
\item Conjecture \ref{con:collapseWrefc} says that the $\Wrefc_k$ hierarchy collapses to $\Wrefc_1$ (and thus to $\Propc_1$), when allowing new variables under the \emph{relative} condition.
\end{enumerate}

\subsection{Separating the different hierarchies}
\label{sec:conclsep}

For stating our three main conjectures relating the three hierarchies, we use the following notions:
\begin{itemize}
\item A sequence $(F_n')_{n \in \NN}$ is called a CNF-representation of $(F_n)_{n \in \NN}$ if for all $n \in \NN$ the clause-set $F_n'$ is a CNF-representation of $F_n$.
\item A \textbf{polysize sequence in $\mc{C} \sse \Cls$} is a sequence $(F_n)_{n \in \NN}$ with $F_n \in \mc{C}$ for all $n \in \NN$, such that $(\ell(F_n))_{n \in \NN}$ is polynomially bounded (i.e., there is a polynomial $p(x)$ with $\ell(F_n) \le p(n)$ for all $n \in \NN$).
\end{itemize}

We conjecture that $\Wrefc_2$ even without new variables offers possibilities for good representations not offered by any $\Urefc_k$:
\begin{conj}\label{con:w2nosimstr}
  There exists a polysize $(F_n)_{n \in \NN}$ in $\Wrefc_2$, such that for no $k \in \NNZ$ there exists a polysize CNF-representation $(F_n')_{n \in \NN}$ of $(F_n)_{n \in \NN}$ in $\Urefc_k$.
\end{conj}
A proof of Conjecture \ref{con:w2nosimstr} needed, besides the novel handling of the new variables, to develop lower-bounds methods specifically for hardness, since the method via trigger hypergraphs yields already lower bounds for w-hardness.

We conjecture that new variables can not simulate higher hardness, strengthening Theorem \ref{thm:separation}, Conjecture \ref{con:sepsharp} and Conjecture \ref{con:sepext}:
\begin{conj}\label{con:newvnhhd}
  For every $k \in \NNZ$ there exists a polysize $(F_n)_{n \in \NN}$ in $\Propc_{k+1}$, such that there is no polysize CNF-representation $(F_n')_{n \in \NN}$ of $(F_n)_{n \in \NN}$ in $\Wrefc_k$.
\end{conj}

Finally we conjecture that there is a sequence of boolean functions which has polysize arc-consistent representations, but no polysize representations of bounded hardness, even for the w-hardness:
\begin{conj}\label{con:relhdstrong}
  There exists a polysize $(F_n)_{n \in \NN}$ in $\Cls$, such that there is a polysize CNF-representation $(F_n')_{n \in \NN}$ of $(F_n)_{n \in \NN}$ with $\hardness^{\var(F_n)}(F_n') \le 1$ for all $n \in \NN$, while for no $k \in \NNZ$ there is a polysize CNF-representation $(F_n'')_{n \in \NN}$ of $(F_n)_{n \in \NN}$ in $\Wrefc_k$.
\end{conj}
Regarding our notion of a ``polysize sequence'' $(F_n)_{n \in \NN}$, this is a very liberal notion, allowing to express arbitrary boolean functions, since the number of variables could be logarithmic in the index, and thus $F_n$ could contain exponentially many clauses in the number of variables. The sequence of Theorem \ref{thm:separation} also fulfils $n(F_n) = \Omega(n)$, and making this provision one could speak of ``simple'' boolean functions, however this would complicate the formulations of our conjectures, and so we abstained from it.

We conclude our considerations on hierarchies by considering the three hierarchies $\Altsluri{k}$ introduced in \cite{Vlcek2010ClassesBoolPolySLUR}, $\Altslurstari{k}$ introduced in \cite{CepekKuceraVlcek2012SLUR}, and $\Canoni{k}$ introduced in \cite{BalyoGurskyKuceraVlcek2012SLURHier}, which we have compared to the UC-hierarchy in \cite{GwynneKullmann2012SlurSOFSEM,GwynneKullmann2012Slur,GwynneKullmann2012SlurJ}. From the point of view of polysize representations without new variables, the hierarchy $\Canoni{k}$ collapses to $\Canoni{0} = \Urefc_0$:
\begin{lem}\label{lem:collapsecanon}
  For $F \in \Cls$ let $k(F)$ be the minimal $k \in \NNZ$ such that $F \in \Canoni{k}$. Then the function $\primec_0: \Cls \ra \Canoni{0} = \Urefc_0$ can be computed in time $O(c(F)^{3 \cdot 2^{k}} \cdot \ell(F))$, when the input is $F$ together with $k := k(F)$.
\end{lem}
\begin{prf}
  Let $K := 2^k$. So for every $C \in \primec_0(F)$ there exists $F' \sse F$ with $F' \models C$ and $c(F') \le K$, since a resolution tree of height $k$ has at most $K$ leaves. Now we compute $\primec_0(F)$ as follows:
  \begin{enumerate}
  \item Set $P := \es$.
  \item Run through all $F' \sse F$ with $c(F') \le K$; their number is $O(c(F)^K)$.
  \item For each $F'$ determine whether $F' \models \purec(F')$ holds, in which case clause $\purec(F')$ is added to $P$; note that the test can be performed in time $O(2^K \cdot K)$.
  \item The final $P$ obtained has $O(c(F)^K)$ many elements. After performing subsumption elimination (in cubic time) we obtain $\primec_0(F)$ (by Lemma \ref{lem:uniquepurec}). \Qed
  \end{enumerate}
\end{prf}
It seems an interesting question whether the two other hierarchies $\Altsluri{k}$, $\Altslurstari{k}$ collapse or not, and whether they can be reduced to some $\Urefc_k$.

\subsection{Compilation procedures}
\label{sec:conclcompil}

For a given boolean function $f$ and $k \in \NNZ$, how do we find algorithmically a ``small'' equivalent $F \in \Urefc_k$ ? In \cite{GwynneKullmann2012SlurJ}, Section 8, the notion of a ``$k$-base for $f$'' is introduced, which is an $F \in \Urefc_k$ equivalent to $f$, with $F \sse \primec_0(f)$ and where no clause can be removed without increasing the hardness or destroying equivalence. It is shown that if $f$ is given as a 2-CNF, then a smallest $k$-base is computable in polynomial time, but even for $f$ with given $\primec_0(f)$, where $\primec_0(f)$ is a Horn clause-set, deciding whether a $k$-base of a described size for a fixed $k \ge 1$ exists is NP-complete.

There are interesting applications where $\primec_0(f)$ is given (or can be computed), and where then some small equivalent $F \in \Urefc_k$ is sought. The most basic approach filters out unneeded prime implicates; see \cite{GwynneKullmann2011TranslationsPrelim,GwynneKullmann2011HardnessPrelim} for some initial applications to cryptanalysis. A simple filtering heuristic, used in \cite{GwynneKullmann2011TranslationsPrelim,GwynneKullmann2011HardnessPrelim}, is to favour (keeping) short-clauses. In a first phase, starting with the necessary elements of $\primec_0(f)$, further elements are added (when needed) in ascending order of size for building up the initial $F \in \Urefc_k$ (which in general is not a base). In the second phase, clauses from $F$ are removed in descending order of size when reducing to a $k$-base. The intuition behind this heuristic is that small clauses cover more total assignments (so fewer are needed), and they are also more likely to trigger $\rk_k$, making them more useful in producing small, powerful representations. Essentially the same heuristic is considered in \cite{BordeauxMarquesSilva2012KnowledgeCompilation} (called ``length-increasing iterative empowerment'') when generating representations in $\Propc$.

For the case that $f$ is given by a CNF $F_0$, in \cite{Val1994UnitResolutionComplete} one finds refinements of the resolution procedure applied to $F_0$, which would normally compute $\primec_0(f)$, i.e., the $0$-base in $\Urefc_0$, and where by some form of ``compression'' now an equivalent $F \in \Urefc_1$ is computed. This approach needed to be generalised to arbitrary $\Urefc_k$.

\subsection{Exploring w-hardness}
\label{sec:concwhard}

It is to be expected that w-hardness can behave very differently from hardness. For example, as expressed by Conjecture \ref{con:w2nosimstr}, already its second level should enable contain short clause-sets not representable in any $\Urefc_k$. However yet we do not have tools at hand to handle w-hardness (we do not even have yet a conjectured example for such a separation). A first task is to investigate which of the results on hardness from this article and from \cite{GwynneKullmann2012SlurJ} can be adapted to w-hardness. In \cite{BeyersdorffGwynneKullmann2013PHPER} we will present some basic methods for w-hardness bounds.

Can the classes $\Wrefc_k$ go beyond monotone circuits, which were shown in \cite{BKNW2009CircuitComplexity} to be strongly related to the expressive power of arc-consistent CNF representations (see the following subsection for some further remarks)? Conjecture \ref{con:collapseWrefc} would show the contrary, namely that in the (unrestricted) presence of new variables also w-hardness boils down, modulo polytime computations, to $\Propc_1$ (under the relative condition!). If this is true, then the believable greater power of $\Wrefc_k$ over $\Urefc_k$ would all take place inside arc-consistency; and by Conjecture \ref{con:relhdstrong} it would take place strictly inside arc-consistency.

\subsection{Hard boolean functions handled by oracles}
\label{sec:conclhard}

Finally we turn to concrete (sequences of) boolean functions which are currently out of reach of good presentations, and where the use of oracles thus is necessary.

Conjecture \ref{con:xorcls} says that systems of XOR-clauses (affine equations) have no good representation, even when just considering arc-consistency. So the conjecture is that here we have another example for the limitations of arc-consistent representations as shown in \cite{BKNW2009CircuitComplexity}. To overcome these (conjectured) limitations, the theory started here has to be generalised via the use of oracles as developed in \cite{Ku99b,Ku00g}, and further discussed in Subsection 9.4 of \cite{GwynneKullmann2012Slur,GwynneKullmann2012SlurJ}. The point of these oracles, which are just sets $\mc{U} \sse \Usat$ of unsatisfiable clause-sets stable under application of partial assignments, is to discover hard \emph{unsatisfiable} (sub-)instances (typically in polynomial time). Thus they are conceptually simpler than the current integration of SAT solvers and methods from linear algebra (see \cite{CryptoSAT2009,Chen2009HybridXOR,Soos2010SATGauss,HanJiang2012GaussianSAT,LaitinenJunttilaNiemela2012ConflictXORLearn}).

An important aspect of the theory to be developed must be the usefulness of the representation (with oracles) in context, that is, as a ``constraint'' in a bigger problem: a boolean function $f$ represented by a clause-set $F$ is typically contained in $F' \supset F$, where $F'$ is the SAT problem to be solved (containing also other constraints). One approach is to require from the oracle also stability under addition of clauses, as we have it already for the resolution-based reductions like $\rk_k$, so that the (relativised) reductions $\rk_k^{\mc{U}}$ can always run on the whole clause-set (an instantiation of $F'$). However for example for the oracle mentioned below, based on semidefinite programming, this would be prohibitively expensive. And for some oracles, like detection of minimally unsatisfiable clause-sets of a given deficiency, the problems would turn from polytime to NP-hard in this way (\cite{FKS00,BueningZhao2002MUsubsets}). Furthermore, that we have some representation of a constraint which would benefit for example from some XOR-oracle, does not mean that in other parts of the problems that oracle will also be of help. So in many cases it is better to restrict the application of the oracle $\mc{U}$ to that subset $F \subset F'$ where to achieve the desired hardness the oracle is actually required.

Another example of a current barrier is given by the satisfiable pigeonhole clause-sets $\php^m_m$, which have variables $p_{i,j}$ for $i, j \in \tb 1m$, and where the satisfying assignments correspond precisely to the permutations of $\tb 1m$. The question is about ``good'' representations. In \cite{BeyersdorffGwynneKullmann2013PHPER} we show $\hardness(\php^m_m) = \whardness(\php^m_m) = m-1$, and so the (standard representation) $\php^m_m \in \Cls$ itself is not a good representation (it is small, but has high w-hardness). Actually, as explained in \cite{BeyersdorffGwynneKullmann2013PHPER}, from \cite{BKNW2009CircuitComplexity} it follows that $\php^m_m$ has no polysize arc-consistent representation at all! So again, here oracles are needed; see Subsection 9.4 of \cite{GwynneKullmann2012Slur,GwynneKullmann2012SlurJ} for a proposal of an interesting oracle (with potentially good stability properties).

\bibliographystyle{plain}

\begin{thebibliography}{10}

\bibitem{AnsoteguiBonetLevyManya2008Hardness}
Carlos Ans{\'{o}}tegui, Mar{\'{\i}}a~Luisa Bonet, Jordi Levy, and Felip
  Many{\`{a}}.
\newblock Measuring the hardness of {SAT} instances.
\newblock In Dieter Fox and Carla Gomes, editors, {\em Proceedings of the 23th
  AAAI Conference on Artificial Intelligence (AAAI-08)}, pages 222--228, 2008.

\bibitem{AtseriasFichteThurley2009ClauseLearningBoundedWidth}
Albert Atserias, Johannes~Klaus Fichte, and Marc Thurley.
\newblock Clause-learning algorithms with many restarts and bounded-width
  resolution.
\newblock In Kullmann \cite{Swansea2009}, pages 114--127.
\newblock ISBN 978-3-642-02776-5.

\bibitem{AudemardSimon2009PredictingLearnt}
Gilles Audemard and Laurent Simon.
\newblock Predicting learnt clauses quality in modern {SAT} solvers.
\newblock In {\em IJCAI'09 Proceedings of the 21st International Joint
  Conference on Artificial intelligence}, pages 399--404. AAAI, 2009.

\bibitem{BailleuzBoufkhad2003CardinalityConstraints}
Olivier Bailleux and Yacine Boufkhad.
\newblock Efficient {CNF} encoding of boolean cardinality constraints.
\newblock In {\em Principles and Practice of Constraint Programming -- CP
  2003}, volume 2833 of {\em Lecture Notes in Computer Science}, pages
  108--122, 2003.

\bibitem{BailleuxBoufkhadRoussel2009PBCNF}
Olivier Bailleux, Yacine Boufkhad, and Olivier Roussel.
\newblock New encodings of pseudo-boolean constraints into {CNF}.
\newblock In Kullmann \cite{Swansea2009}, pages 181--194.
\newblock ISBN 978-3-642-02776-5.

\bibitem{BalyoGurskyKuceraVlcek2012SLURHier}
Tom\'{a}\v{s} Balyo, \v{S}tefan Gursk\'{y}, Petr Ku{\v{c}}era, and V{\'{a}}clav
  Vl{\v{c}}ek.
\newblock On hierarchies over the {SLUR} class.
\newblock In {\em Twelfth International Symposium on Artificial Intelligence
  and Mathematics (ISAIM 2012)}, January 2012.
\newblock Available at
  \url{http://www.cs.uic.edu/bin/view/Isaim2012/AcceptedPapers}.

\bibitem{SW98}
Eli Ben-Sasson and Avi Wigderson.
\newblock Short proofs are narrow --- resolution made simple.
\newblock In {\em Proceedings of the 31th Annual ACM Symposium on Theory of
  Computing (STOC'99)}, pages 517--526, May 1999.

\bibitem{BKNW2009CircuitComplexity}
Christian Bessiere, George Katsirelos, Nina Narodytska, and Toby Walsh.
\newblock Circuit complexity and decompositions of global constraints.
\newblock In {\em Proceedings of the Twenty-First International Joint
  Conference on Artificial Intelligence (IJCAI-09)}, pages 412--418, 2009.

\bibitem{BeyersdorffGwynneKullmann2013PHPER}
Olaf Beyersdorff, Matthew Gwynne, and Oliver Kullmann.
\newblock Hardness measures and resolution lower bounds, with applications to
  {P}igeonhole principles.
\newblock In preparation, to appear on arXiv, July 2013.

\bibitem{Biere2008picosat}
Armin Biere.
\newblock Picosat essentials.
\newblock {\em Journal on Satisfiability, Boolean Modeling and Computation},
  4:75--97, 2008.

\bibitem{Biere2009Precosat}
Armin Biere.
\newblock P\{re,i\}co{SAT}@{SC}'09.
\newblock \url{http://fmv.jku.at/precosat/preicosat-sc09.pdf}, 2009.

\bibitem{Biere2010PicoSATLingelingDesc}
Armin Biere.
\newblock Lingeling, {P}lingeling, {PicoSAT} and {PrecoSAT} at {SAT} {R}ace
  2010.
\newblock Technical report, Institute for Formal Models and Verification,
  Johannes Kepler University, Altenbergerstr. 69, 4040 Linz, Austria, August
  2010.
\newblock \url{http://fmv.jku.at/papers/Biere-FMV-TR-10-1.pdf}.

\bibitem{Biere2012LingelingDesc}
Armin Biere.
\newblock Lingeling and friends entering the {SAT} {C}hallenge 2012.
\newblock In Adrian Balint, Anton Belov, Daniel Diepold, Simon Gerber, Matti
  J\"{a}rvisalo, and Carsten Sinz, editors, {\em Proceedings of {SAT} Challenge
  2012: Solver and Benchmark Descriptions}, volume B-2012-2 of {\em Department
  of Computer Science Series of Publications B}, pages 33--34. University of
  Helsinki, 2012.
\newblock
  \url{https://helda.helsinki.fi/bitstream/handle/10138/34218/sc2012_proceedings.pdf}.

\bibitem{2008HandbuchSAT}
Armin Biere, Marijn~J.H. Heule, Hans van Maaren, and Toby Walsh, editors.
\newblock {\em Handbook of Satisfiability}, volume 185 of {\em Frontiers in
  Artificial Intelligence and Applications}.
\newblock IOS Press, February 2009.

\bibitem{BordeauxMarquesSilva2012KnowledgeCompilation}
Lucas Bordeaux and Joao Marques-Silva.
\newblock Knowledge compilation with empowerment.
\newblock In M{\'{a}}ria Bielikov{\'{a}}, Gerhard Friedrich, Georg Gottlob,
  Stefan Katzenbeisser, and Gy{\"{o}}rgy Tur{\'{a}}n, editors, {\em SOFSEM
  2012: Theory and Practice of Computer Science}, volume 7147 of {\em Lecture
  Notes in Computer Science}, pages 612--624. Springer, 2012.

\bibitem{BueningZhao2002MUsubsets}
Hans~Kleine B{\"{u}}ning and Xishun Zhao.
\newblock The complexity of read-once resolution.
\newblock {\em Annals of Mathematics and Artificial Intelligence},
  36(4):419--435, December 2002.

\bibitem{BuroKleineBuening1996ResolutionShortClauses}
Michael Buro and Hans {Kleine B{\"u}ning}.
\newblock On resolution with short clauses.
\newblock {\em Annals of Mathematics and Artificial Intelligence},
  18(2-4):243--260, 1996.

\bibitem{CadoliDonini1997SurveyKnowledgeComp}
Marco Cadoli and Francesco~M. Donini.
\newblock A survey of knowledge compilation.
\newblock {\em AI Communications}, 10(3,4):137--150, December 1997.

\bibitem{CepekKuceraVlcek2012SLUR}
Ond{\v{r}}ej {\v{C}}epek, Petr Ku{\v{c}}era, and V{\'{a}}clav Vl{\v{c}}ek.
\newblock Properties of {SLUR} formulae.
\newblock In M{\'{a}}ria Bielikov{\'{a}}, Gerhard Friedrich, Georg Gottlob,
  Stefan Katzenbeisser, and Gy{\"{o}}rgy Tur{\'{a}}n, editors, {\em SOFSEM
  2012: Theory and Practice of Computer Science}, volume 7147 of {\em LNCS
  Lecture Notes in Computer Science}, pages 177--189. Springer, 2012.

\bibitem{Chen2009HybridXOR}
Jiangchao Chen.
\newblock Building a hybrid {SAT} solver via conflict-driven, look-ahead and
  {XOR} reasoning techniques.
\newblock In Kullmann \cite{Swansea2009}, pages 298--311.
\newblock ISBN 978-3-642-02776-5.

\bibitem{Co73}
Stephen~A. Cook.
\newblock An exponential example for analytic tableaux.
\newblock Manuscript (see \cite{Urq95}, page 432), 1973.

\bibitem{CramaHammer2011BooleanFunctions}
Yves Crama and Peter~L. Hammer.
\newblock {\em Boolean Functions: Theory, Algorithms, and Applications}, volume
  142 of {\em Encyclopedia of Mathematics and Its Applications}.
\newblock Cambridge University Press, 2011.
\newblock ISBN 978-0-521-84751-3.

\bibitem{CreignouKolaitisVollmer2008ComplexityConstraints}
Nadia Creignou, Phokion Kolaitis, and Heribert Vollmer, editors.
\newblock {\em Complexity of Constraints: An Overview of Current Research
  Themes}, volume 5250 of {\em Lecture Notes in Computer Science (LNCS)}.
\newblock Springer, 2008.
\newblock ISBN-10 3-540-92799-9.

\bibitem{DH09HBSAT}
Evgeny Dantsin and Edward~A. Hirsch.
\newblock Worst-case upper bounds.
\newblock In Biere et~al. \cite{2008HandbuchSAT}, chapter~12, pages 403--424.

\bibitem{DarwicheMarquis2002KCmap}
Adnan Darwiche and Pierre Marquis.
\newblock A knowledge compilation map.
\newblock {\em Journal of Artificial Intelligence Research}, 17:229--264, 2002.

\bibitem{Val1994UnitResolutionComplete}
Alvaro del Val.
\newblock Tractable databases: How to make propositional unit resolution
  complete through compilation.
\newblock In {\em Proceedings of the 4th International Conference on Principles
  of Knowledge Representation and Reasoning (KR'94)}, pages 551--561, 1994.

\bibitem{Val2000UnitResolutionComplete}
Alvaro del Val.
\newblock On some tractable classes in deduction and abduction.
\newblock {\em Artificial Intelligence}, 116(1-2):297--313, January 2000.

\bibitem{Een2006Translating}
Niklas E{\'e}n and Niklas S{\"o}rensson.
\newblock Translating pseudo-boolean constraints into {SAT}.
\newblock {\em Journal on Satisfiability, Boolean Modeling and Computation},
  2:1--26, March 2006.

\bibitem{FargierMarquis2008ExtendKCMapKromHornAff}
H\'{e}l\`{e}ne Fargier and Piere Marquis.
\newblock Extending the knowledge compilation map: {K}rom, {H}orn, {A}ffine and
  beyond.
\newblock In Malik Ghallab, Constantine~D. Spyropoulos, Nikos Fakotakis, and
  Nikos Avouris, editors, {\em ECAI 2008}, volume 178 of {\em Frontiers in
  Artificial Intelligence and Applications}, pages 50--54. IOS Press, 2008.

\bibitem{FKS00}
Herbert Fleischner, Oliver Kullmann, and Stefan Szeider.
\newblock Polynomial--time recognition of minimal unsatisfiable formulas with
  fixed clause--variable difference.
\newblock {\em Theoretical Computer Science}, 289(1):503--516, November 2002.

\bibitem{FrGe98}
John Franco and Allen~Van Gelder.
\newblock A perspective on certain polynomial-time solvable classes of
  satisfiability.
\newblock {\em Discrete Applied Mathematics}, 125:177--214, 2003.

\bibitem{Gent2002ArcConsistency}
Ian~P. Gent.
\newblock Arc consistency in {SAT}.
\newblock In Frank van Harmelen, editor, {\em Proceedings of the 15th European
  Conference on Artificial Intelligence (ECAI 2002)}, pages 121--125. IOS
  Press, 2002.

\bibitem{GW00}
Jan~Friso Groote and Joost~P. Warners.
\newblock The propositional formula checker {H}eer{H}ugo.
\newblock {\em Journal of Automated Reasoning}, 24:101--125, 2000.

\bibitem{GwynneKullmann2011HardnessPrelim}
Matthew Gwynne and Oliver Kullmann.
\newblock Towards a better understanding of hardness.
\newblock In {\em The Seventeenth International Conference on Principles and
  Practice of Constraint Programming (CP 2011): Doctoral Program Proceedings},
  pages 37--42, September 2011.
\newblock Proceedings available at
  \url{http://www.dmi.unipg.it/cp2011/downloads/dp2011/DP_at_CP2011.pdf}.

\bibitem{GwynneKullmann2011TranslationsPrelim}
Matthew Gwynne and Oliver Kullmann.
\newblock Towards a better understanding of {SAT} translations.
\newblock In Ulrich Berger and Denis Therien, editors, {\em Logic and
  Computational Complexity (LCC'11), as part of LICS 2011}, June 2011.
\newblock 10 pages, available at \url{http://www.cs.swansea.ac.uk/lcc2011/}.

\bibitem{GwynneKullmann2012SlurSOFSEM}
Matthew Gwynne and Oliver Kullmann.
\newblock Generalising and unifying {SLUR} and unit-refutation completeness.
\newblock In Peter van Emde~Boas, Frans C.~A. Groen, Giuseppe~F. Italiano,
  Jerzy Nawrocki, and Harald Sack, editors, {\em SOFSEM 2013: Theory and
  Practice of Computer Science}, volume 7741 of {\em Lecture Notes in Computer
  Science (LNCS)}, pages 220--232. Springer, 2013.

\bibitem{GwynneKullmann2012Slur}
Matthew Gwynne and Oliver Kullmann.
\newblock Generalising unit-refutation completeness and {SLUR} via nested input
  resolution.
\newblock Technical Report arXiv:1204.6529v5 [cs.LO], arXiv, January 2013.

\bibitem{GwynneKullmann2012SlurJ}
Matthew Gwynne and Oliver Kullmann.
\newblock Generalising unit-refutation completeness and {SLUR} via nested input
  resolution.
\newblock {\em Journal of Automated Reasoning}, 2013.
\newblock To appear.

\bibitem{HanJiang2012GaussianSAT}
Cheng-Shen Han and Jie-Hong~Roland Jiang.
\newblock When boolean satisfiability meets {G}aussian elimination in a
  {S}implex way.
\newblock In P.~Madhusudan and Sanjit~A. Seshia, editors, {\em Computer Aided
  Verification}, volume 7358 of {\em Lecture Notes in Computer Science}, pages
  410--426. Springer, July 2012.

\bibitem{Har96}
John Harrison.
\newblock St{\r{a}}lmarck's algorithm as a {HOL} derived rule.
\newblock In {\em Theorem proving in higher order logics: 9th International
  Conference, TPHOLs'96}, Lecture Notes in Computer Science 1125, pages
  221--234, 1996.

\bibitem{HeuleMaaren2005MarchEq}
Marijn Heule, Mark Dufour, Joris van Zwieten, and Hans van Maaren.
\newblock Marcheq: Implementing additional reasoning into an efficient
  look-ahead {SAT} solver.
\newblock In Hoos and Mitchell \cite{Vancouver2004b}, pages 345--359.
\newblock
  \url{http://www.st.ewi.tudelft.nl/\%7Emarijn/publications/35420345.pdf}.

\bibitem{HvM09HBSAT}
Marijn J.~H. Heule and Hans van Maaren.
\newblock Look-ahead based {SAT} solvers.
\newblock In Biere et~al. \cite{2008HandbuchSAT}, chapter~5, pages 155--184.

\bibitem{Vancouver2004b}
Holger~H. Hoos and David~G. Mitchell, editors.
\newblock {\em Theory and Applications of Satisfiability Testing 2004}, volume
  3542 of {\em Lecture Notes in Computer Science}, Berlin, 2005. Springer.
\newblock ISBN 3-540-27829-X.

\bibitem{JacksonSheridan2004CircuitstoSAT}
Paul Jackson and Daniel Sheridan.
\newblock Clause form conversions for boolean circuits.
\newblock In Hoos and Mitchell \cite{Vancouver2004b}, pages 183--198.
\newblock ISBN 3-540-27829-X.

\bibitem{JarvisaloJunttila2009LimitRestrictedLearning}
Matti J\"arvisalo and Tommi Junttila.
\newblock Limitations of restricted branching in clause learning.
\newblock {\em Constraints}, 14(3):325--356, 2009.

\bibitem{JarvisaloMatsliahNordstromZivny2012ProofCompHardnessSAT}
Matti J\"{a}rvisalo, Arie Matsliah, Jakob Nordstr\"{o}m, and Stanislav
  \v{Z}ivn\'{y}.
\newblock Relating proof complexity measures and practical hardness of {SAT}.
\newblock In Michela Milano, editor, {\em Principles and Practice of Constraint
  Programming - CP 2012}, volume 7514 of {\em Lecture Notes In Computer
  Science}, pages 316--331. Springer, 2012.

\bibitem{JarvisaloNiemala2008StructuralBranchingExperiments}
Matti J\"{a}rvisalo and Ilkka Niemel\"{a}.
\newblock The effect of structural branching on the efficiency of clause
  learning {SAT} solving: An experimental study.
\newblock {\em Journal of Algorithms}, 63(1):90--113, July 2008.

\bibitem{JovanovicKreuzer2010AlgAttackSAT}
Philipp Jovanovic and Martin Kreuzer.
\newblock Algebraic attacks using {SAT}-solvers.
\newblock {\em Groups-Complexity-Cryptology}, 2(2):247--259, November 2010.

\bibitem{BarahomaJungKatsirelosWalsh2008EncodingDNNF}
Jean~Christoph Jung, Pedro Barahoma, George Katsirelos, and Toby Walsh.
\newblock Two encodings of {DNNF} theories, July 2008.
\newblock Presented at ECAI'08 Workshop on Inference methods based on Graphical
  Structures of Knowledge. Proceedings at \url{http://www.irit.fr/LC/}.

\bibitem{KeanTsiknis1990IncrementalPrimeImp}
Alex Kean and George Tsiknis.
\newblock An incremental method for generating prime implicants/implicates.
\newblock {\em Journal of Symbolic Computation}, 9(2):185--206, February 1990.

\bibitem{Kl93}
Hans {Kleine B{\"u}ning}.
\newblock On generalized {H}orn formulas and {$k$}-resolution.
\newblock {\em Theoretical Computer Science}, 116:405--413, 1993.

\bibitem{Kullmann2007HandbuchMU}
Hans {Kleine B{\"{u}}ning} and Oliver Kullmann.
\newblock Minimal unsatisfiability and autarkies.
\newblock In Biere et~al. \cite{2008HandbuchSAT}, chapter~11, pages 339--401.

\bibitem{Ku99b}
Oliver Kullmann.
\newblock Investigating a general hierarchy of polynomially decidable classes
  of {CNF}'s based on short tree-like resolution proofs.
\newblock Technical Report TR99-041, Electronic Colloquium on Computational
  Complexity (ECCC), October 1999.

\bibitem{Ku97b}
Oliver Kullmann.
\newblock New methods for 3-{SAT} decision and worst-case analysis.
\newblock {\em Theoretical Computer Science}, 223(1-2):1--72, July 1999.

\bibitem{Ku96c}
Oliver Kullmann.
\newblock On a generalization of extended resolution.
\newblock {\em Discrete Applied Mathematics}, 96-97:149--176, October 1999.

\bibitem{Ku99dKo}
Oliver Kullmann.
\newblock An application of matroid theory to the {SAT} problem.
\newblock In {\em Fifteenth Annual IEEE Conference on Computational Complexity
  (2000)}, pages 116--124. IEEE Computer Society, July 2000.

\bibitem{Ku2002h}
Oliver Kullmann.
\newblock Investigating the behaviour of a {SAT} solver on random formulas.
\newblock Technical Report CSR 23-2002, Swansea University, Computer Science
  Report Series (available from
  \url{http://www-compsci.swan.ac.uk/reports/2002.html}), October 2002.
\newblock 119 pages.

\bibitem{Ku2003e}
Oliver Kullmann.
\newblock The combinatorics of conflicts between clauses.
\newblock In Enrico Giunchiglia and Armando Tacchella, editors, {\em Theory and
  Applications of Satisfiability Testing 2003}, volume 2919 of {\em Lecture
  Notes in Computer Science}, pages 426--440, Berlin, 2004. Springer.
\newblock ISBN 3-540-20851-8.

\bibitem{Ku00g}
Oliver Kullmann.
\newblock Upper and lower bounds on the complexity of generalised resolution
  and generalised constraint satisfaction problems.
\newblock {\em Annals of Mathematics and Artificial Intelligence},
  40(3-4):303--352, March 2004.

\bibitem{Kullmann2007HandbuchTau}
Oliver Kullmann.
\newblock Fundaments of branching heuristics.
\newblock In Biere et~al. \cite{2008HandbuchSAT}, chapter~7, pages 205--244.

\bibitem{Swansea2009}
Oliver Kullmann, editor.
\newblock {\em Theory and Applications of Satisfiability Testing - SAT 2009},
  volume 5584 of {\em Lecture Notes in Computer Science}. Springer, 2009.
\newblock ISBN 978-3-642-02776-5.

\bibitem{Kullmann2007ClausalFormZII}
Oliver Kullmann.
\newblock Constraint satisfaction problems in clausal form {II}: Minimal
  unsatisfiability and conflict structure.
\newblock {\em Fundamenta Informaticae}, 109(1):83--119, 2011.

\bibitem{KullmannZhao2012Confluence}
Oliver Kullmann and Xishun Zhao.
\newblock On {D}avis-{P}utnam reductions for minimally unsatisfiable
  clause-sets.
\newblock Technical Report arXiv:1202.2600v5 [cs.DM], arXiv, December
  \noopsort{b}2012.

\bibitem{LaitinenJunttilaNiemela2012ConflictXORLearn}
Tero Laitinen, Tommi Junttila, and Ilkka Niemel\"{a}.
\newblock Conflict-driven {XOR}-clause learning.
\newblock In Alessandro Cimatti and Roberto Sebastiani, editors, {\em Theory
  and Applications of Satisfiability Testing – SAT 2012}, volume LNCS 7317 of
  {\em Lecture Notes in Computer Science}, pages 383--396. Springer, 2012.
\newblock ISBN-13 978-3-642-31611-1.

\bibitem{LA1996}
Chu~Min Li and Anbulagan.
\newblock Heuristics based on unit propagation for satisfiability problems.
\newblock In {\em Proceedings of 15th International Joint Conference on
  Artificial Intelligence (IJCAI'97)}, pages 366--371. Morgan Kaufmann
  Publishers, 1997.

\bibitem{MarquesSilva2012MUS}
Joao Marques-Silva.
\newblock Computing minimally unsatisfiable subformulas: State of the art and
  future directions.
\newblock {\em Journal of Multiple-Valued Logic and Soft Computing},
  19(1-3):163--183, 2012.

\bibitem{MSLM09HBSAT}
Joao~P. Marques-Silva, Ines Lynce, and Sharad Malik.
\newblock Conflict-driven clause learning {SAT} solvers.
\newblock In Biere et~al. \cite{2008HandbuchSAT}, chapter~4, pages 131--153.

\bibitem{DarwichePipatsrisawat2009ClauseLearnRes}
Knot Pipatsrisawat and Adnan Darwiche.
\newblock On the power of clause-learning {SAT} solvers with restarts.
\newblock In Ian~P. Gent, editor, {\em Principles and Practice of Constraint
  Programming - CP 2009}, volume 5732 of {\em Lecture Notes in Computer
  Science}, pages 654--668. Springer, 2009.

\bibitem{DarwichePipatsrisawat2011ClauseLearnRes}
Knot Pipatsrisawat and Adnan Darwiche.
\newblock On the power of clause-learning {SAT} solvers as resolution engines.
\newblock {\em Artificial Intelligence}, 175(2):512--525, 2011.

\bibitem{GreenbaumPlaisted1986ClauseFormTrans}
David~A. Plaisted and Steven Greenbaum.
\newblock A structure-preserving clause form translation.
\newblock {\em Journal of Symbolic Computation}, 2(3):293--304, 1986.

\bibitem{Pr96}
Daniele Pretolani.
\newblock Hierarchies of polynomially solvable satisfiability problems.
\newblock {\em Annals of Mathematics and Artificial Intelligence},
  17(3-4):339--357, 1996.

\bibitem{RM09HBSAT}
Olivier Roussel and Vasco Manquinho.
\newblock Pseudo-boolean and cardinality constraints.
\newblock In Biere et~al. \cite{2008HandbuchSAT}, chapter~22, pages 695--733.

\bibitem{SSt98}
Mary Sheeran and Gunnar St{\r{a}}lmarck.
\newblock A tutorial on {S}t{\r{a}}lmarck's proof procedure for propositional
  logic.
\newblock In {\em FMCAD'98}, volume 1522 of {\em Lecture Notes in Computer
  Science}, pages 82--99, 1998.

\bibitem{Sinz2005CardinalityConstraints}
Carsten Sinz.
\newblock Towards an optimal {CNF} encoding of boolean cardinality constraints.
\newblock In {\em Principles and Practice of Constraint Programming -- CP
  2005}, volume 3709 of {\em Lecture Notes in Computer Science (LNCS)}, pages
  827--831. Springer, 2005.

\bibitem{SloanSzoerenyiTuran2005Primimplikanten_1}
Robert~H. Sloan, Bal{\'{a}}zs S{\"{o}}r{\'{e}}nyi, and Gy{\"{o}}rgy
  Tur{\'{a}}n.
\newblock On $k$-term {DNF} with the largest number of prime implicants.
\newblock {\em SIAM Journal on Discrete Mathematics}, 21(4):987--998, 2007.

\bibitem{Soos2010CMSDesc}
Mate Soos.
\newblock Cryptominisat 2.5.0.
\newblock
  \url{http://baldur.iti.uka.de/sat-race-2010/descriptions/solver_13.pdf},
  2010.

\bibitem{Soos2010SATGauss}
Mate Soos.
\newblock Enhanced {G}aussian elimination in {DPLL}-based {SAT} solvers.
\newblock Pragmatics of SAT, 2010.
\newblock
  \url{http://www.msoos.org/wordpress/wp-content/uploads/2010/08/PoS10-Soos.pdf}.

\bibitem{CryptoSAT2009}
Mate Soos, Karsten Nohl, and Claude Castelluccia.
\newblock Extending {SAT} solvers to cryptographic problems.
\newblock In Kullmann \cite{Swansea2009}, pages 244--257.
\newblock
  \url{http://planete.inrialpes.fr/~soos/publications/Extending_SAT_2009.pdf}.

\bibitem{Sorensson2010Minisat22}
Niklas S\"{o}rensson.
\newblock Minisat 2.2 and minisat++ 1.1.
\newblock
  \url{http://baldur.iti.uka.de/sat-race-2010/descriptions/solver_25+26.pdf},
  2010.

\bibitem{Sperner1928SubsetsFiniteSets}
Emanuel Sperner.
\newblock Ein {S}atz \"{u}ber {U}ntermengen einer endlichen {M}enge.
\newblock {\em Mathematische Zeitschrift}, 27(1):544--548, 1928.

\bibitem{SS90}
Gunnar St{\r{a}}lmarck and M.~S{\"{a}}flund.
\newblock Modeling and verifying systems and software in propositional logic.
\newblock In B.K. Daniels, editor, {\em Safety of Computer Control Systems
  (SAFECOMP'90)}, pages 31--36, 1990.

\bibitem{TamuraTagaKitagawaBanbara2009OrderEncoding}
Naoyuki Tamura, Akiko Taga, Satoshi Kitagawa, and Mutsunori Banbara.
\newblock Compiling finite linear {CSP} into {SAT}.
\newblock {\em Constraints}, 14(2):254--272, 2009.

\bibitem{2011CompactOrderEncoding}
Tomoya Tanjo, Naoyuki Tamura, and Mutsunori Banbara.
\newblock A compact and efficient {SAT}-encoding of finite domain {CSP}.
\newblock In Karem~A. Sakallah and Laurent Simon, editors, {\em Theory and
  Applications of Satisfiability Testing - SAT 2011}, volume LNCS 6695 of {\em
  Lecture Notes in Computer Science}, pages 375--376. Springer, 2011.
\newblock ISBN-13 978-3-642-14185-0.

\bibitem{Urq95}
Alasdair Urquhart.
\newblock The complexity of propositional proofs.
\newblock {\em The Bulletin of Symbolic Logic}, 1(4):425--467, 1995.

\bibitem{Ma99j}
Hans van Maaren.
\newblock A short note on some tractable cases of the satisfiability problem.
\newblock {\em Information and Computation}, 158(2):125--130, May 2000.

\bibitem{Vlcek2010ClassesBoolPolySLUR}
V.~Vl{\v{c}}ek.
\newblock Classes of boolean formulae with effectively solvable {SAT}.
\newblock In Jana Safrankova and Jiri Pavlu, editors, {\em Proceedings of the
  19th Annual Conference of Doctoral Students - WDS 2010}, volume~1, pages
  42--47. Matfyzpress, 2010.

\end{thebibliography}

\newcommand{\noopsort}[1]{}

\end{document}